%% file: main.tex
\documentclass[10pt,twocolumn,letterpaper]{article}

\usepackage{cvpr}              

\usepackage{xspace}
\usepackage[dvipsnames]{xcolor}
\usepackage[ruled,vlined]{algorithm2e}
\usepackage{algpseudocode}
\usepackage{tabularx}
\usepackage{multirow}
\usepackage[pagebackref=true,breaklinks=true,colorlinks,bookmarks=false]{hyperref}

\usepackage{url}
\usepackage[capitalize]{cleveref}

\Crefname{figure}{Fig.}{Figs.}
\Crefname{table}{Tab.}{Tabs.}
\Crefname{section}{Sec.}{Secs.}
\Crefname{appendix}{App.}{Apps.}

\definecolor{vangoghblue}{rgb}{0.1,0.5,1.0}
\definecolor{turtlegreen}{rgb}{0.3,0.7,0.2}

\newcommand{\mypara}[1]{{\noindent\textbf{#1.}\enspace}}

\AtBeginDocument{%
  }

\newcommand{\method}{Meta 3D TextureGen\xspace}

\begin{document}

\title{\method: Fast and Consistent Texture Generation for 3D Objects}

\author{Raphael Bensadoun$^*$,\: Yanir Kleiman$^*$,\: Idan Azuri,\: Omri Harosh, \\ \vspace{-0.35cm} \\ Andrea Vedaldi,\: Natalia Neverova,\: Oran Gafni 
\\
\\
GenAI, Meta\\ 
{\tt\small \{raphaelbens,yanirk,idanazuri,omrih,vedaldi,nneverova,oran\}@meta.com}}\vspace{-1.5cm}

\twocolumn[{%
\renewcommand\twocolumn[1][]{#1}%
\maketitle
\vspace{-0.7cm}
\begin{center}
    \centering
    \captionsetup{type=figure}

    \includegraphics[width=\textwidth]{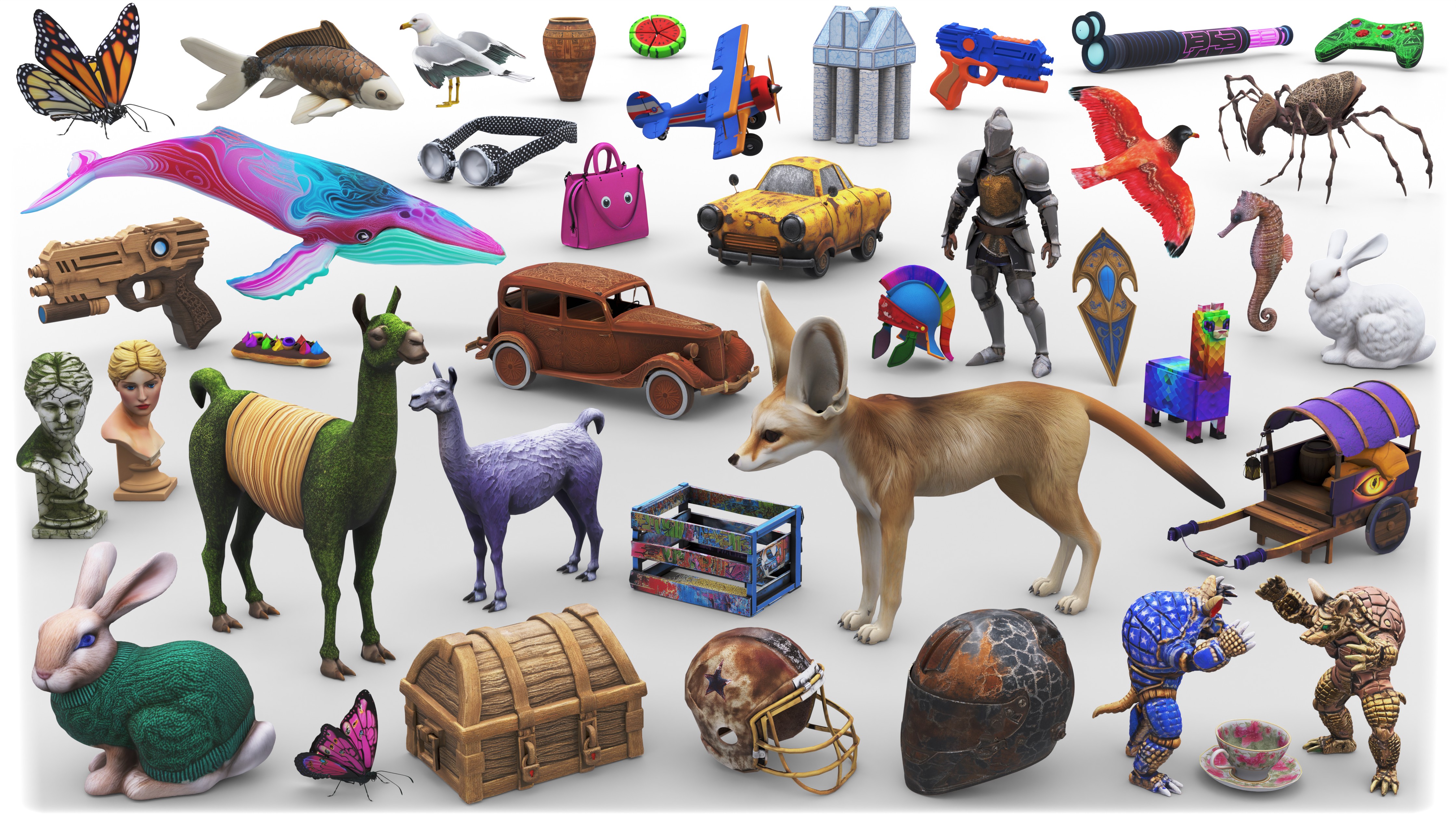}
    \vspace{-0.5cm}
    \captionof{figure}{\method: examples of generated textures. Given a 3D shape and a textual prompt, our method generates globally consistent, high-quality textures in under $20$ seconds, while maintaining text faithfulness for both realistic and stylized text prompts.}
    \label{fig:teaser}
\end{center}%
}]
\def\thefootnote{*}\footnotetext{Equal contribution}


\begin{abstract}

The recent availability and adaptability of text-to-image models has sparked a new era in many related domains that benefit from the learned text priors as well as high-quality and fast generation capabilities, one of which is texture generation for 3D objects.
Although recent texture generation methods achieve impressive results by using text-to-image networks, the combination of global consistency, quality, and speed, which is crucial for advancing texture generation to real-world applications, remains elusive.
  
To that end, we introduce \method: a new feedforward method comprised of two sequential networks aimed at generating high-quality and globally consistent textures for arbitrary geometries of any complexity degree in less than $20$ seconds.
Our method achieves state-of-the-art results in quality and speed by conditioning a text-to-image model on 3D semantics in 2D space and fusing them into a complete and high-resolution UV texture map, as demonstrated by extensive qualitative and quantitative evaluations.
In addition, we introduce a texture enhancement network that is capable of up-scaling any texture by an arbitrary ratio, producing $4k$ pixel resolution textures.

\end{abstract}

\input{sections/1_intro}

\input{sections/fig_arch}

\input{sections/2_related}

\input{sections/3_method}

\input{sections/4_experiments}

\input{sections/5_limitations}

\input{sections/6_conclusions}

\clearpage

{
    \small
    \bibliographystyle{ieeenat_fullname}
    \bibliography{sample-base}
}

\clearpage

\input{appendix}

\end{document}

%% file: sections/1_intro.tex
\section{Introduction}%
\label{sec:intro}

3D generative models have advanced considerably, in part thanks to the impressive progress in text-to-image~\cite{ramesh2021zero,gafni2022make,ramesh2022hierarchical,saharia2022photorealistic,rombach2022high,dai2023emu} and text-to-video~\cite{singer2022make,ho2022imagen,girdhar2023emu} generation.
These advances concern three related fronts:
(i) generation of 3D shapes, including the development of new and powerful shape representations~\cite{yariv2023mosaic,siddiqui2023meshgpt,nash2020polygen,alliegro2023polydiff,chen2020bsp},
(ii) generation of textures~\cite{metzer2023latent,cao2023texfusion,chen2023text2tex,richardson2023texture}; and
(iii) combined generation of shape and texture, often called `text-to-3D'~\cite{li2023instant3d,shi2023mvdream,poole2022dreamfusion,wang2024prolificdreamer,lin2023magic3d}.
As new shape representations usually include appearance information too, areas (i) and (iii) are converging.
However, texture generation remains important, as it allows to control appearance independently of shape, and is applicable to any 3D asset, whether produced by an artist or generated automatically.

``Moonlight is sculpture; sunlight is painting’’. After the subtleties of geometry, textures and colors add a remarkable layer of expressiveness, as implied in this famous quote by Nathaniel Hawthorne~\cite{hawthorne1896passages}.
Creating textures is a key mode of expression for 3D artists and crucial to the impact of 3D content in applications such as gaming, animation, and virtual/mixed reality.
However, creating high-quality and diverse textures, whether realistic or stylized, is difficult and time-consuming, particularly for complex 3D shapes, and requires specific professional skills.

Contrary to image and video generation, where billions of images and videos are available for training, 3D generation is hampered by the lack of large-scale 3D datasets.
For this reason, 3D generation networks, including texture generation, are often \emph{derived} from pre-trained image or video generation networks.
This allows texture generators to inherit some of the qualities of their peers, including realism, faithfulness and open-ended nature, while only utilizing a comparatively small amount of 3D training data.
However, there are still significant quality and speed gaps between texture and 2D image and video generation:

{\textbf{(i) Global consistency and text faithfulness}}. 
The gap between the image-text relationship when generating a single image compared to generating a sequence of images or views, translates to a lack of global consistency and text faithfulness in the generated texture. This is further intensified by the strong bias of text-to-image models towards frontal views, as well as their lack of 3D understanding. These inconsistencies range from small texture misalignments (often referred to as ``seams''), to a lack of symmetry or an overall incoherent look, to catastrophic failures such as the ``Janus effect''~\cite{wiki:janus}, where multiple instances of a given anatomical feature (e.g. a face or an eye) appear in multiple places across the object.

\textbf{(ii) Semantic alignment with the target 3D shape}.
The text-to-image model is required to generate texture that fits the given 3D object, and must thus be \emph{conditioned} on its shape.
However, fusing fine 3D shape information into 2D space in a coherent manner, such that fine 3D information is preserved yet translated efficiently to 2D space is difficult to achieve.
Previous attempts generated texture by either conditioning in UV space on vertex or normal maps~\cite{yu2023texture}, or in image space, on depth maps~\cite{zeng2023paint3d}.
However, they struggle with precise alignment and fine-detail preservation, resulting in lower texture quality for highly detailed 3D objects, which is a considerable limitation.

{\textbf{(iii) Inference speed}}.
While previous methods rely on iterative generation for improving global consistency and gaining complete shape coverage, they require multiple generation steps, ranging from several to thousands of forward passes, such as via Score Distillation Sampling (SDS)~\cite{poole2022dreamfusion}.
This results in a long inference time of minutes, which is compute intensive and renders these methods unsuitable for many practical use cases, such as user-generated content applications, or allowing designers to perform quick iterations as part of their creative process.

We introduce \method, a new texture generation method that successfully addresses these gaps, while attaining state-of-the-art results.
Our method is fast, as it only requires a single forward pass over two diffusion processes. The method achieves excellent view and shape consistency, as well as text fidelity, by conditioning the first fine-tuned text-to-image model on 2D renders of 3D features, and generating all texture views jointly, accounting for their statistical dependencies and effectively eliminating global consistency issues such as the Janus problem.

The second image-to-image network operates in UV space, it creates a high-quality output by completing missing information, removing residual artifacts, and enhancing the effective resolution, bringing our generated textures to being close to application-ready.
Moreover, we introduce an additional network that enhances the texture quality and increases resolution by an arbitrary ratio, effectively achieving a $4$k pixel resolution for the generated textures.

To the best of our knowledge, this is the first approach to achieve high quality and diverse texturing of arbitrary meshes using merely two diffusion-based processes, without resorting to costly interleaved rendering or optimization-based stages. Moreover, this is the first work to explicit condition networks on geometry in 2D, such as position and normal renders in order to encourage local and global consistency, finally alleviating the Janus effect.

Samples of our generated textures are provided on a diverse set of shapes and prompts throughout the paper, as well as on static and animated shapes in the \href{https://youtu.be/110Rr2ABCY8}{video}.

%% file: sections/fig_arch.tex
\begin{figure*}[h]
  \centering
  \includegraphics[width=\textwidth]{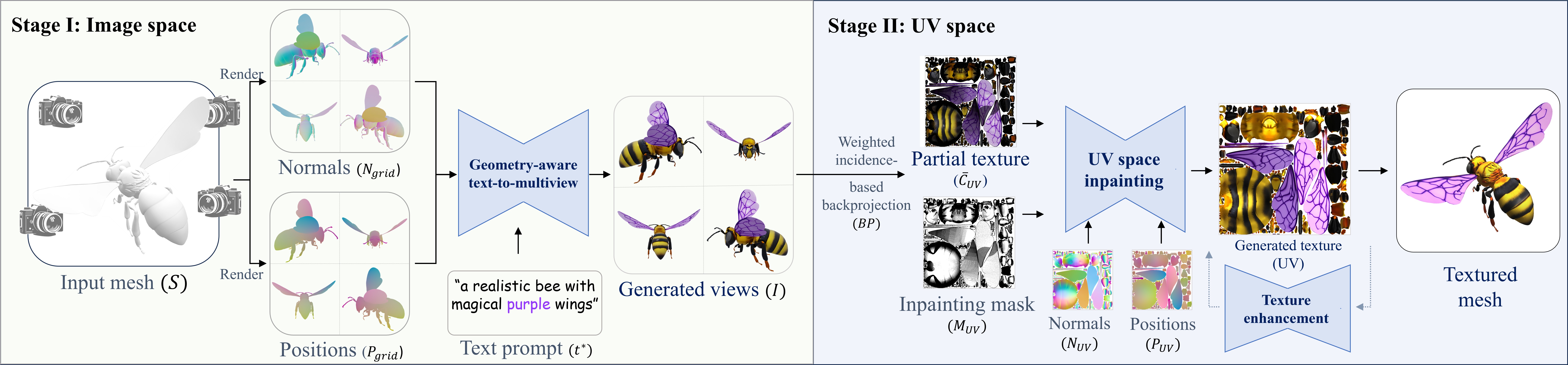}
  \caption{\textbf{Method overview.} Given an input shape and a text prompt, \method generates a globally consistent high-quality texture in less than $20$ seconds.
  The first stage (left) consists of a geometry-aware text-to-image model that generates a multi-view image of the generated texture, conditioned on renders of the normal and position maps over the input mesh.
  The second stage (right) consists of a projection of the generated texture renders back to UV space while taking into account the normals and camera angles (weighted incidence).
  The combined backprojections are then fed into the UV-space inpainting network along with a guiding inpainting mask, as well as the vertex and position UV maps, which generates a complete texture map in UV space.
  The generated texture map can optionally go through a MultiDiffusion texture enhancement network to increase the resolution by an arbirary ratio.}

  \label{fig:architecture}
\end{figure*}

%% file: sections/2_related.tex
\section{Related work}
\label{sec:related}
\subsection{Image generation}
A number of architectures have been proposed for text-to-image synthesis, including earlier efforts using Generative Adversarial Networks~\cite{goodfellow2014generative}. Some more recent variants are based on transformers (e.g. 
DALL-E~\cite{dalleSpotlight}, 
CogView~\cite{ding2021cogview}, 
Make-a-Scene~\cite{gafni2022make}, 
Parti~\cite{yu2022scaling}) and Muse~\cite{chang2023muse}). Another popular class of text-to-image generators builds on pixel-space or latent diffusion models~\cite{ho2020denoising}, including 
eDiff-I~\cite{balaji2022ediffi}, 
Imagen~\cite{saharia2022photorealistic}, 
unCLIP~\cite{ramesh2022hierarchical}, 
Stable Diffusion~\cite{rombach2022high}, 
SDXL~\cite{podell2023sdxl}, 
EMU~\cite{dai2023emu} and others. In this work, we are starting with a pre-trained latent diffusion model with an architecture similar to EMU~\cite{dai2023emu} and further extend it to our task.

\subsection{Multi-view generation} 

The field of multi-view generation, which involves the generation of multiple perspectives of a single object or scene from noise or a few reference images, has demonstrated its utility in the generation of 3D shapes. Zero-1-to-3~\cite{liu2023zero} and Consistent-1-to-3~\cite{ye2023consistent} generate novel views through viewpoint-conditioned diffusion model.  Zero123++~\cite{shi2023zero123++}, MVDream~\cite{shi2023mvdream} and Instant 3D~\cite{li2023instant3d} opt for a grid-like generation of six and four views respectively.
ConsistNet ~\cite{yang2023consistnet} use a different diffusion process for each view and introduce a 3D pooling mechanism to share information between views.
Additional layers and architectures to enhance multi-view consistency are proposed by SyncDreamer~\cite{liu2023syncdreamer}, Consistent123~\cite{weng2023consistent123}, DMV3D~\cite{xu2023dmv3d} and MVDiffusion++~\cite{tang2024mvdiffusion++} which denoise multiple views of the 3D object simultaneously.
The obtained multi-view images in these works are then utilized as guidance to reconstruct the texture and geometry of a 3D object.

In contrast to our task of texture generation, these models are designed for the generation of 3D objects, where the geometry is not predetermined and is concurrently produced with the texture. This application inherently provides the flexibility to modify the geometry to achieve more consistent multi-view images, for both texture and geometry.

\subsection{Texture generation}\label{sec:texture-generation}
Texture generation aims to create high-quality and realistic or stylized textures for 3D objects based on textual descriptions.
Early works, such as CLIP-Mesh~\cite{mohammad2022clip} and Text2Mesh~\cite{michel2022text2mesh} proposed to optimize a texture via differentiable rendering, using CLIP~\cite{radford2021learning} guidance to match the text prompt.
Other optimization-based methods such as Fantasia3D~\cite{chen2023fantasia3d}, Latent-Paint ~\cite{metzer2023latent} and Paint-It~\cite{youwang2023paint}, combine differentiable rendering with SDS~\cite{poole2022dreamfusion} to utilize gradients from diffusion models.
Texturify~\cite{siddiqui2022texturify}
and Mesh2Tex~\cite{bokhovkin2023mesh2tex} opt for a GAN-based approach incorporating a latent texture code and a mapping network similarly to StyleGAN~\cite{karras2019style}.
The rapid emergence of large-scale text-to-image models, particularly diffusion models, has led to several advancements in texture generation. Several methods, such as TexDreamer~\cite{liu2024texdreamer} and Geometry Aware Texturing~\cite{10.1145/3610542.3626152} aim to generate a UV map in a straight-forward manner, applying the diffusion process directly in UV space.
While these methods tend to be fast, they are limited to human texture generation and clothing items respectively, and cannot generalize to arbitrary objects. Point-UV Diffusion~\cite{yu2023texture} proposes a point-cloud diffusion approach to generate a colored point-cloud, which colors are subsequently projected onto the UV map for further refinement, yet requires to train a separate model for each object category, and does not generalize to arbitrary objects. 

A significant area of work, which includes TEXTure~\cite{richardson2023texture}, Text2Tex~\cite{chen2023text2tex}, Intex~\cite{tang2024intex} and Paint3D~\cite{zeng2023paint3d}, consists of iterative inpainting using pre-trained depth-to-image diffusion models in a zero-shot manner.
This involves generating a single view at a time and iteratively rotating the mesh until a sufficient area is covered, using interleaved renderings as guidance for further inpainting steps. While these approaches are training-free, their inference runtime is significant and can take a few minutes for a single generated texture. Moreover, they are not 3D-aware and are prone to producing artifacts such as the ``Janus'' effect. SyncMVD~\cite{liu2023text} adopted the same zero-shot approach while employing different diffusion processes for each view and synchronizing the output at each step, leading to better quality textures, yet suffering from the same global consistency issues.
TexFusion~\cite{cao2023texfusion} alleviates consistency issues by adding a module which performs denoising diffusion iterations in multiple camera views and aggregates them through a latent texture map after every denoising step. 
FlashTex~\cite{deng2024flashtex}, similarly to our approach, trains on a 3D dataset and generates a four-view grid. As conditioning, they use renderings of the shape with three different materials, which are then combined into a single three-channel image.
Subsequently, they use an SDS optimization-based stage to distill information from their trained multi-view model, resulting in a significant runtime of $2$ minutes. Meshy~\cite{meshy3p0}, a commercial product for which we do not have the complete technical details, tends to produce better results quality-wise than some methods mentioned above. Yet, their textures exhibit global inconsistencies, as well as over-saturated colors, text alignment issues and blurred inpainting of self occlusions.

%% file: sections/3_method.tex
\section{Preliminaries and data processing}

Our method takes a representation of the 3D shape features in the form of rendered images and baked texture maps in UV space, which are used in the first and second stage respectively.
Here we detail the different channels that we render for each shape.

\subsection{Shape renders}

We render the following channels for each shape. Each channel is rendered from four views which are stitched to a single image.

\mypara{Combined pass} 
As ground truth data used for training, which is not extracted at inference time,
we render the shape with all material properties.
This render, often referred to as ``beauty pass'', preserves lighting effects and material properties that are applied to the object.
These are crucial to preserve to correctly represent different types of materials such as wood, plastic, metal, etc., which react differently to light and thus cannot be represented faithfully using only their diffuse color.
We use Blender~\cite{blender} to render the combined pass with even lighting from all directions.

\mypara{Position and normal passes}
These are used as conditioning for training and inference. Each pixel in the position pass represents the XYZ position of the corresponding point on the shape, and each pixel in the normal pass represents the normal direction of the shape at the corresponding point. Both are normalized to the range $[0,1]$ and rendered without lighting, hence written as-is to the output image.

\subsection{UV maps}
We bake each channel into a texture in UV space.
This process involves producing a UV layout for each shape and baking the texture to an image.

\mypara{UV layout}
Our in-house dataset contains objects from various sources which may have various UV layouts, from layouts meticulously created by an artist, to scanned objects with a procedurally generated layout, and objects with partial or corrupt UV layout.
A single object may contain many texture files, in which case the UV layout of each part may overlap the layout of parts that are mapped to a different texture.
For our method, we require a UV layout that maps the shape onto a single square texture with no overlapping UV islands, so we automatically rearrange the UV islands of the shape such that there is no overlap between them.
For objects that do not have a suitable UV map, we generate a new UV map using Blender's \emph{Smart Project} feature, and filter out objects for which this process fails to produce a desirable UV layout.

\mypara{Baked channels}
We use Blender to bake the \emph{combined}, \emph{position}, and \emph{normal} passes mentioned above to the UV space. Baking a texture is a similar process to rendering an object, but the rendered pixel are written to the corresponding location on the UV map rather than being painted in the render view.
The \emph{combined} pass is used as the target image for training, while the \emph{position} and \emph{normal} passes are used as conditioning for the network.

\input{figures/depth/fig_depth}

\mypara{Backprojected textures}
To simulate the input textures that are produced by the first stage, we take the color renders of the shape and project them onto the texture in UV space, using the same process as described in \cref{sec:backproj}.
The network goal is to reconstruct the full texture map from these partial views.

\input{figures/comparisons/fig_comparison_short}

\section{Method}%
\label{sec:method}

Given a 3D object and description of a desired texture, \method produces as output a corresponding texture in UV space.
As shown in \cref{fig:architecture}, \method employs a two-stage approach.
The first stage operates in image space, conditioned on a text description and renders of the 3D shape features, and produces renders of the textured shape from multiple views.
The second stage operates in UV space, taking a weighted incidence-based backprojection of the first stage output as condition, as well as the 3D shape features used for the first stage, but in UV space.
The end result of the second stage is a complete UV texture map which is consistent between different views and matches the text prompt.
An optional extension of the second stage is a texture enhancement network that extends the MultiDiffusion~\cite{bar2023multidiffusion} approach from 1D to 2D image-patch overlaps, increasing the texture map resolution by $\times4$.

\input{figures/bunnies/fig_bunnies}

As demonstrated in our experiments (\cref{sec:experiments}), by conditioning the fine-tuned text-to-image model on renders of 3D shape features while generating all views in tandem, the first stage is able to generate diverse yet globally consistent renders of textured 3D shapes, while the second stage focuses on generating the missing areas that are occluded in image space and improving the overall quality of the generated texture map.

Next, we provide a detailed overview of each stage.
We focus here on the novel or unusual aspects of our method and refer the reader to the supplement for details. 

\subsection{Stage I: Generation in image space}%
\label{sec:stage1}

The goal of the first stage is to generate globally consistent 
images of a given 3D object based on a textual description of the desired output.
To this end, we use a diffusion-based neural network fine-tuned from a pre-trained image generator.
In order to produce consistent views that match the given 3D object, the network takes as input a grid of position and normal renders from multiple angles, in addition to the text conditioning.
Specifically, for each channel we produce a grid of $4$ matching viewpoints and combine them to a single image. The four viewpoints are fixed at training and inference time, and provide a $360${\textdegree} view of the object at $90${\textdegree} intervals, with a fixed elevation angle of $20${\textdegree}.

\subsubsection{Geometry-aware 2D conditioning.}
Multiple methods~\cite{yu2023texture,chen2023text2tex,zeng2023paint3d} use depth maps as a way to represent 3D assets in 2D images leveraging depth-conditioned pre-trained diffusion models in a zero-shot manner.
In contrast, we advocate for the use of position and normal renders.

As seen in \cref{fig:depth}, the additional information in these representations provides the following benefits for using them as conditioning:
(i) position values are global and not view-dependent, providing point correspondence between the same points on the object in different views, thus encouraging 3D consistency;
(ii) normal renders provide orientation information and fine geometric details of the mesh to guide the generation model, which can be difficult to capture with depth.

\subsubsection{Multi-view image generation from text}
The first stage consists of a U-Net based latent diffusion model, fine-tuned from a model with a similar architecture to Emu~\cite{dai2023emu} denoted by $f$.
Its goal is to generate a grid of four consistent views of an arbitrary mesh $S$, in image space, guided by a text prompt $t^*$, denoted by $I$.
For this purpose, the diffusion model is conditioned on two grids of matching position and normal renders, denoted as $\text{P}_\text{grid}(S)$ and $\text{N}_\text{grid}(S)$ respectively. 

The generated multi-view image grid $I$ can then be formulated as follows:
\begin{equation}
        I(S, t^*) = f(z, t^*, \text{P}_\text{grid}(S), \text{N}_\text{grid}(S)),
\end{equation}
where $z$ is 2D noise map where each pixel is sampled i.i.d.~from a standard Gaussian distribution.
Note that in this equation $t^*$ denotes the textual prompt; in practice, this network is also conditioned on the diffusion step, sometime called `time'.
We do not show it here explicitly for succinctness and clarity.

\subsection{Stage II: Generation in UV space}

The goal of the second stage is to generate the final texture in UV space. Given the viewpoints from the first stage output, the network aims at inpainting missing areas due to self occlusions and improving the overall quality of the generated texture, in UV space.
The inputs for the second stage are the partial texture map, obtained by backprojecting and blending the views generated by the first stage, in addition to the position and normal UV maps.

\subsubsection{Backprojection and incidence-based weighted blending}%
\label{sec:backproj}

Backprojection is a technique where a 2D image or projection is mapped onto the UV texture map of a 3D model.
This involves identifying the corresponding face on the 3D model for each non-background pixel in the image and assigning the color value at the corresponding coordinate in the texture map.

Although the first stage results in highly consistent views of the generated texture due to the conditioning on 3D semantics, we have observed, similarly to previous works~ \cite{yu2023texture,chen2023text2tex,liu2023text}, that textures generated over areas that are not facing the camera (low incidence angles) are less reliable.
This can lead to artifacts when na{\"\i}vely averaging different texture views together, particularly in areas with high frequency details such as fine patterns or writings.
To overcome this issue, similarly to SyncMVD~\cite{liu2023text}, we blend the backprojections into a single UV map using a weighted average by the incidence angles.
Specifically, we utilize the cosine similarity between the viewing direction and per-pixel normal vectors in image space to determine per-pixel weight contributions to the blended texture. 
Formally, the incidence of a pixel $p$ in a rendering $I^i_S$ of a 3D shape $S$, for each view $i$ (which we denote by $\phi(I^i_{S}, p)$) is defined as $\phi(I^i_{S}, p) = cos(\theta_{\vec{v_{i}}(p),\vec{n}(I^i_{S}, p)})$ where $\theta_{\vec{x},\vec{y}}$ is the angle between $\vec{x}$ and $\vec{y}$, $\vec{v_{i}}(p)$ is the viewing direction from camera $i$ to pixel $p$, and $\vec{n}(I^i_{S}, p)$ is the normal vector at pixel $p$ of the rendered shape $S$ from camera $i$.\\
Finally, denoting the backprojection operation as $\operatorname{BP}$, we define each pixel $p$ of the blended partial texture $\bar{C}_\text{UV}(S, t^*)$ as follows:
\begin{equation}
    \bar{C}_{UV}^{p}(S, t^*) = \frac{\sum_{j=0}^n \operatorname{BP}(I(S, t^*)_{j}^{p}) \odot \operatorname{BP}(\phi(I(S,t^*)_{j} , p)^\alpha)}{\sum_{j=0}^n \operatorname{BP}(\phi(I(S,t^*)_j, p))^\alpha) + \epsilon},
\end{equation}
where $I(S,t^*)_{j}$ is the $j$'th view of $I(S,t^*)$ and $I(S,t)_{j}^p$ is the pixel $p$ of $I(S,t^*)_{j}$.
$\epsilon$ is a small constant to avoid zero division.
We use $n=4$, as the number of generated views and  $\alpha = 6$ for all of our experiments.

\subsubsection{UV-space inpainting network}

The first stage followed by the weighted backprojection operator results in a texture map that is sparse in a varying degree, depending on the input shape.
The degree of sparsity is determined by two factors:
(i) occlusions caused by insufficient coverage of the selected views in respect to the shape structure, resulting in missing areas, and
(ii) pixel-level ``holes'' resulting from the absence of one-to-one correspondence between each occupied pixel in the generated rendering and the UV map.
To obtain the full texture, we opt for an inpainting approach.

Similarly to Stage I, the inpainting is modeled by a U-Net based latent diffusion model fine-tuned from the same pre-trained network which we denote by $g$.
$g$ is conditioned on the blended partial map $\bar{C}_\text{UV}(S, t^*)$, the inpainting mask denoting the missing areas and pixels to inpaint $M_\text{UV}(S)$, along with $\text{P}_\text{UV}(S)$ and $\text{N}_\text{UV}(S)$, to obtain the final texture map $\text{Texture}(S, t^*)$ as follows:
\begin{equation}
\text{Texture}(S, t^*) = g(z, \bar{C}_\text{UV}(S, t^*), M_\text{UV}(S), \text{P}_\text{UV}(S),\text{N}_\text{UV}(S)),
\end{equation}
where $z$ is a 2D noise map where each pixel is sampled i.i.d.~from a standard Gaussian distribution. 

\subsubsection{Texture enhancement network}
Our two-stage texture generation approach yields a text-aligned, high-quality and consistent UV texture map at a resolution of $1024\times1024$ pixels.
While this resolution is satisfying for some applications, other applications may require a higher resolution of $4$k ($4096\times4096)$ pixels.
To that end, we introduce an additional, yet optional component to the second stage for up-scaling the generated texture map resolution and quality. This is the  texture enhancement network, which is flexible in terms of the output resolution and ratio, as it operates in a patched-based fashion.

The reason for employing a patch-based approach~\cite{ozdenizci2023restoring,wang2023exploiting} is due to the memory limitations of current GPUs that do not support the generation of $4$k resolution images.
As patch-based prediction results in inconsistencies between different patches, manifesting both locally (seams) and globally as pattern/color mismatches, we extend the MultiDiffusion~\cite{bar2023multidiffusion} approach from 1D image-patch overlaps to 2D (panoramas to square-shaped images) to mitigate these issues, aggregating the different latent patches and applying a weighted Gaussian average at each diffusion time step. In addition, we employ a tiled-VAE approach for the encoder-decoder to enable the encoding and decoding of high-resolution textures.

%% file: figures/depth/fig_depth.tex
\begin{figure}[t]
\centering

\begin{minipage}{\linewidth}
	\centering

    \begin{minipage}{0.32\linewidth}
        \centering
        \includegraphics[width=0.9\linewidth]{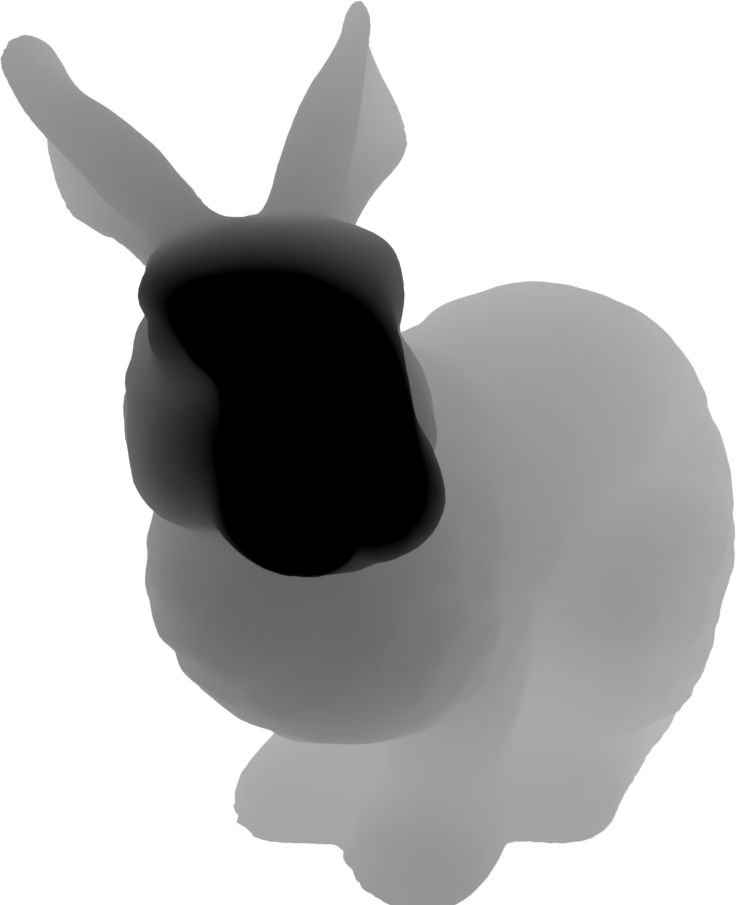}
        \includegraphics[width=0.9\linewidth]{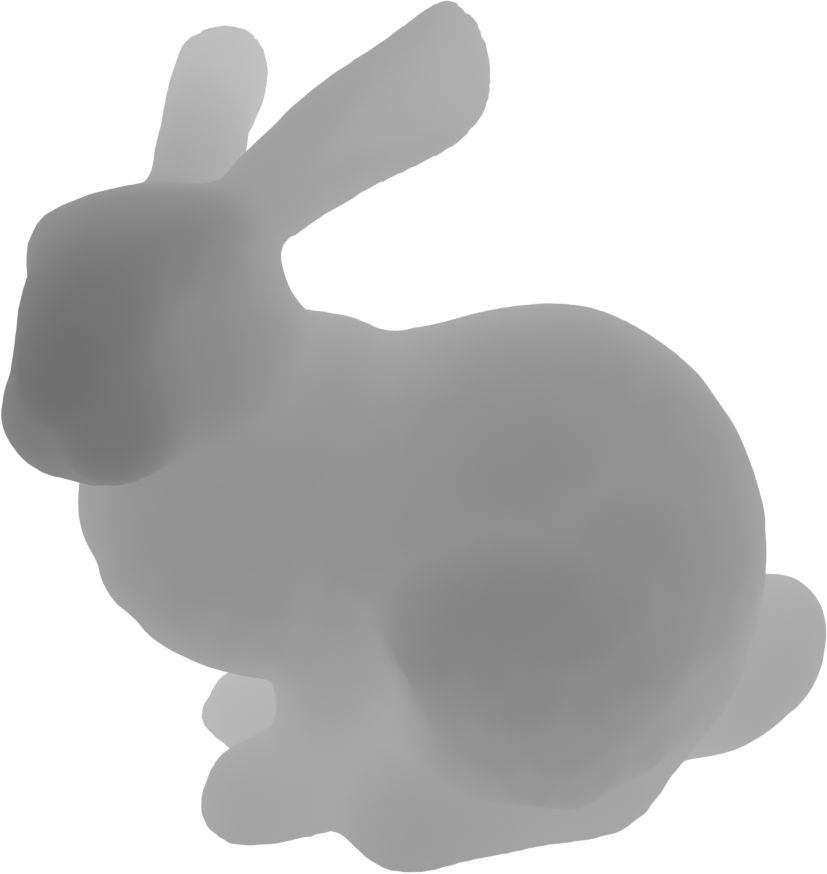}
        \\
	    (a)
	\end{minipage}
    \begin{minipage}{0.32\linewidth}
        \centering
        \includegraphics[width=0.9\linewidth]{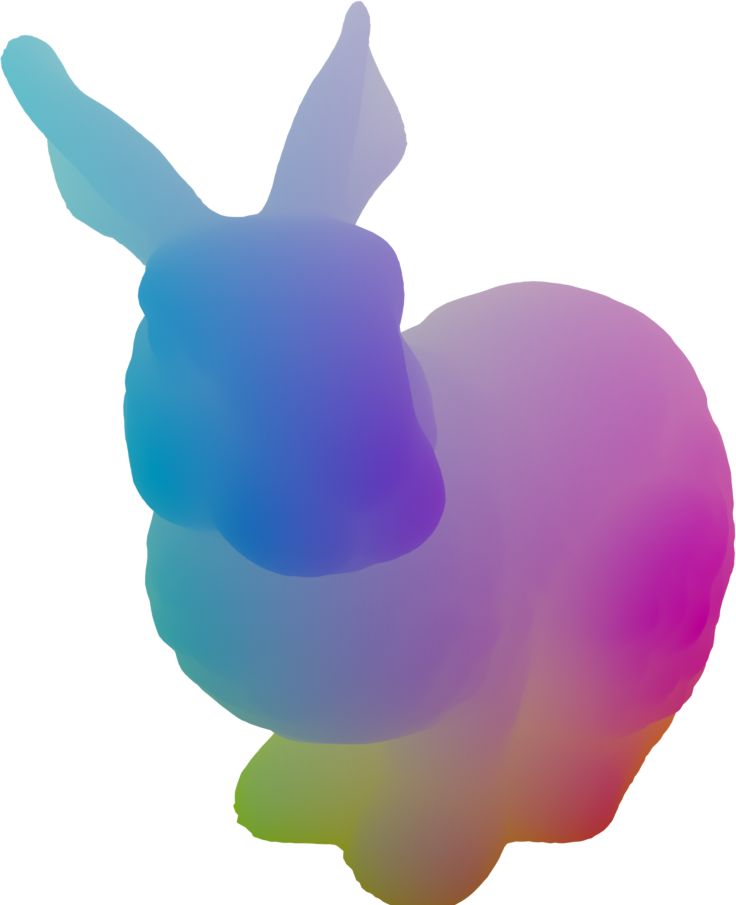}
        \includegraphics[width=0.9\linewidth]{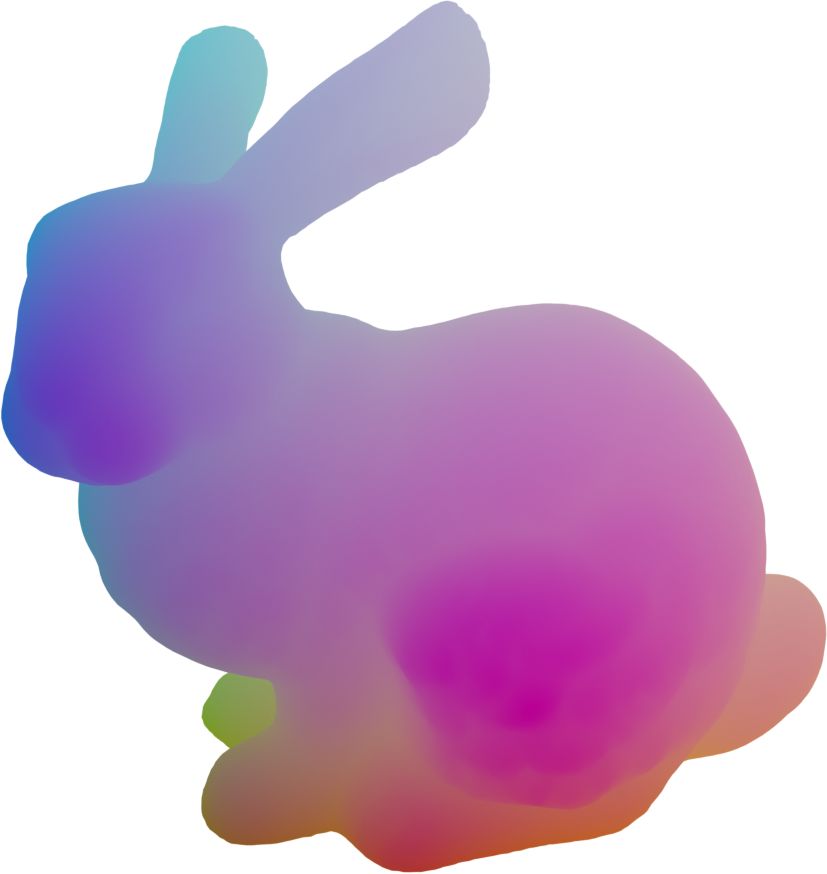}
        \\
	    (b)
	\end{minipage}
    \begin{minipage}{0.32\linewidth}
        \centering
        \includegraphics[width=0.9\linewidth]{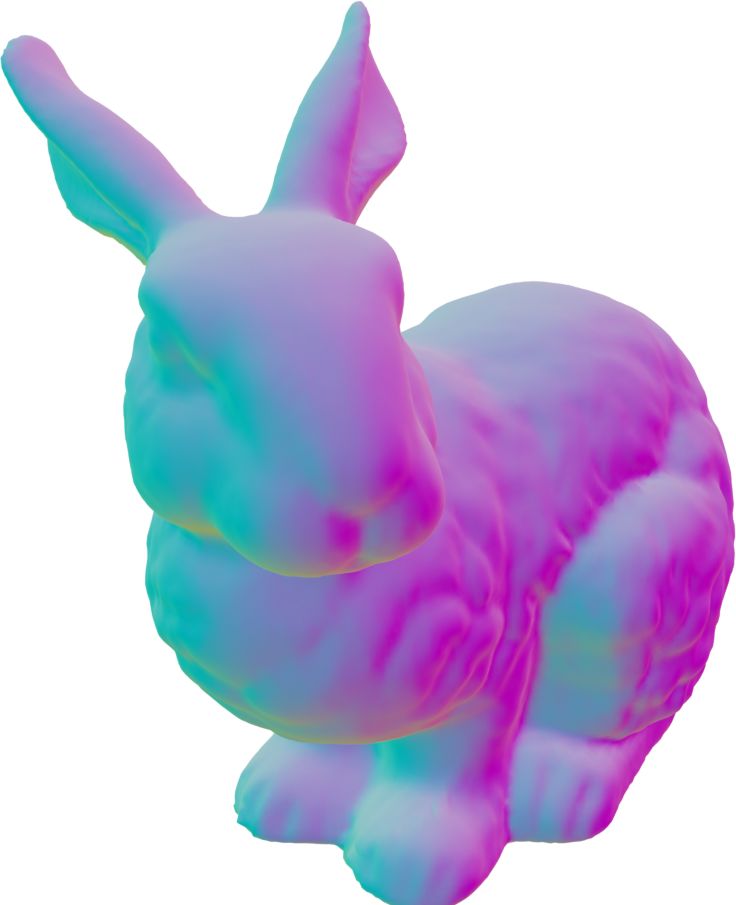}
        \includegraphics[width=0.9\linewidth]{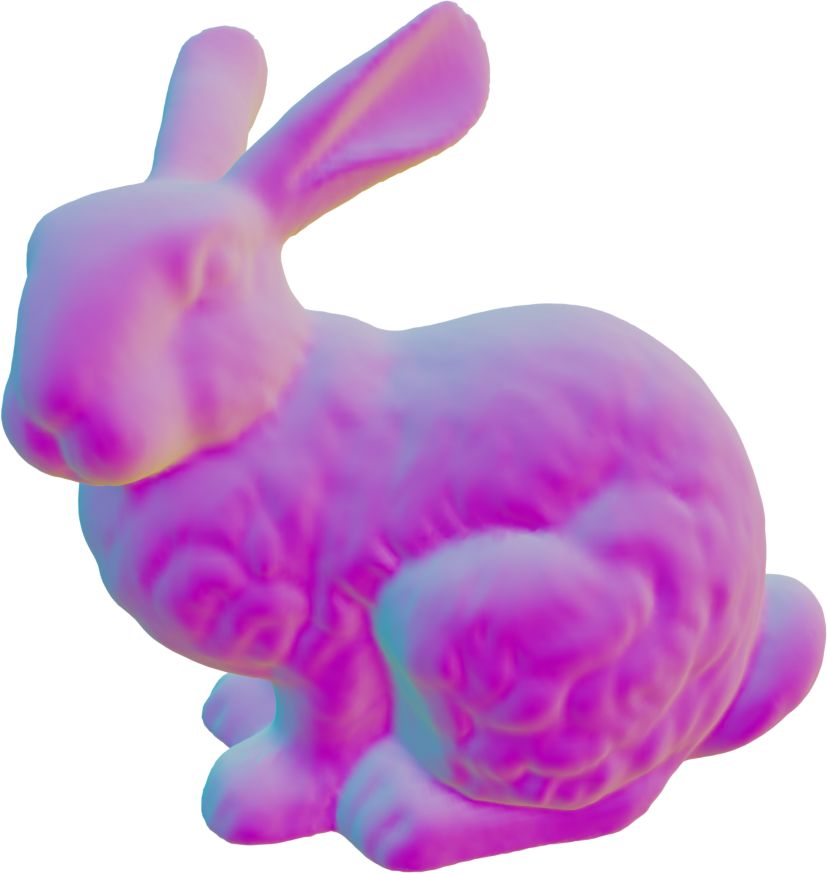}
        \\
	    (c)
	\end{minipage}


\end{minipage}

%
		%
		%
%
%

\caption{Contrary to (a) depth renders, (b) position renders are global rather than view-dependent, and (c) normal renders contain high-frequency details.} 
  \label{fig:depth}
\end{figure}

%% file: figures/comparisons/fig_comparison_short.tex
\begin{figure*}[t]
\centering

\begin{minipage}{0.98\linewidth}
	\centering

	\begin{minipage}{0.16\linewidth}
	    \centering
        \textbf{Ours}
	\end{minipage}
	\begin{minipage}{0.16\linewidth}
	    \centering
        TEXTure
	\end{minipage}
	\begin{minipage}{0.16\linewidth}
	    \centering
        Text2Tex
	\end{minipage}
	\begin{minipage}{0.16\linewidth}
	    \centering
        SyncMVD
	\end{minipage}
	\begin{minipage}{0.16\linewidth}
	    \centering
        Paint3D
	\end{minipage}
	\begin{minipage}{0.16\linewidth}
	    \centering
        Meshy 3.0
	\end{minipage}
    \vspace{2mm}

	\begin{minipage}{0.16\linewidth}
	    \centering
	    \includegraphics[width=0.48\linewidth]{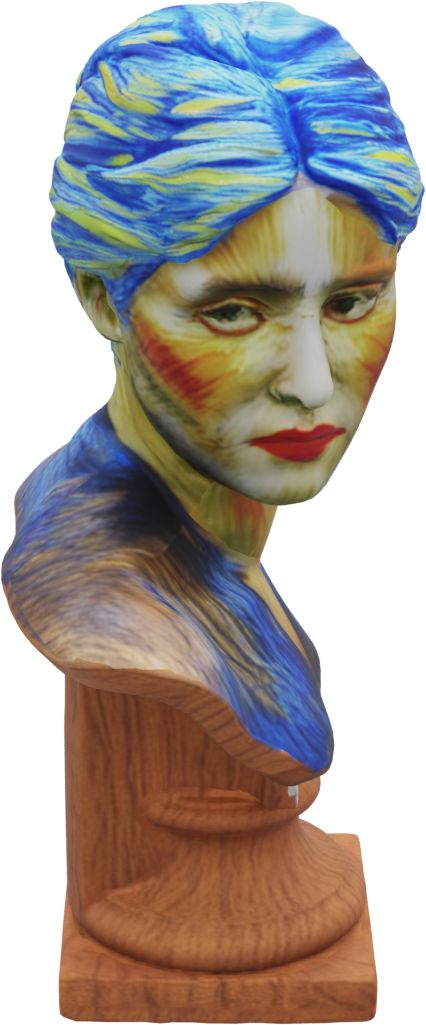}
	    \includegraphics[width=0.48\linewidth]{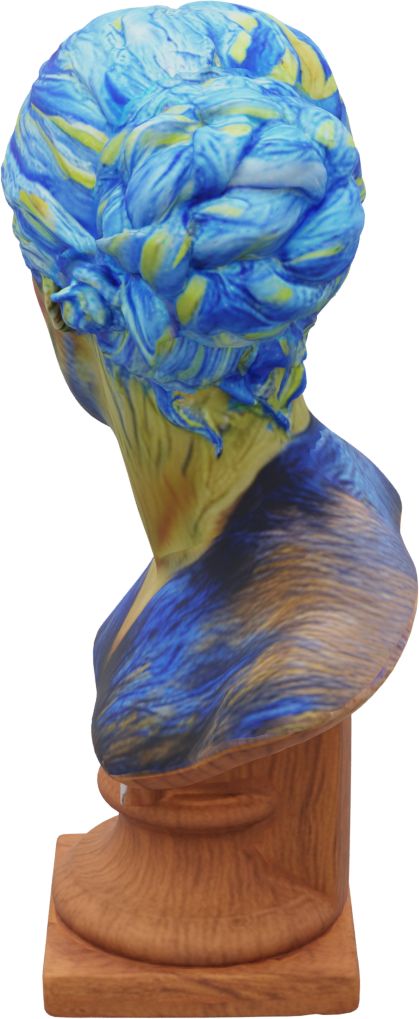}
	\end{minipage}
	\begin{minipage}{0.16\linewidth}
	    \centering
	    \includegraphics[width=0.48\linewidth]{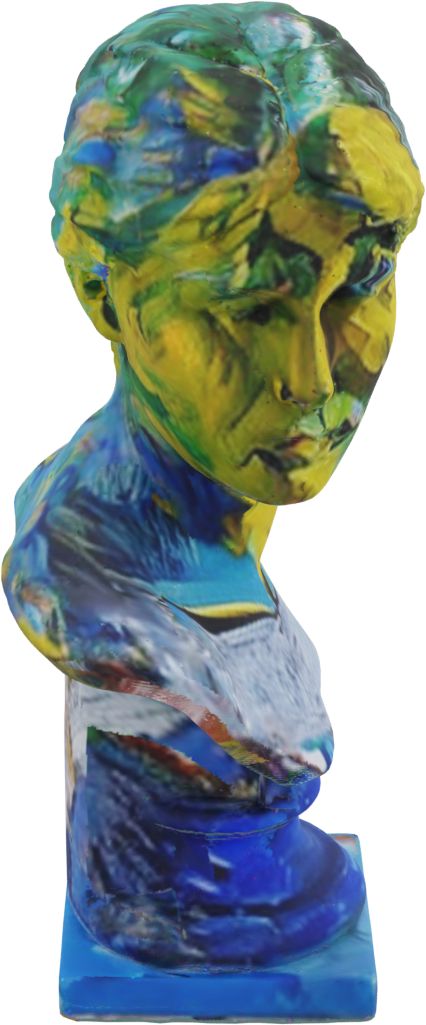}
	    \includegraphics[width=0.48\linewidth]{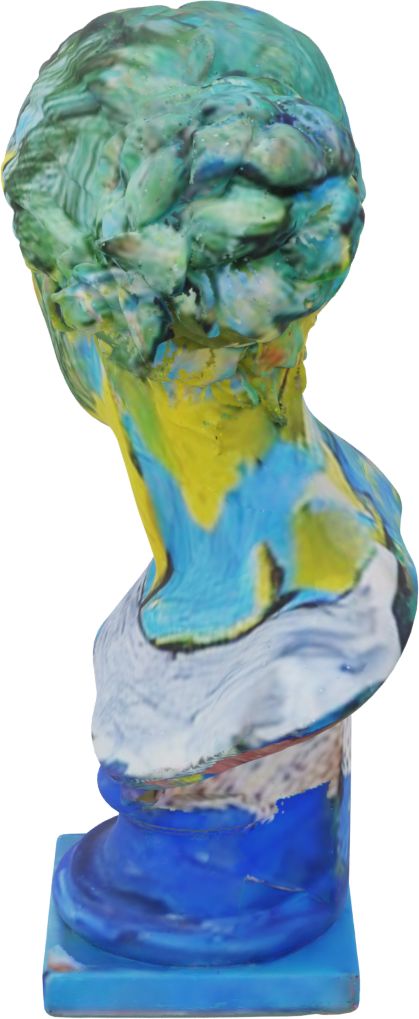}
	\end{minipage}
	\begin{minipage}{0.16\linewidth}
	    \centering
	    \includegraphics[width=0.48\linewidth]{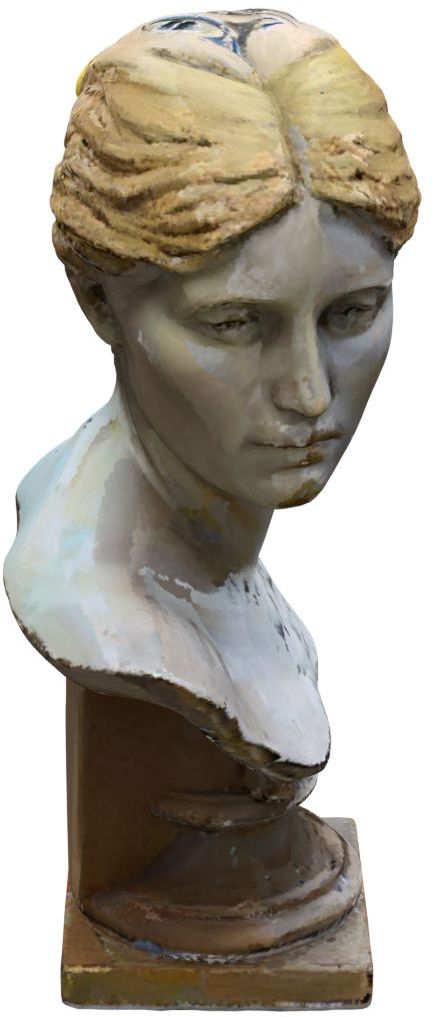}
	    \includegraphics[width=0.48\linewidth]{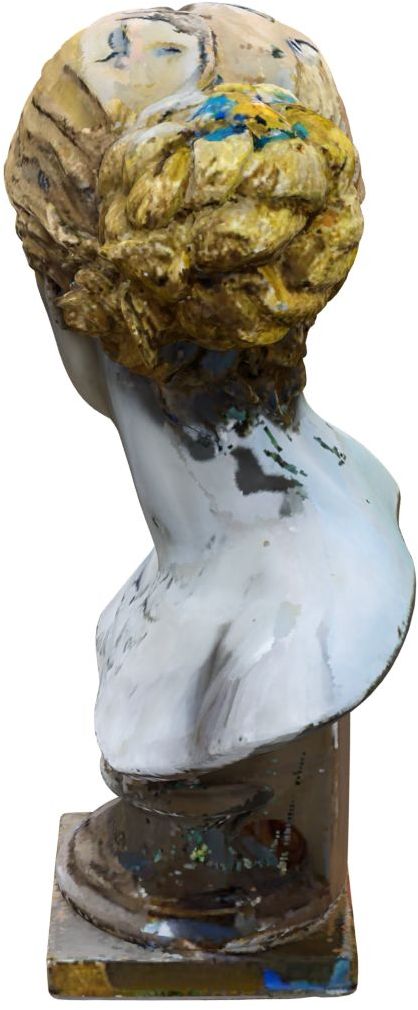}
	\end{minipage}
	\begin{minipage}{0.16\linewidth}
	    \centering
	    \includegraphics[width=0.48\linewidth]{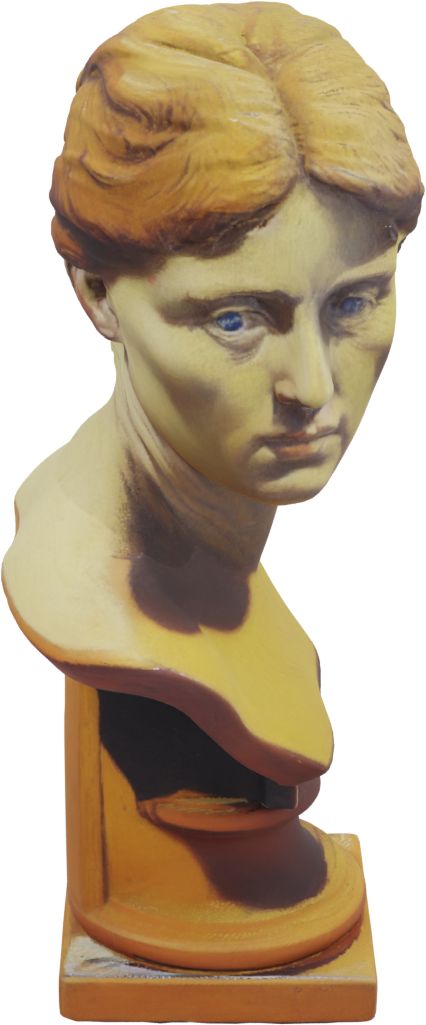}
	    \includegraphics[width=0.48\linewidth]{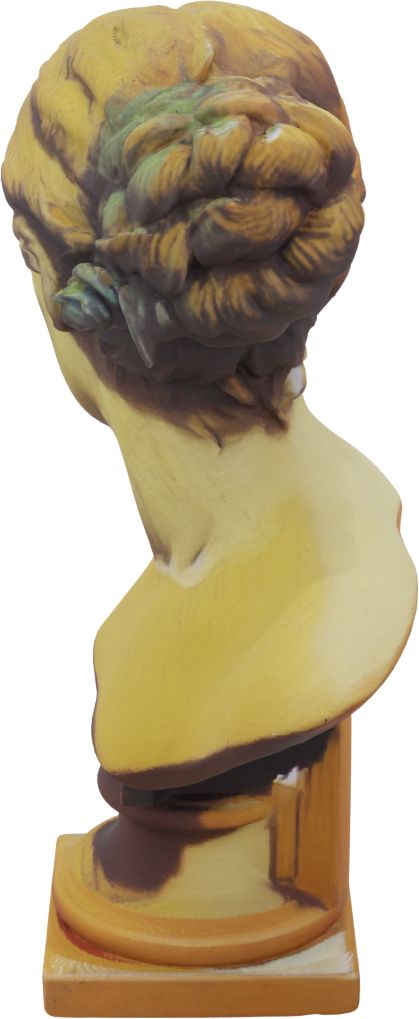}
	\end{minipage}
	\begin{minipage}{0.16\linewidth}
	    \centering
	    \includegraphics[width=0.48\linewidth]{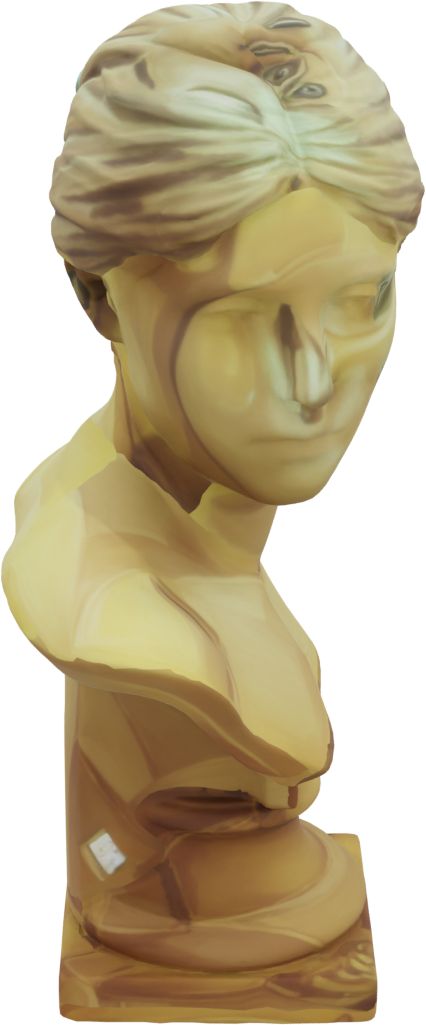}
	    \includegraphics[width=0.48\linewidth]{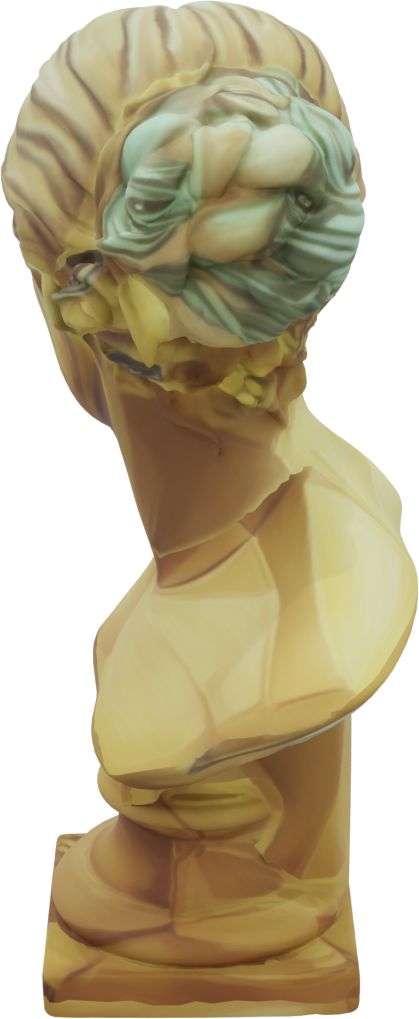}
	\end{minipage}
	\begin{minipage}{0.16\linewidth}
	    \centering
	    \includegraphics[width=0.48\linewidth]{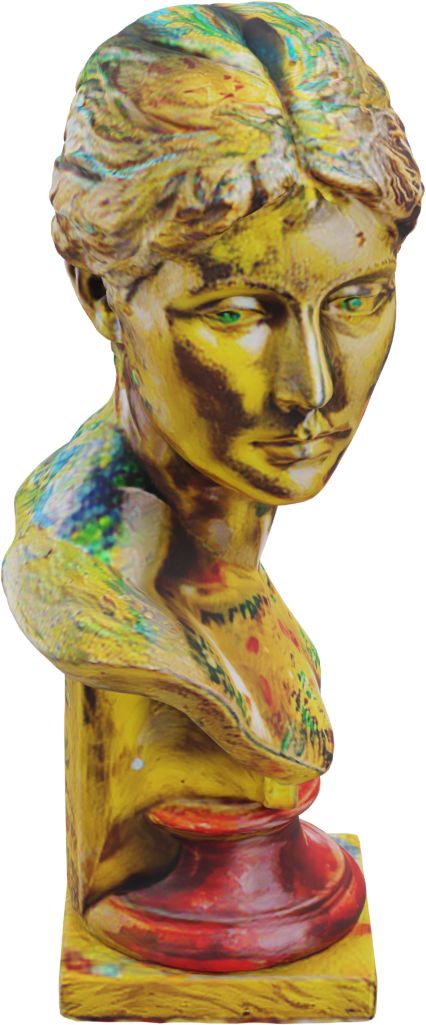}
	    \includegraphics[width=0.48\linewidth]{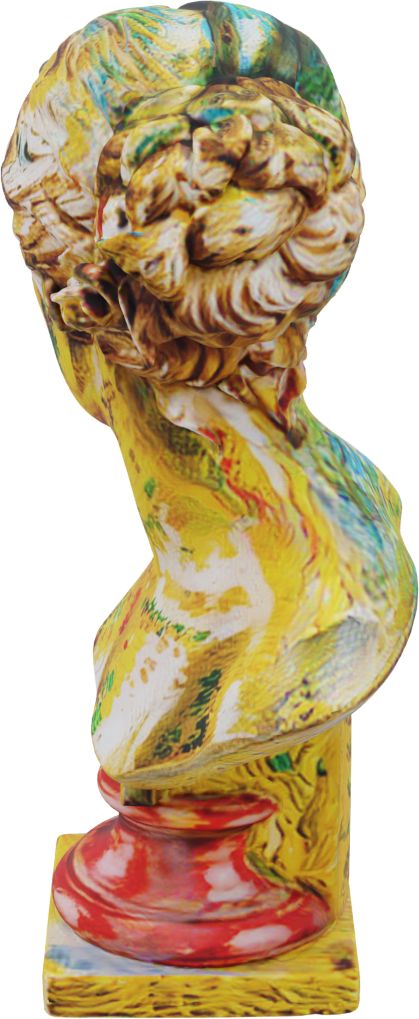}
	\end{minipage}
    \vspace{2mm}

    \Large{\emph{``a sculpture of a woman painted in the \textbf{\textcolor{vangoghblue}{style of Van Gogh}}''}}
    \vspace{2mm}

	\begin{minipage}{0.16\linewidth}
	    \centering
	    \includegraphics[width=\linewidth]{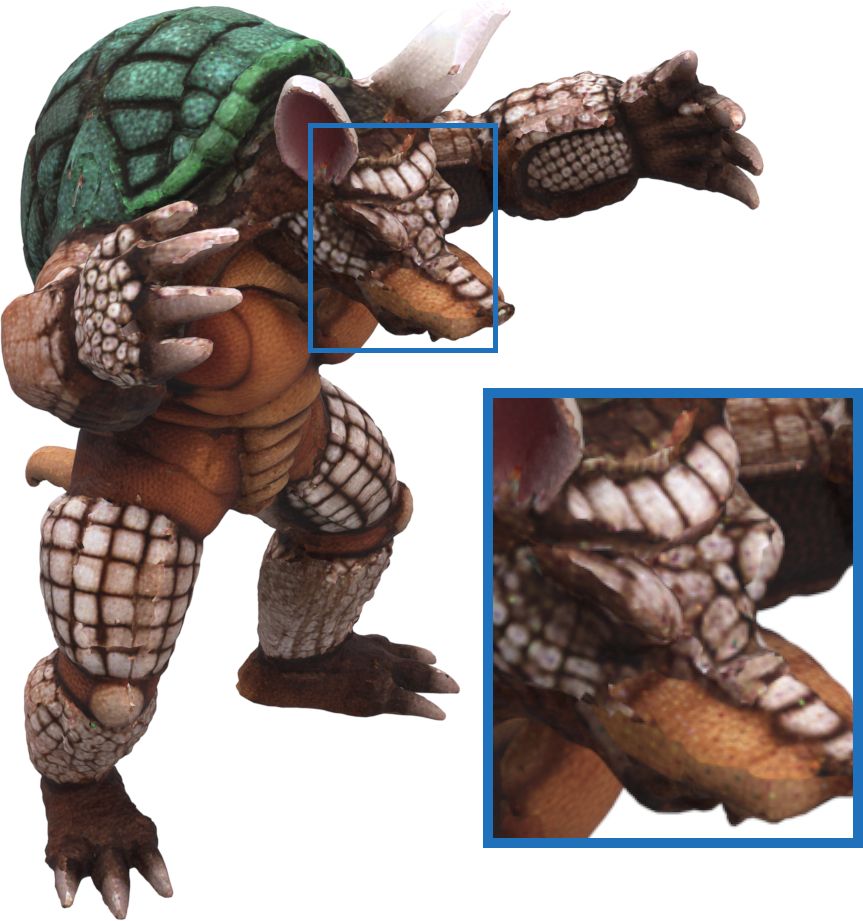}
	\end{minipage}
	\begin{minipage}{0.16\linewidth}
	    \centering
	    \includegraphics[width=\linewidth]{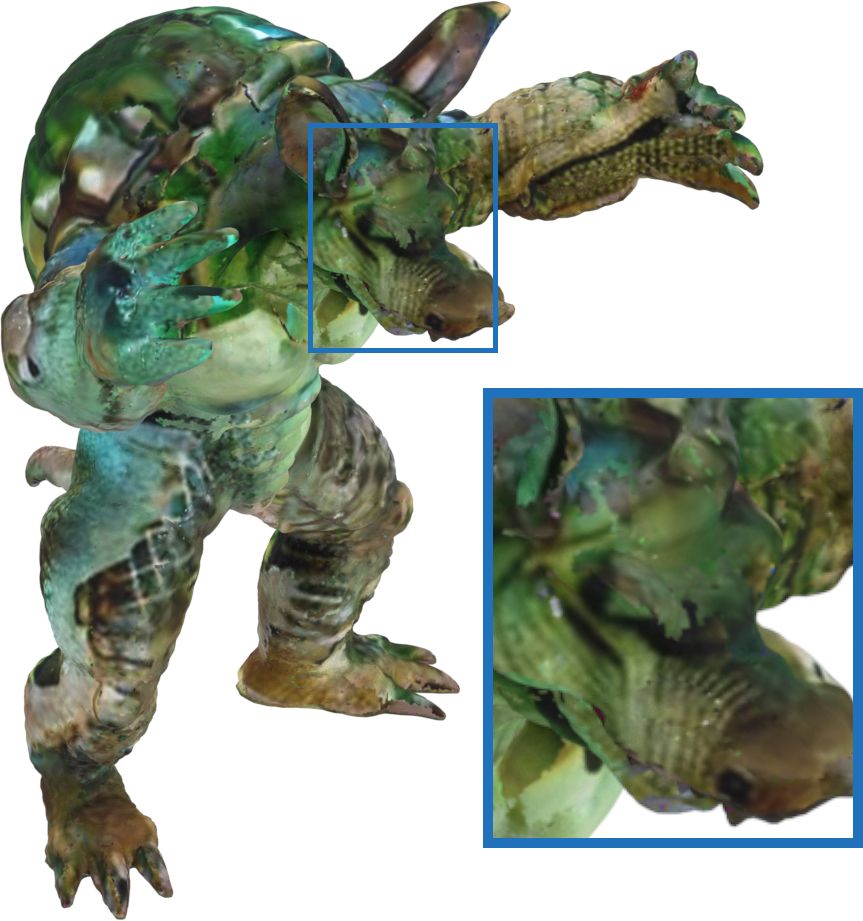}
	\end{minipage}
	\begin{minipage}{0.16\linewidth}
	    \centering
	    \includegraphics[width=\linewidth]{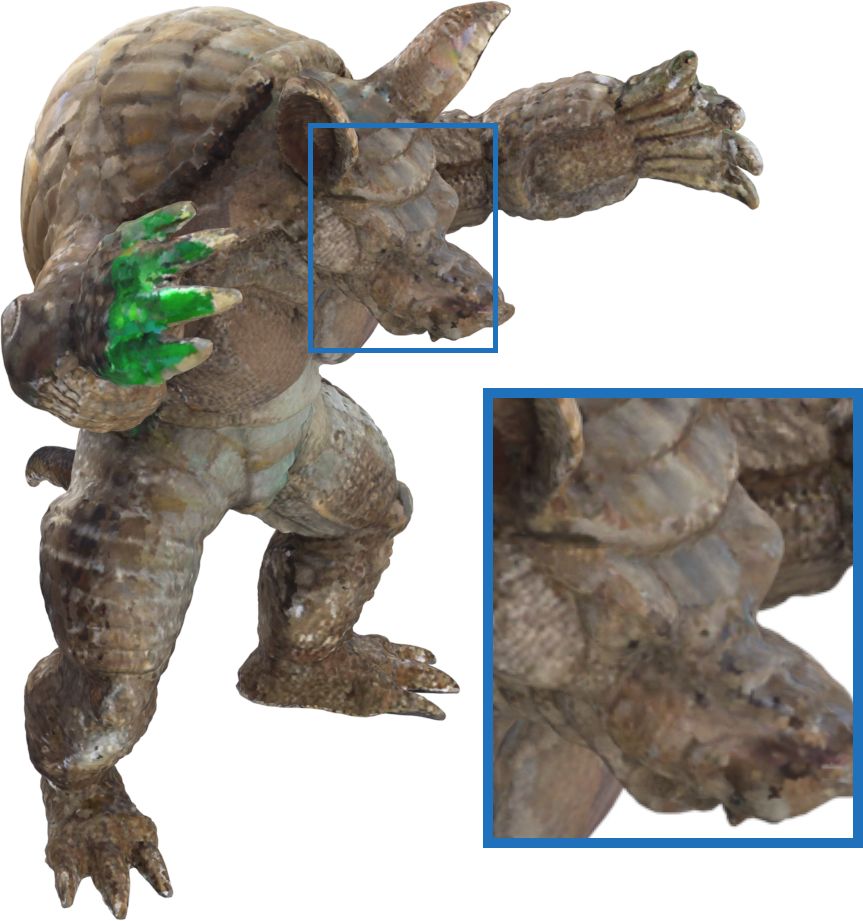}
	\end{minipage}
	\begin{minipage}{0.16\linewidth}
	    \centering
	    \includegraphics[width=\linewidth]{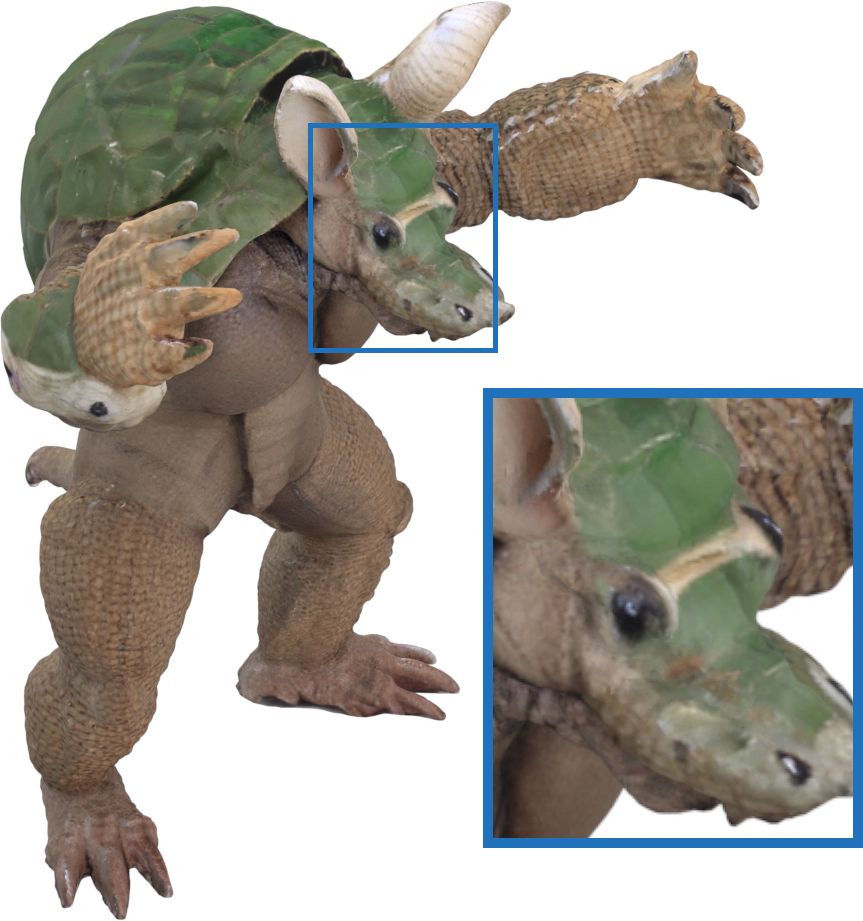}
	\end{minipage}
	\begin{minipage}{0.16\linewidth}
	    \centering
	    \includegraphics[width=\linewidth]{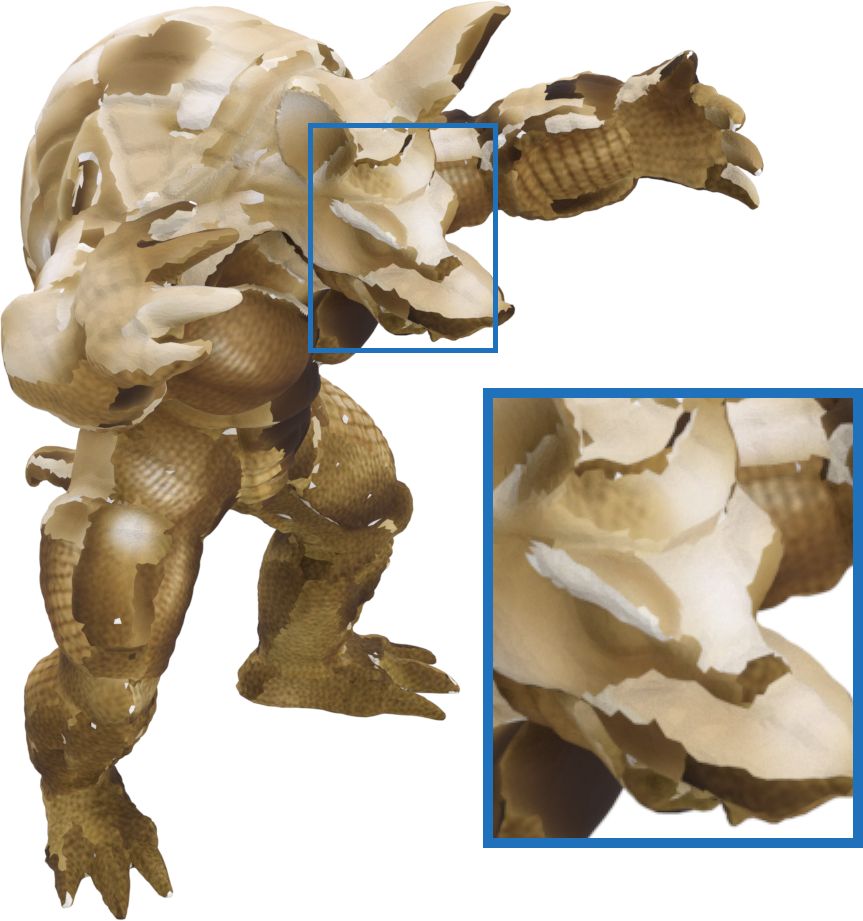}
	\end{minipage}
	\begin{minipage}{0.16\linewidth}
	    \centering
	    \includegraphics[width=\linewidth]{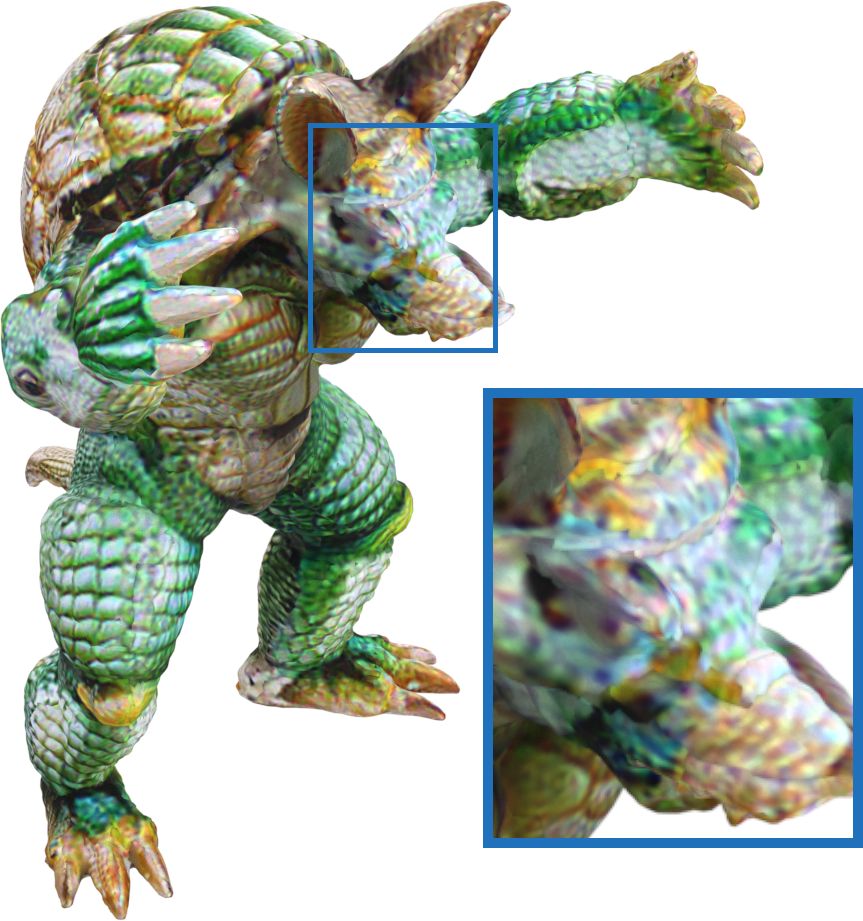}
	\end{minipage}
    \vspace{2mm}


    \Large{\emph{``a realistic armadillo creature with a \textbf{\textcolor{turtlegreen}{shell like a green turtle}} on its back''}}

\end{minipage}

%
		%
		%
%
%
\vspace{0.5cm}
\caption{\textbf{Qualitative comparison with previous work (local consistency, quality and text alignment).} Compared with previous work, our method results in higher-quality textures, while preserving local consistency and adhering to the text prompt.}
\label{fig:comparison_short}
\end{figure*}

%% file: figures/bunnies/fig_bunnies.tex
\begin{figure*}[t]
\centering

\begin{minipage}{\linewidth}
	\centering

	\begin{minipage}{0.33\linewidth}
	    \centering
	    \includegraphics[trim={0 0 0 0},clip,width=\linewidth]{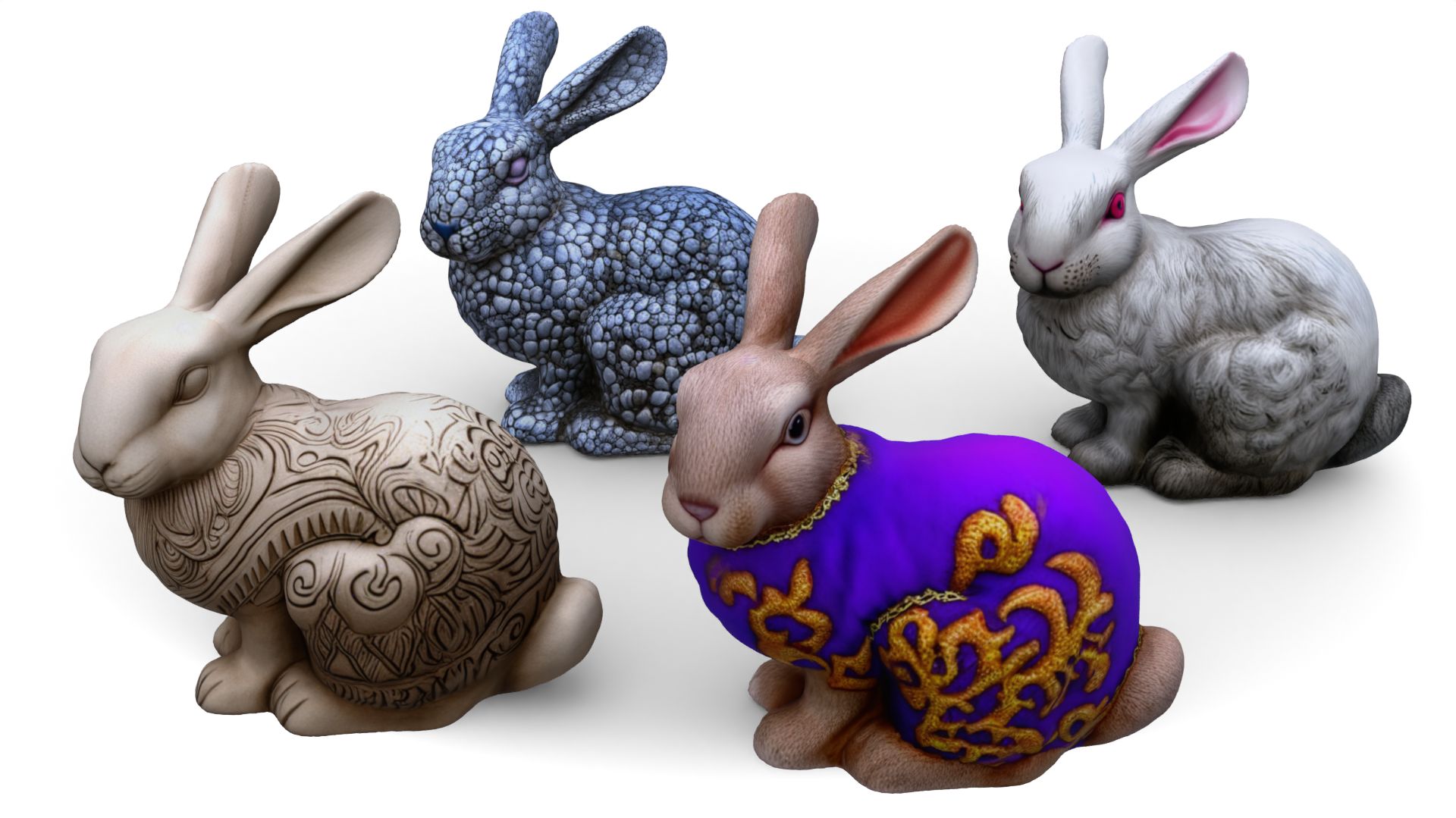}
        \textbf{Ours}
	\end{minipage}
	\begin{minipage}{0.33\linewidth}
	    \centering
	    \includegraphics[trim={0 0 0 0},clip,width=\linewidth]{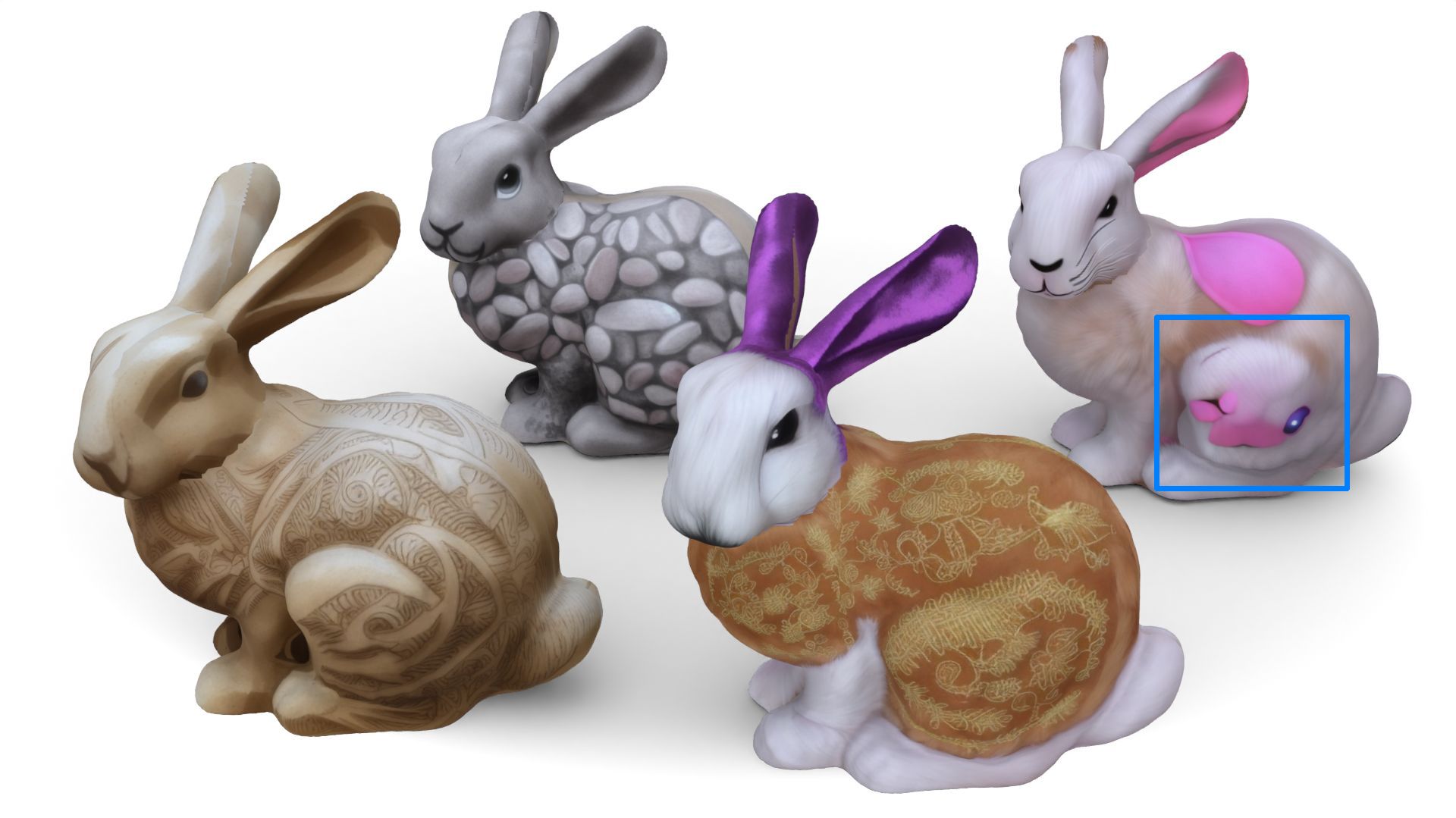}
        Paint3D
	\end{minipage}
	\begin{minipage}{0.33\linewidth}
	    \centering
	    \includegraphics[trim={0 0 0 0},clip,width=\linewidth]{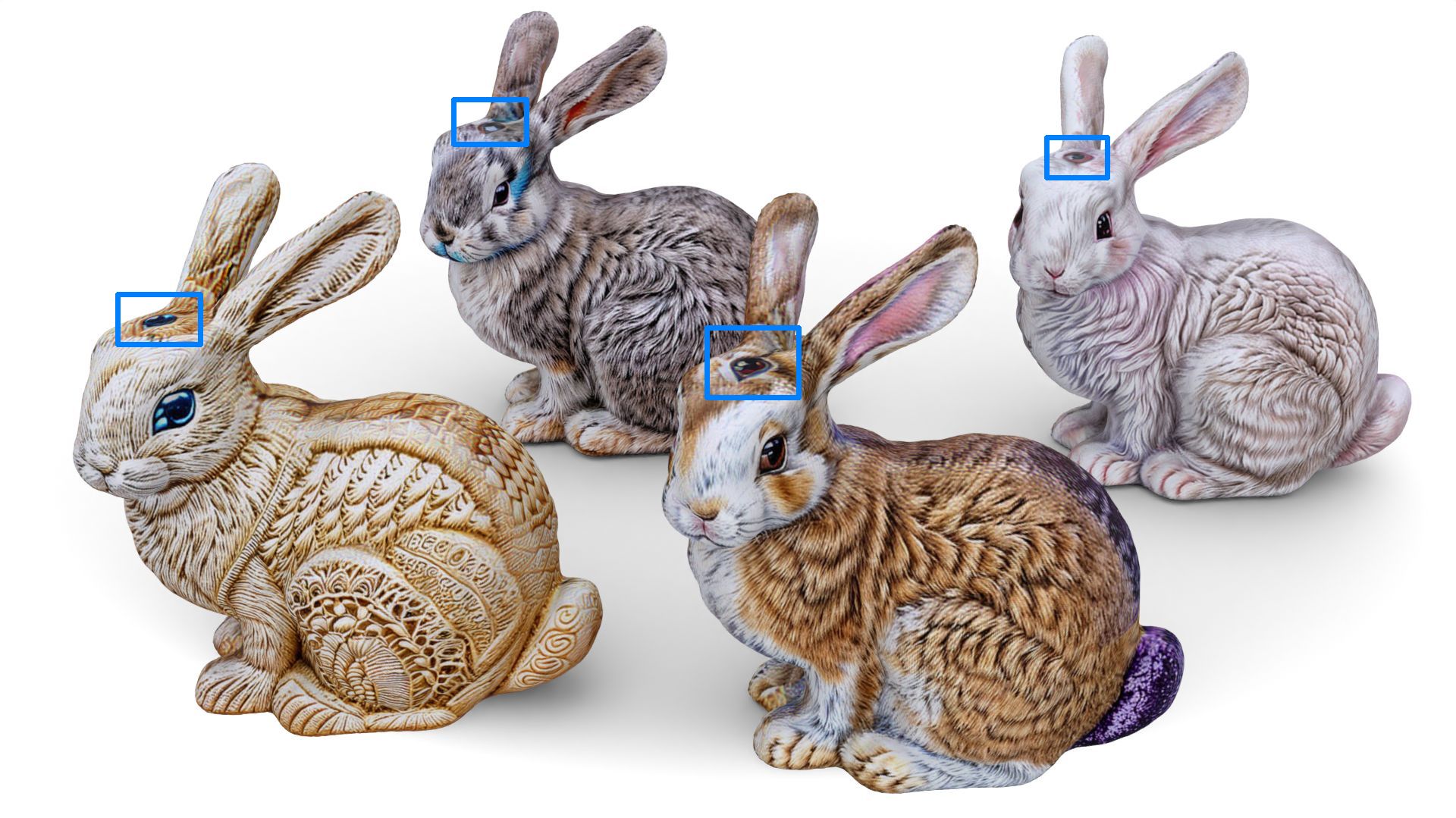}
        Meshy 3.0
	\end{minipage}
    \vspace{2mm}

	\begin{minipage}{0.33\linewidth}
	    \centering
	    \includegraphics[width=\linewidth]{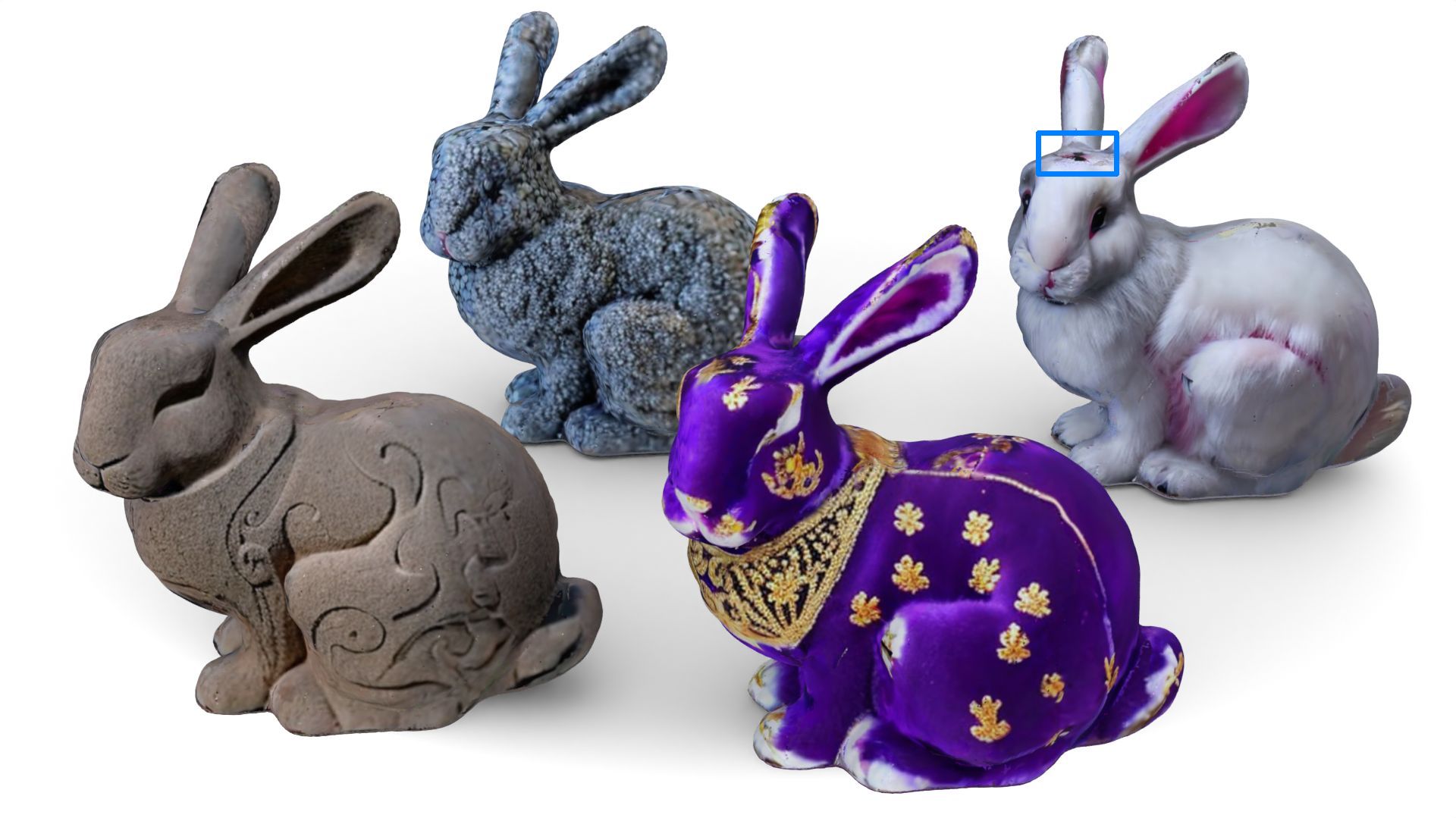}
        TEXTure
	\end{minipage}
	\begin{minipage}{0.33\linewidth}
	    \centering
	    \includegraphics[width=\linewidth]{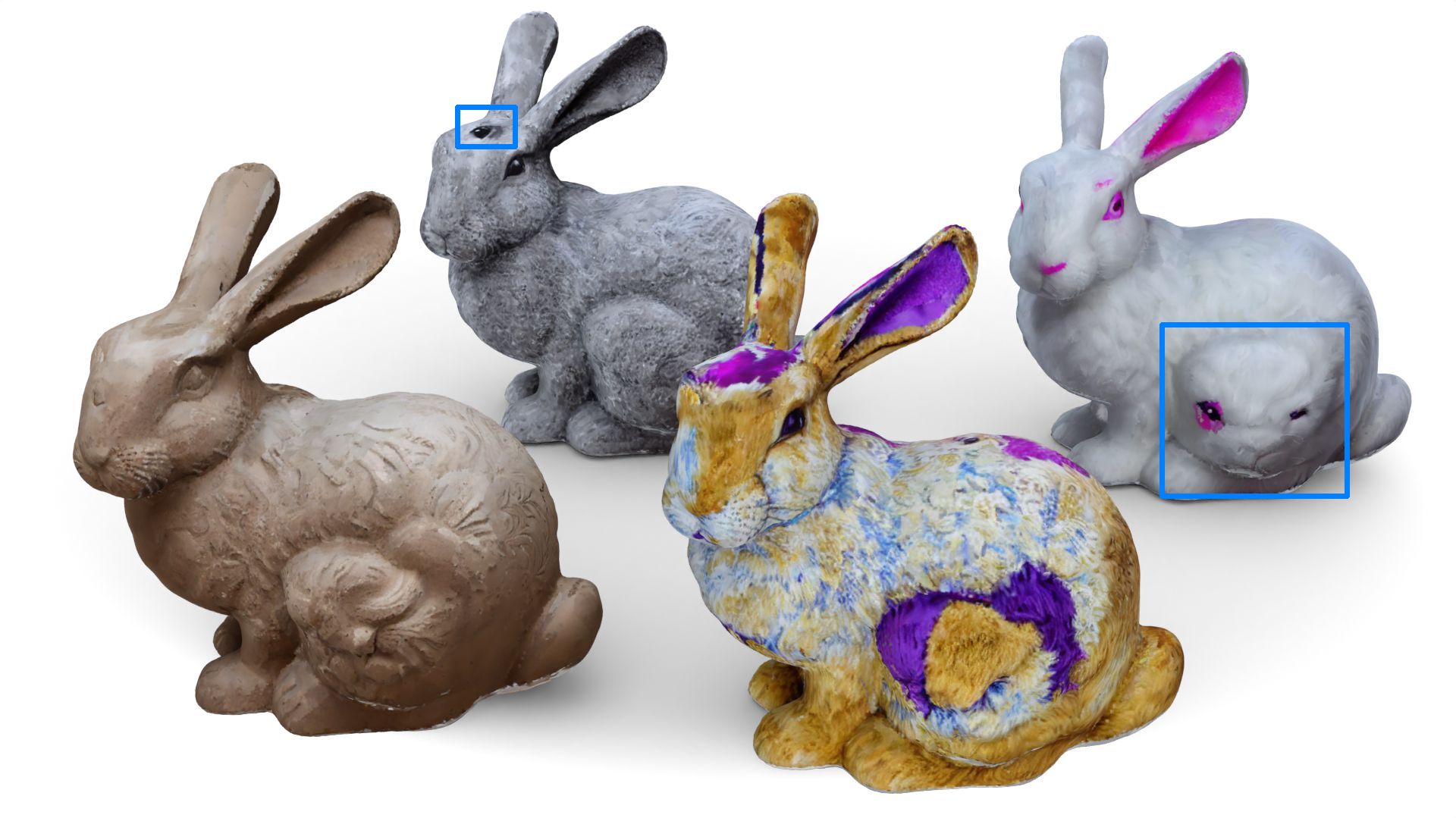}
        Text2Tex
	\end{minipage}
	\begin{minipage}{0.33\linewidth}
	    \centering
	    \includegraphics[width=\linewidth]{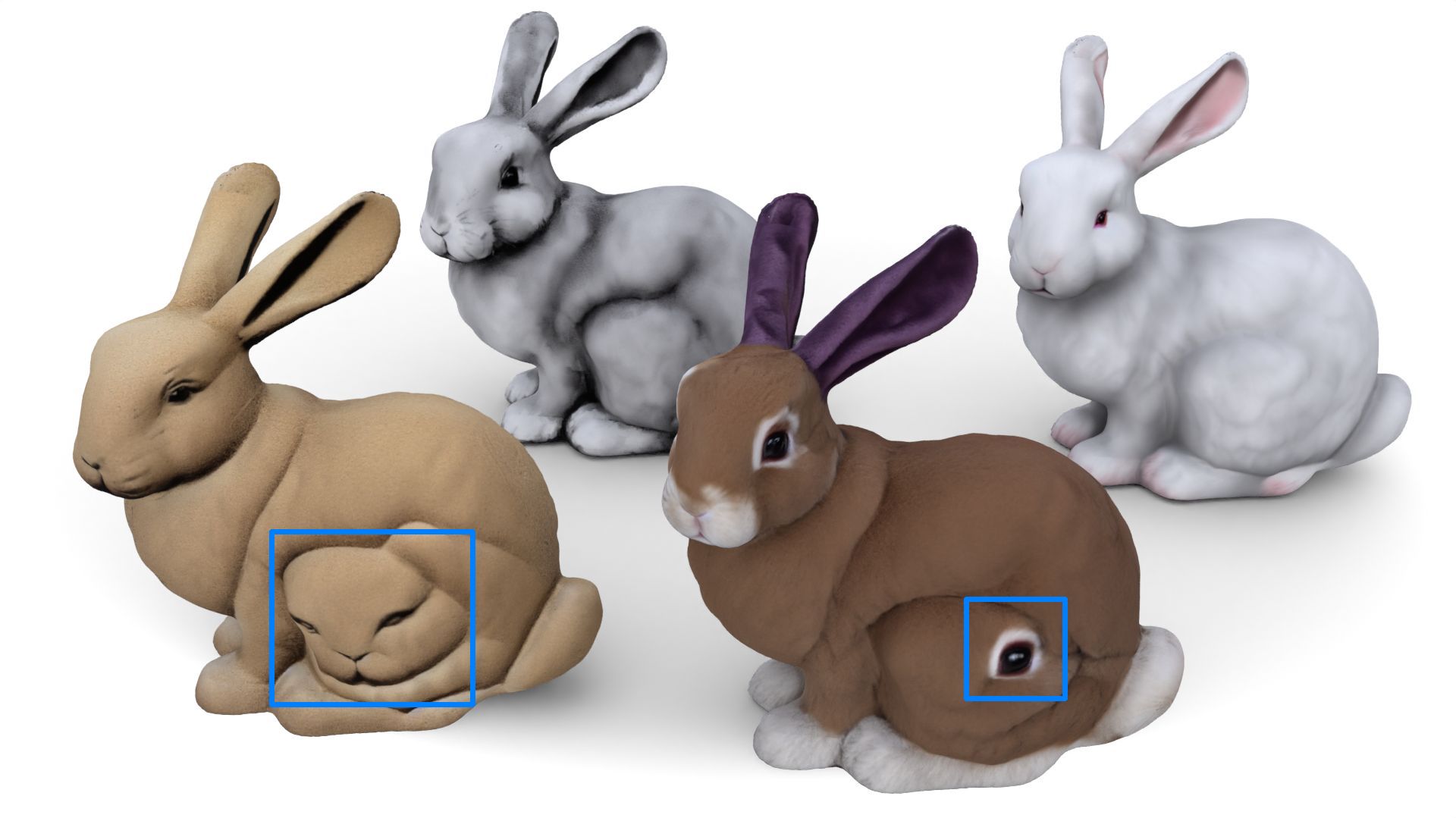}
        SyncMVD
    \end{minipage}


\end{minipage}

%
		%
		%
%
%

\caption{\textbf{Qualitative comparison with previous work (global consistency, quality and text alignment).} While previous methods result in global inconsistencies such as the Janus effect (\textcolor{blue}{blue rectangles}), as well as text mis-alignments, our method returns a globally consistent and highly text-aligned textures. Text prompts: \textbf{(i) top-left:} \emph{``a bunny made out of \textbf{small pebbles} of many \textbf{\textcolor{gray}{shades of {gray}}}''}, \textbf{(ii) top-right:} \emph{``a realistic white rabbit with long fur, \textbf{\textcolor{purple}{pink eyes}}, and \textbf{black paws}''}, \textbf{(iii) bottom-left:} \emph{``a \textbf{\textcolor{brown}{sand sculpture}} of a bunny with engraving of an intricate pattern''}, \textbf{(iv) bottom-right:} \emph{``a bunny with a \textbf{\textcolor{violet}{velvet purple coat}} with \textbf{\textcolor{yellow}{intricate gold}} embroidery along the edges''}.}
\label{fig:bunnies}
\end{figure*}

%% file: sections/4_experiments.tex
\section{Experiments}%
\label{sec:experiments}

We evaluate our method in comparison to state-of-the-art previous work, namely TEXTure~\cite{richardson2023texture}, Text2tex~\cite{chen2023text2tex}, SyncMVD~\cite{liu2023text}, Paint3D~\cite{zeng2023paint3d}, and the commercial product Meshy 3.0~\cite{meshy3p0}. 
Our method achieves state-of-the-art results according to user studies and numerical metric comparisons. Samples supporting the qualitative advantage are provided in \cref{fig:comparison_short,fig:bunnies}, while quantitative comparisons are provided in \cref{tab:freq}.
Additionally, we provide a qualitative ablation study in \cref{fig:ablation} to better assess the effects of different contributions. Diverse sets of generated samples are provided in \cref{fig:teaser,}, \cref{fig::diverse_llamas}, and \cref{fig::envs}, as well as in the appendix, including animated samples in the  \href{https://youtu.be/110Rr2ABCY8}{video}.

\begin{table}
  \caption{Quantitative comparison with previous work. We evaluate the win-rate of our method in terms of better representation of the prompt and fewer artifacts compared with previous methods, as well as FID, KID ($\times 10^{-3}$), and runtime. Overall, our textures were preferable over all baselines. The quantitative metrics show that we achieve better visual fidelity on this task of texturing artist-made assets.}
  \label{tab:freq}
  \begin{tabular}{p{1cm}ccccc}
  \toprule
       & \multicolumn{1}{p{1.4cm}}{\centering Preference}  
       & \multicolumn{1}{p{0.3cm}}{\centering Artifacts} 
       & \multicolumn{1}{p{1cm}}{\centering FID$\downarrow$}  
       &   \multicolumn{1}{p{0.8cm}}{KID$\downarrow$}
       & \multicolumn{1}{p{0.8cm}}{Runtime}  \\
    \midrule
    TEXTure & 78.5\% & 76.5\% & 91.4 & 8.4 & 90s \\
    Text2Tex & 81.9\% & 84.2\% & 92.1 & 6.9 & 287s   \\
    SyncMVD & 67.4\% & 66.7\% & 77.7 & 3.8 & 81s \\
    Paint3D & 78.9\% & 79.5\% & 86.1 & 5.2 & 66s\\
    Meshy~3.0 & 64.5\% & 68.4\% & 99.7 & 10.7 & 85s\footnotemark \\
    \textbf{Ours} & - & - & \textbf{73.0} & \textbf{3.6} & \textbf{19s} \\ 
  \bottomrule
\end{tabular}
\end{table}

\begin{figure*}[t]
  \centering
  \includegraphics[height=4.3cm]{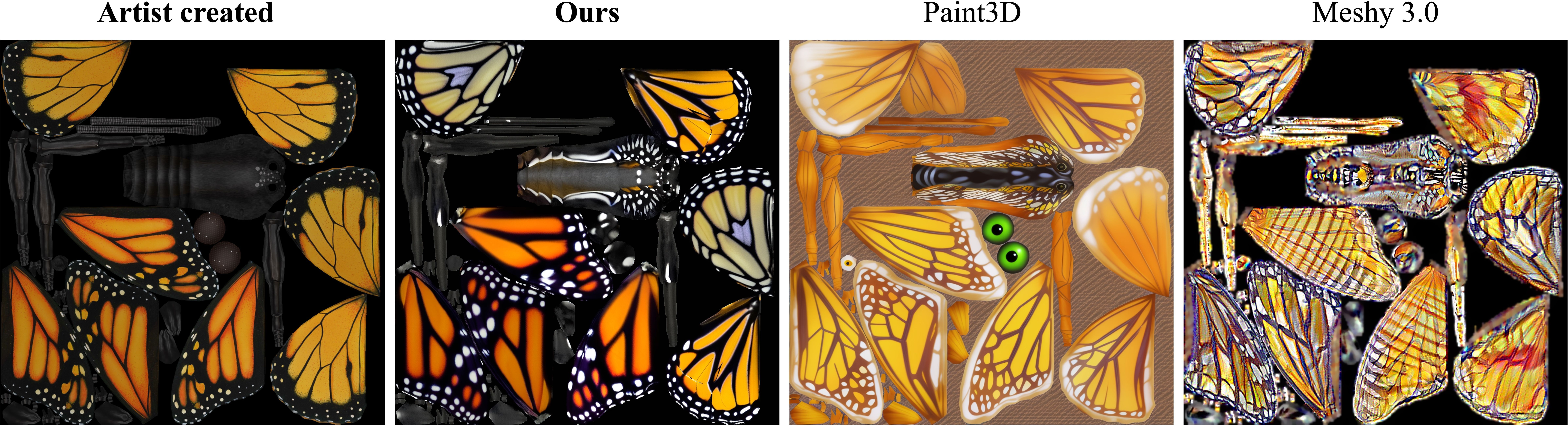}
  \includegraphics[height=4.3cm]{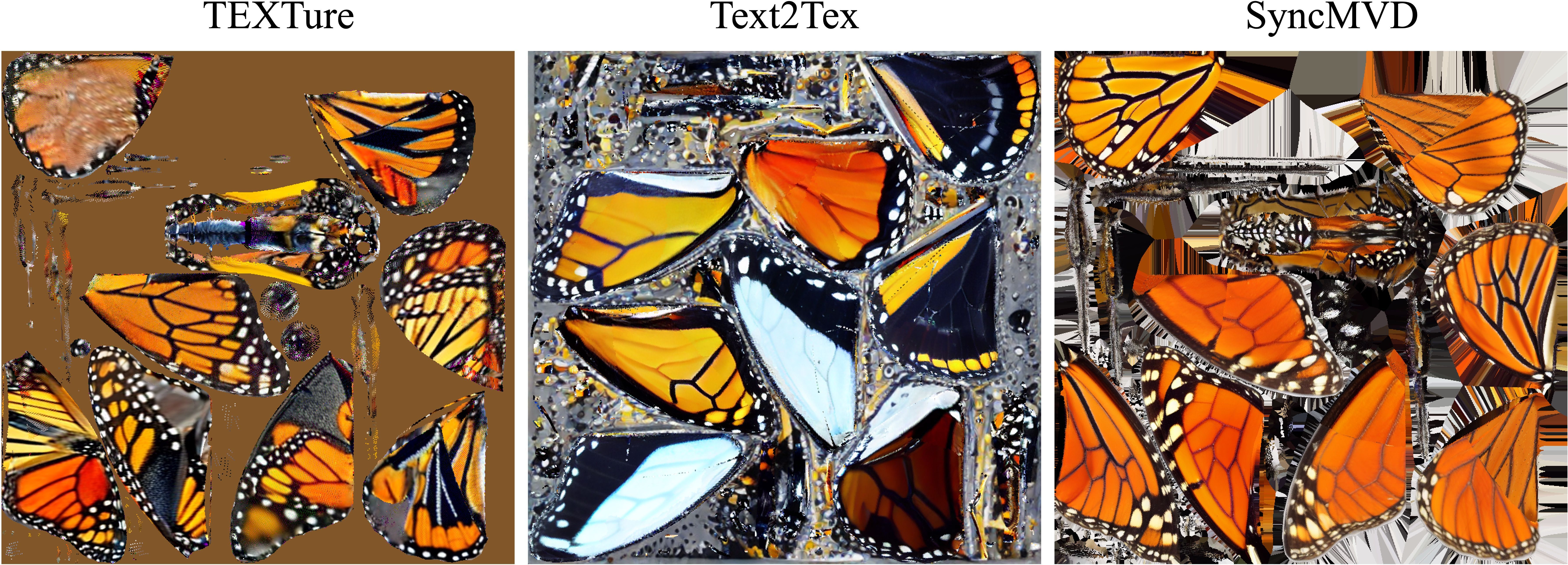}
  
  \caption{\textbf{Qualitative comparison with previous work (texture UV maps).} Our method produces a texture map which is cleaner and closer to the artist generated texture, making it more usable as part of the creative process.}
  \label{fig:artist_compare}
\end{figure*}

\subsection{Data}
\subsubsection{Training data}
Our dataset consists of $260$k textured 3D objects sourced from an in-house collection. 
Text captions are extracted for each object similarly to Cap3D~\cite{luo2024scalable}.

\subsubsection{Evaluation data}
To evaluate our methods and the baselines quantitatively and qualitatively, we use a set of $54$ objects with CC license that do not have a `No-AI' tag from the Sketchfab website. In addition, we use $2$ objects from the Stanford 3D Scanning Repository. For each $3$D object, we provide 4 creative text prompts for generations, which we use for our user study.
Additionally, we provide a single text prompt describing the original texture of each object, which is necessary for evaluation using metrics such as FID (Frechet Inception Distance)~\cite{heusel2017gans} and KID (Kernel Inception Distance)~\cite{binkowski2018demystifying}.
The complete list of objects and prompts is provided in the supplementary.
Additionally, in all qualitative and quantitative comparisons, we do not employ the texture enhancement network in order to allow for fair comparison in terms of resolution, as both our method and the baselines generate texture maps at a resolution of $1024 \times 1024$. 

\subsection{Quantitative comparisons}
In order to quantitatively evaluate our method, we employ the FID and KID metrics.
These metrics aim at evaluating the quality of the generated textures.
Furthermore, we conduct a user study to evaluate how well the generated texture represent the objects in terms of visual quality and text alignment, as well as the presence of artifacts.

\begin{figure*}[h]
  \centering
  \includegraphics[width=\linewidth]{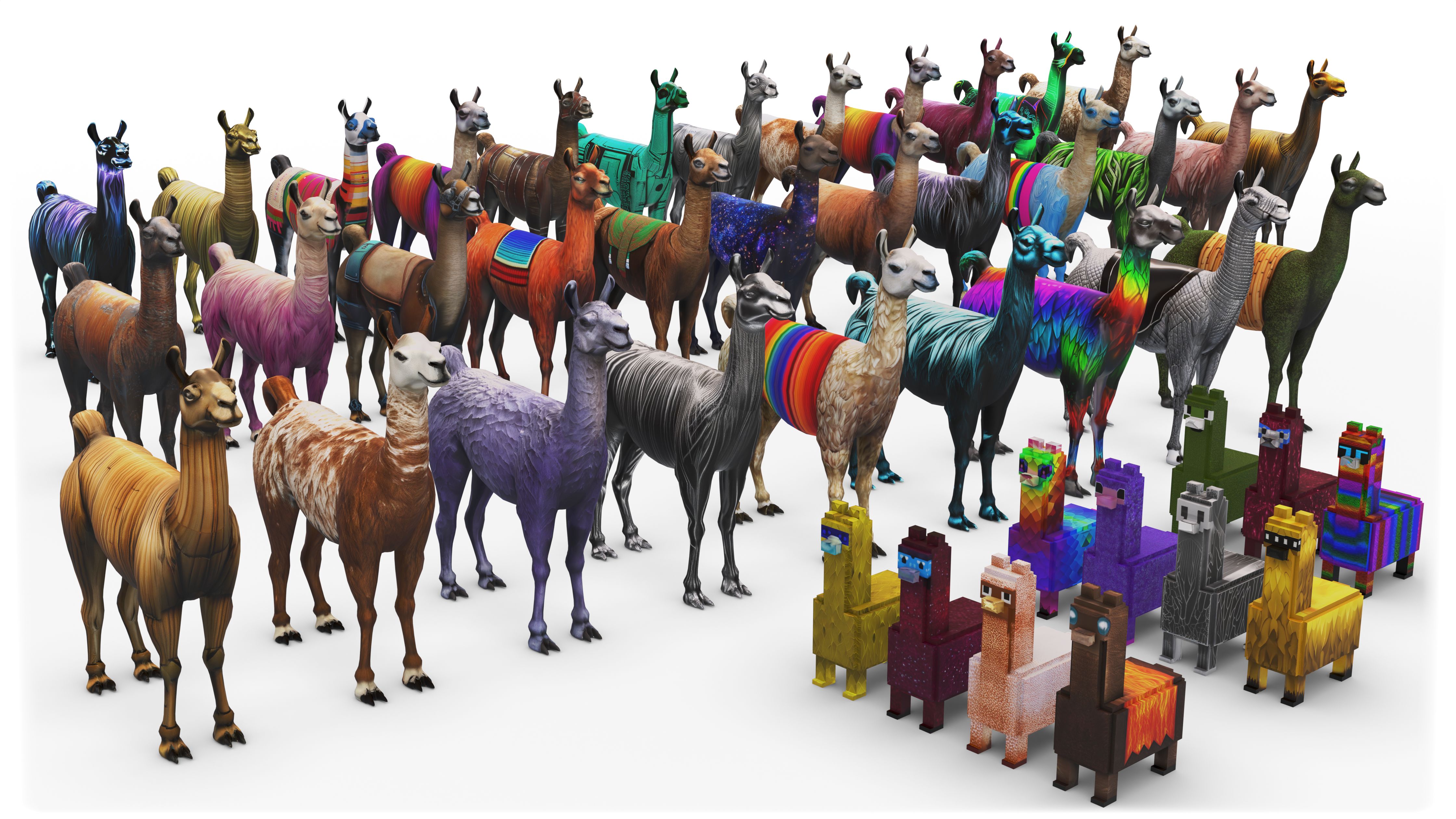}
  \caption{\textbf{Diversity of prompts.} Our method enables generating diverse prompts ranging from realistic to extremely fantastical creations. Here we show 33 different textures on the same llama model, and 11 textures for a voxel model for which the text prompts emphasize creation of low poly assets.}
  \label{fig::diverse_llamas}
\end{figure*}

\subsubsection{User study}
For the user study, we rendered $360${\textdegree} rotation videos of the generated textured meshes from our evaluation set. In each question we present two videos side-by-side, one generated by our model and another generated by one of the baselines, along with the text prompt used to generate the textures. The order of meshes, prompts and baselines are randomized, as well as the left-right ordering of the baseline and our method in order to eliminate bias. 
Similarly to~\cite{chen2023text2tex}, participants were asked to choose which object best represents the given prompt. This question captures both text alignment and overall visual quality, as textures of low quality do not represent the desired object well.
In addition, we ask which object displays fewer visual artifacts to capture cases in which an object is generally of better quality, e.g., more detailed or realistic, but includes some errors or inconsistencies.
The decision for each texture is determined by max-voting. An example question screenshot is provided in the supplementary. $33$ users participated in the study, with $754$ responses.
A breakdown comparing our method with the baselines (see \cref{sec:texture-generation}) can be seen in \cref{tab:freq}.
Overall, our method was preferred over all baselines, both in terms of overall quality and when considering artifacts.

\subsubsection{Metrics}
For the FID and KID calculations, we render the ground-truth textured meshes and the generated textured meshes from $32$ evenly spaced viewpoints under identical conditions, the standard image FID and KID scores are then calculated between these two sets of rendered images.
For runtime, we compare inference time for each method, where we define inference time as the time it takes to generate a complete texture map for a given text prompt and pre-defined mesh. Even though we report faster runtime for the baselines compared with the numbers reported in the original papers, we emphasize that we could not run them in the same exact setup as ours (single H100 vs. A100 GPU), which should translate to some reduction in runtime. However, given our method's advantage of not running multiple generation iterations, combined with the significant runtime difference, we expect that our method would be the fastest when running on the same GPU.

\def\thefootnote{*}\footnotetext{Runtime for Meshy is estimated using the Meshy 3.0 API.}

\subsection{Qualitative comparisons}
We provide several qualitative comparisons with previous work, focusing on different aspects of visual quality: text fidelity, global consistency, local consistency, and texture map usability.
\Cref{fig:comparison_short} emphasizes the challenge in adhering to the text prompt while generating visually pleasing and geometrically coherent textures (e.g. style of Van Gogh, specifically texturing the shell as green, as well as fine-details of the armadillo's face).
\Cref{fig:bunnies} focuses on global consistency, where the Janus effect can be seen clearly in all baselines as additional sets of eyes or faces, as well as text fidelity, where previous methods struggle in maintaining alignment. Finally, UV texture maps generated by the different methods are illustrated to assess the potential usability of these maps, while comparing them with the original artist-created map in \cref{fig:artist_compare}.

\subsection{Ablation study}
In order to assess the importance of different contributions to our method, we provide an ablation study in \cref{fig:ablation}. We compare five cases: (a) excluding the first stage (no image space), (b) excluding the second stage (no UV space), (c) excluding the weighted-incidence blending (simple averaging), (d) our method without texture enhancement (SR), and (e) our method with texture enhancement. 

\input{figures/ablation/fig_ablation}

\textbf{Omitting stage I} (no image space).
In this scenario we fine-tuned a diffusion model that operates in UV space exclusively, similarly to the second stage. However, we omit the partial texture and inpainting mask conditioning and provide only position and normal UV maps ($\text{P}_\text{UV}$,$\text{N}_\text{UV}$) as visual conditions. We additionally enable text conditioning for guidance. This setup proved to be challenging for a standard diffusion model, as it struggled to capture the 3D semantics presented in the exclusive form of UV maps. This resulted in generated textures that exhibit text-alignment issues, especially for non-global prompts, as well as significant local consistency issues (``seam'') appearing at the boundaries of UV fragments.

\textbf{Omitting stage II} (no UV space).
Next, we directly evaluate the generated output of stage I after backprojection. In most cases four views are insufficient to cover an entire 3D object, resulting in several ``unpainted'' areas. Furthermore, we observed that the quality of the backprojected texture is inferior to that of the full method. This suggests that our UV-space stage not only inpaints the occluded areas, but also refines the existing areas of the partial texture and mitigates backprojection artifacts, thereby enhancing its quality and effective resolution.

\textbf{Average blending.}
Lastly, we evaluate a straightforward averaging approach to merge the generated views, as opposed to using a weighted incidence-based blending technique. This results in a final output with blurry areas while lacking fine details.

%% file: figures/ablation/fig_ablation.tex
\begin{figure*}[t]
\centering

\begin{minipage}{\linewidth}
	\centering

	\begin{minipage}{0.19\linewidth}
	    \centering
	    \includegraphics[width=\linewidth]{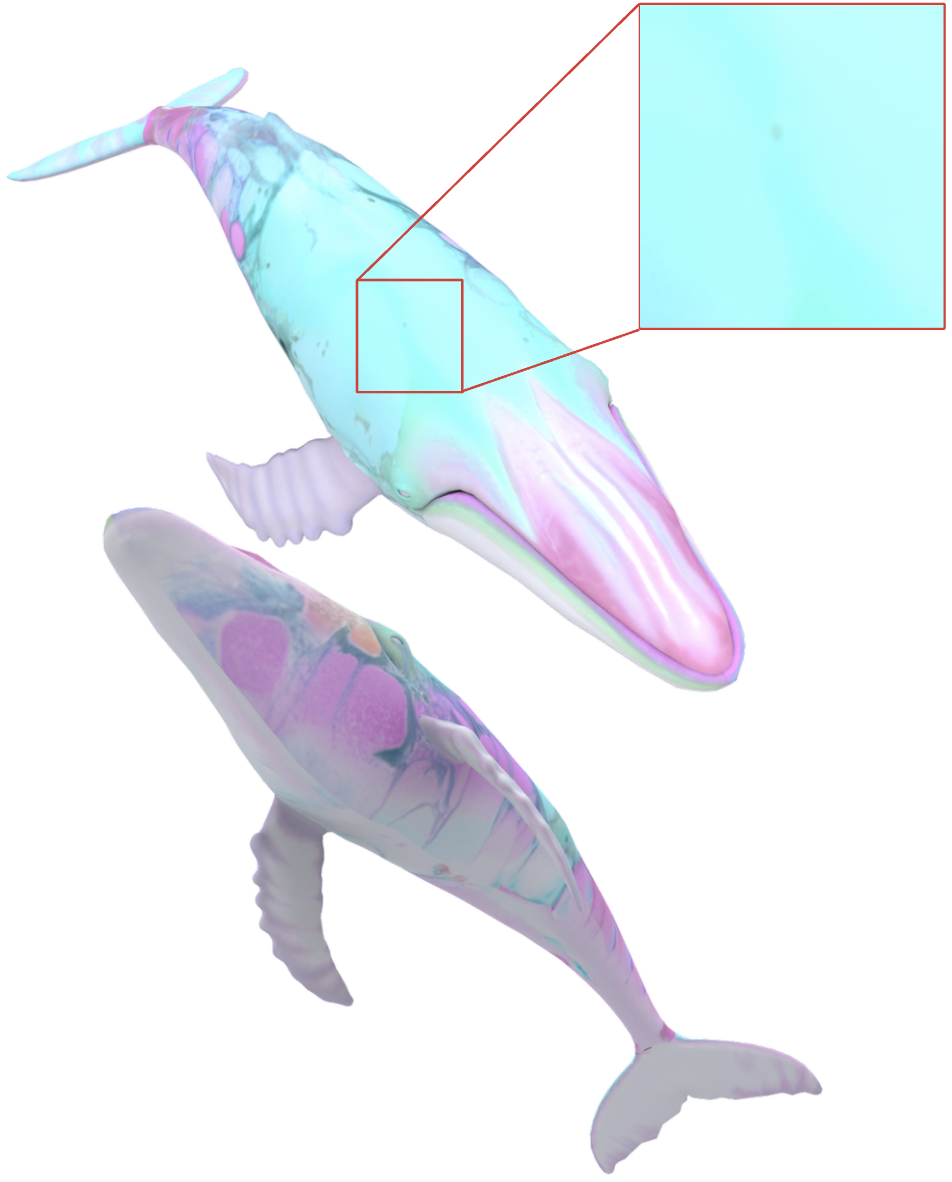}
	\end{minipage}
	\begin{minipage}{0.19\linewidth}
	    \centering
	    \includegraphics[width=\linewidth]{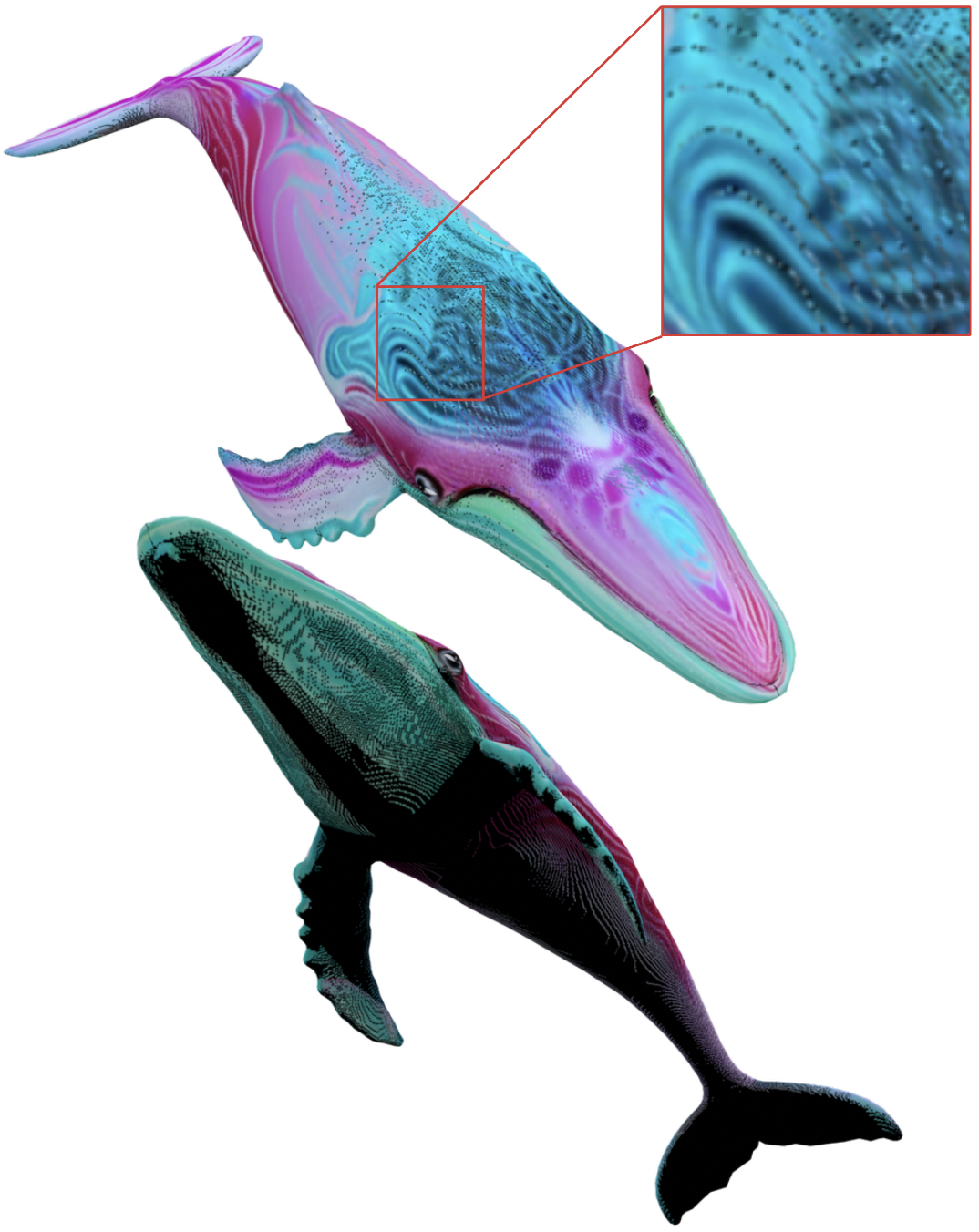}
	\end{minipage}
	\begin{minipage}{0.19\linewidth}
	    \centering
	    \includegraphics[width=\linewidth]{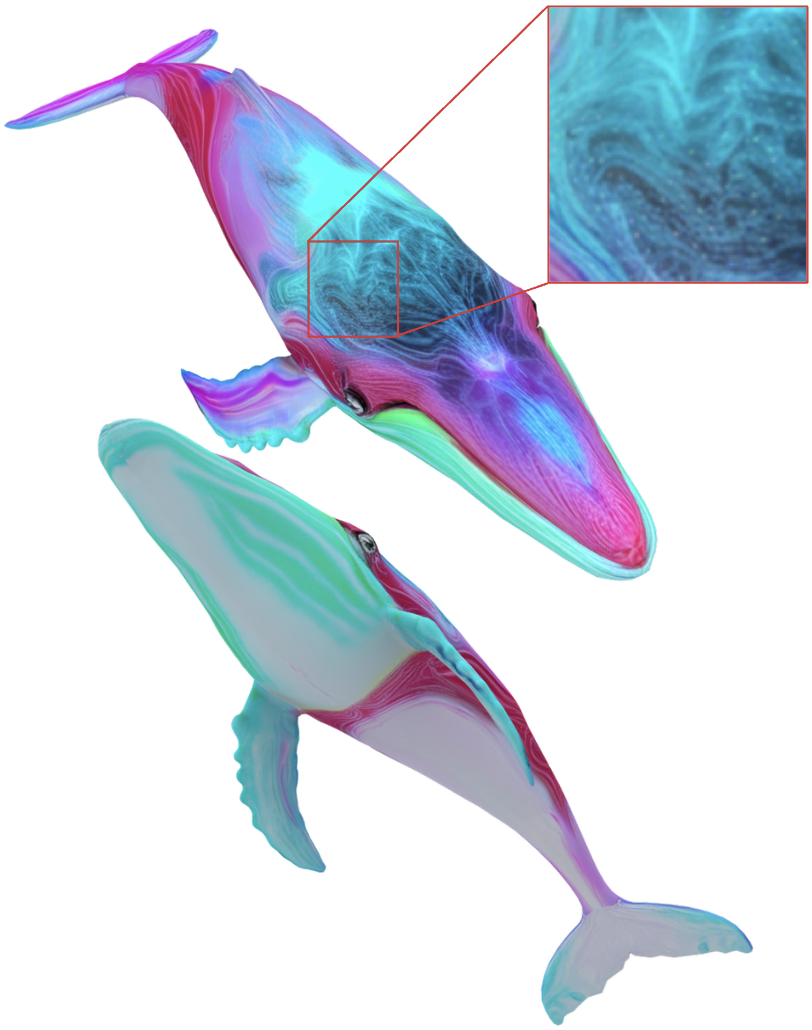}
	\end{minipage}
	\begin{minipage}{0.19\linewidth}
	    \centering
	    \includegraphics[width=\linewidth]{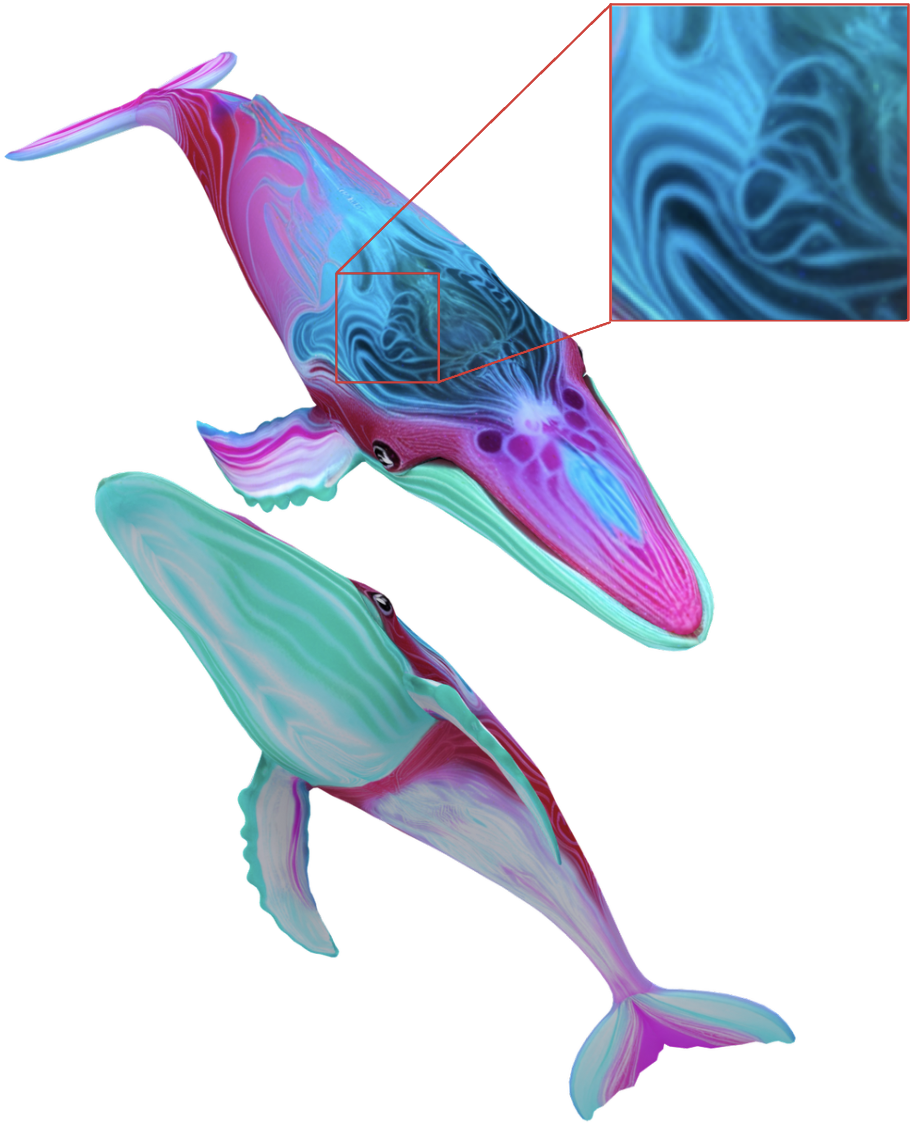}
	\end{minipage}
	\begin{minipage}{0.19\linewidth}
	    \centering
	    \includegraphics[width=\linewidth]{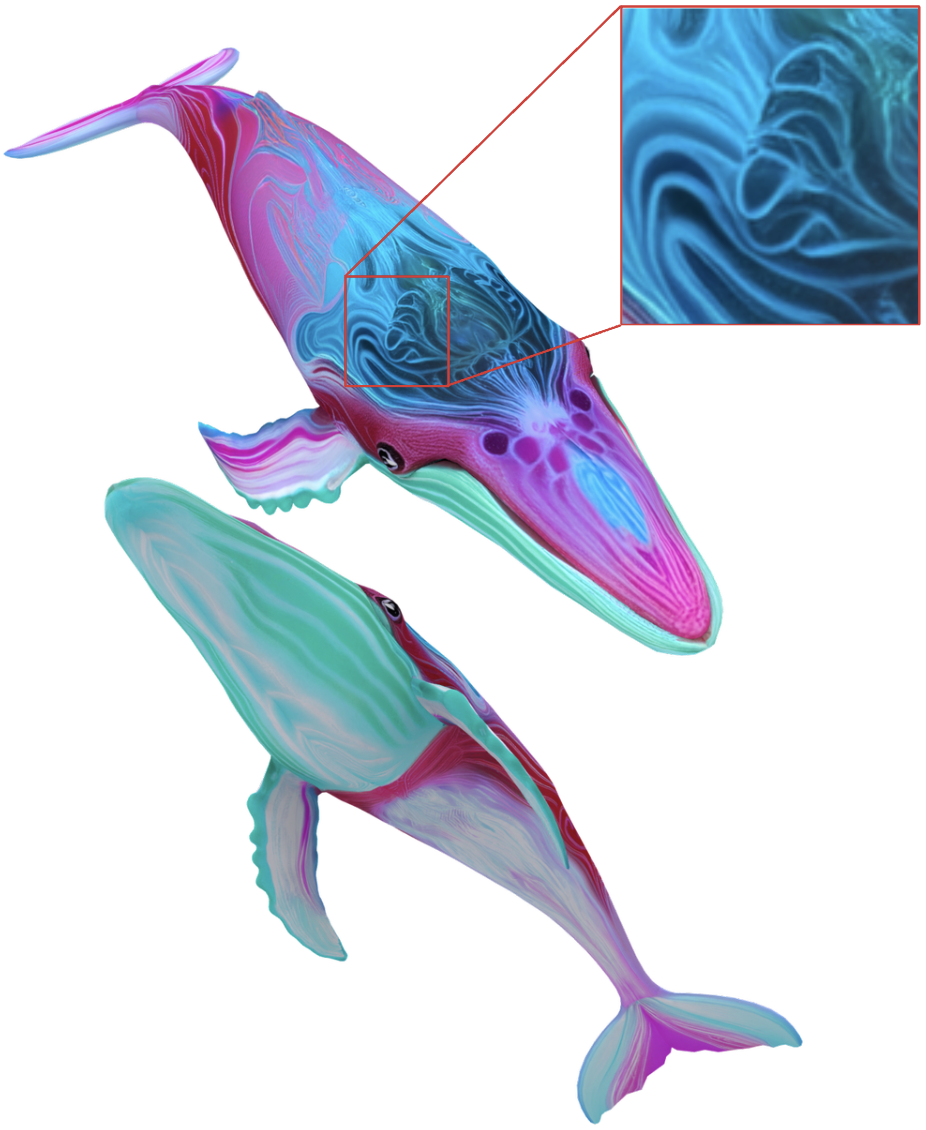}
	\end{minipage}

	\begin{minipage}{0.19\linewidth}
	    \centering
        (a) w/o stage I
	\end{minipage}
	\begin{minipage}{0.19\linewidth}
	    \centering
        (b) w/o stage II
	\end{minipage}
	\begin{minipage}{0.19\linewidth}
	    \centering
        (c) Mean blending
	\end{minipage}
	\begin{minipage}{0.19\linewidth}
	    \centering
        (d) \textbf{Ours}
	\end{minipage}
	\begin{minipage}{0.19\linewidth}
	    \centering
        (e) \textbf{Ours + SR}
	\end{minipage}

\end{minipage}

\caption{\textbf{Qualitative ablation results} for the text prompt \emph{``A whale with a pastel pink skin with swirls of mint green, lavender and blue creating a marbled effect''}. Five scenarios are evaluated: \textbf{(a)} omitting stage I (no image space), \textbf{(b)} omitting stage II (no UV space), \textbf{(c)} backprojection average blending, \textbf{(d)} our result, and \textbf{(e)} our result with the texture enhancement network.}
\label{fig:ablation}
\end{figure*}

%% file: sections/5_limitations.tex
\section{Limitations}
\label{sec:limitations}

The generation of PBR material maps, such as tangent normals, metallic and roughness are not covered by this method, and are left as future work. Although \method is currently the fastest method for texture generation, it is not real-time nor fast enough to cover all possible applications. However, the introduction of recent methods of speeding-up text-to-image models, such as Imagine-Flash~\cite{kohler2024imagine}, could directly translate into real-time texture generation, given that the bottlenecks are the text-to-image forward passes. While training on 3D datasets is crucial for achieving global consistency, the reliance on 3D datasets is somewhat limiting for training large models compared with the size of image and video datasets.

%% file: sections/6_conclusions.tex
\section{Ethical considerations}
The application of generative methods in general extends to a wide range of use cases, many of which are not covered in this work. Before implementing these methods in real-world scenarios, it is crucial to thoroughly examine the data, model, its potential uses, as well as considerations of safety, risk, bias, and societal impact. In the specific case of texture generation, the limitations of the existing shape provide some risk mitigation, as users would be bound to a pre-defined structure.

\section{Conclusions}%
\label{sec:conclusions}
We introduce \method, a new method for texturing 3D objects from text descriptions. While there has been impressive progress in this domain, our method brings texture generation to be significantly closer to an applicable tool for 3D artists and general users to create diverse textures for assets in gaming, animation and VR/MR. This is done by providing global consistency (e.g. eliminating the Janus problem), strong control (adherence to text prompts), speed, and high-resolution ($4$k) to the generation process.

\section{Acknowledgements}
\label{sec:ack}
We are grateful for the instrumental support of the multiple collaborators at Meta who helped us in this work. Emilien Garreau, Ali Thabet, Albert Pumarola, Markos Georgopoulos, Jonas Kohler, Filippos Kokkinos, Yawar Siddiqui, Uriel Singer, Lior Yariv, Amit Zohar, Yaron Lipman, Itai Gat, Ishan Misra, Mannat Singh, Zijian He, Jialiang Wang, Roshan Sumbaly. 
We thank Ahmad Al-Dahle and Manohar Paluri for their support.

%% file: appendix.tex
\appendix

\label{s:appendix}

\input{figures/diversity/fig_diversity_bunny}

\begin{figure*}[h]
  \centering
  \includegraphics[width=\linewidth]{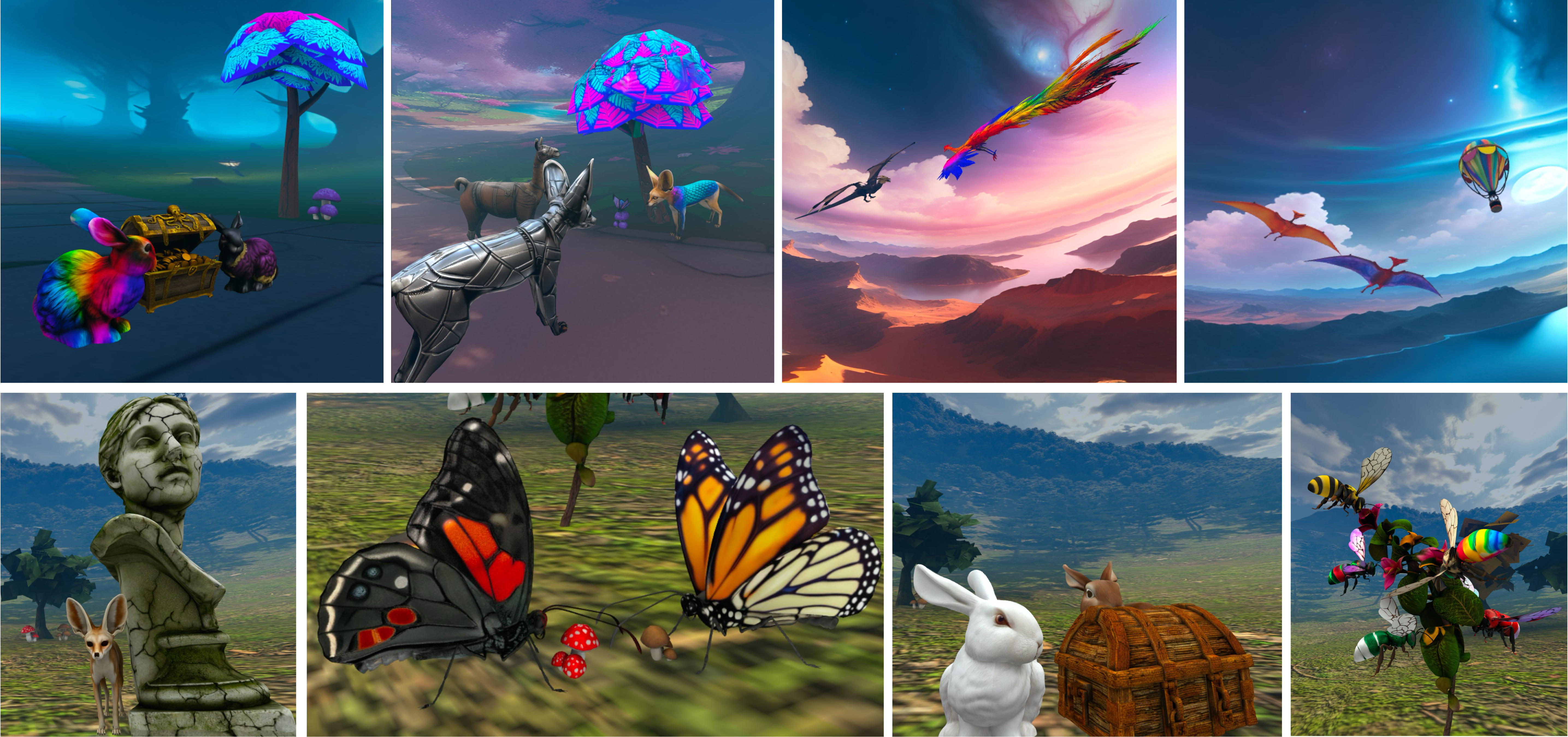}
  \caption{Generated textures in realistic and stylized VR environments. Excluding the skybox (background), all textures are generated.}
  \label{fig::envs}
\end{figure*}

\section{Additional implementation details}

\subsection{Training details}

All of the models presented in the manuscript have a similar architecture, and are fine-tuned from the same base text-to-image generation model that operates at a resolution of $1024\times1024$. Their multiple conditionings are encoded via the original image encoder matching to our base model and are concatenated altogether via channel-wise concatenation. To adapt the architecture to these new inputs, we simply add the relevant number of additional channels as zero-weighted input channels for the first convolution layer. The text-to-multiview network (Stage I) was fine-tuned to minimize the L$2$ loss, and both the UV space inpainting (Stage II) and texture enhancement networks to minimize the L$1$ loss. We use v-prediction formulation where the noise schedule was rescaled to enforce zero terminal SNR~\cite{lin2024common}.  We empirically found that the latter is beneficial when training diffusion models on renderings and UV maps, which possess large background areas, such as rendering background and unmapped pixels for UV maps.
\\
We fine-tune all of our models with a learning rate of 1e-5 and a batch size of 256 on 32 H100 gpus.
Stage I and Stage II models were fine-tuned for 15k steps each and the texture enhancement model was trained for 28k steps.
For stage I and stage II we employ DDPM solver and use 60 diffusion steps for inference. For the multi-diffusion texture enhancement we employ DDIM solver with 50 diffusion steps.

\subsection{Texture enhancement model training pipeline}

Our training pipeline for the diffusion model enhances image quality by addressing artifacts and upscaling the texture by an arbitrary ratio. 
The design of our upsampler draws inspiration from the widely utilized open-source Real-ESRGAN framework \cite{wang2021realesrgan}. Despite its effectiveness, Real-ESRGAN's degradations often produce artifacts such as over-smoothed textures, excessively sharpened edges, and patterns with high contrast, leading to noticeable ringing effects. We have noticed that our method does not exhibit these issues.
Besides changing the architecture to a diffusion model and training on high quality texture maps, we modified the data degradation pipeline to empirically better match our needs, omitting the Unsharp Masking operation as well as the additive Gaussian noise.  Our patch-based approach, followed by Multi-Diffusion blending, allows us to upsample an image by an arbitrary ratio, without introducing seams or noticeable artifacts between the patches. Moreover, we employ a tiled-VAE approach in order to overcome memory issues arising from encoding and decoding large images and latent maps. These choices resulted in a robust upsampler, tailor-made for upsampling texture maps to a very high resolution.

\begin{figure*}[h]
  \centering
  \begin{minipage}{.45\linewidth}
    \centering
    \includegraphics[width=\linewidth]{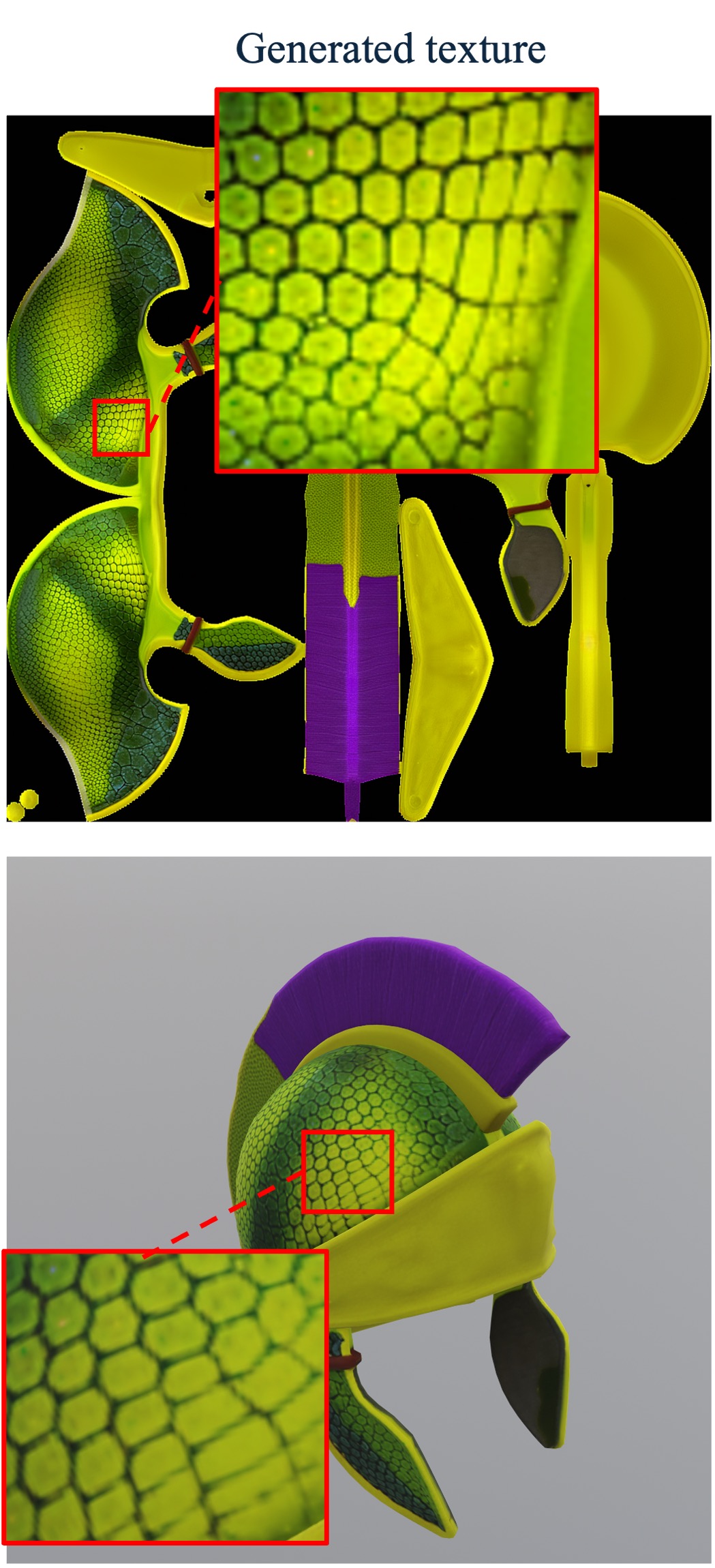}
    \caption*{(a)}
  \end{minipage}\hfill
  \begin{minipage}{.45\linewidth}
    \centering
    \includegraphics[width=\linewidth]{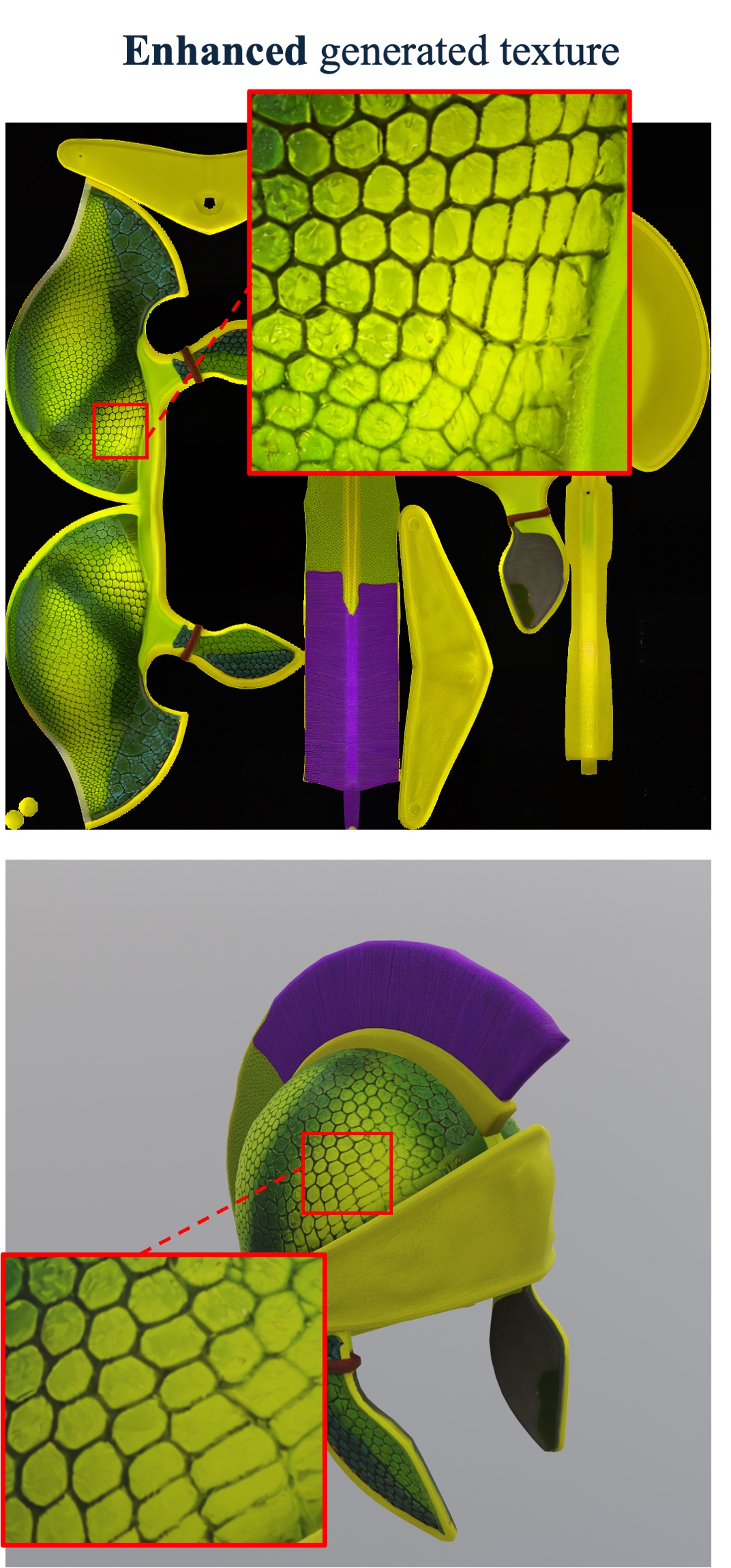}
    \caption*{(b)}
  \end{minipage}
\caption{\textbf{Enhancing textures in UV space and 3D.} (a) Generated textures and (b) enhanced generated textures for the text prompt: ``a yellow-green helmet made of snakeskin with a purple ruffle on top''. The top image represents a texture UV map, while the bottom image showcases a 3D render of the same patch from the UV space.} 
  \label{fig::texture_enhancement}
\end{figure*}

\begin{figure*}[h]
  \centering
  \begin{minipage}{.45\linewidth}
    \centering
    \includegraphics[width=\linewidth]{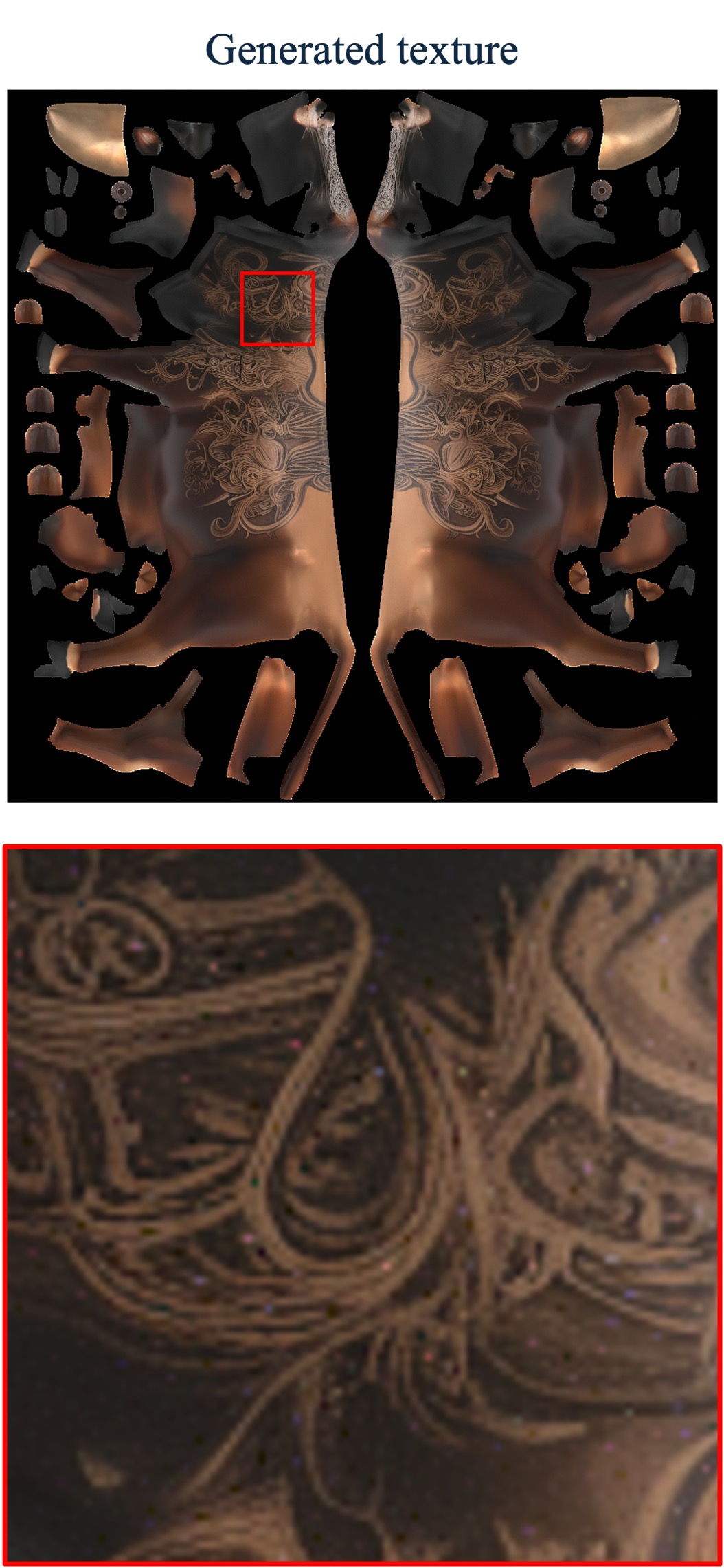}
    \caption*{(a)}
  \end{minipage}\hfill
  \begin{minipage}{.45\linewidth}
    \centering
    \includegraphics[width=\linewidth]{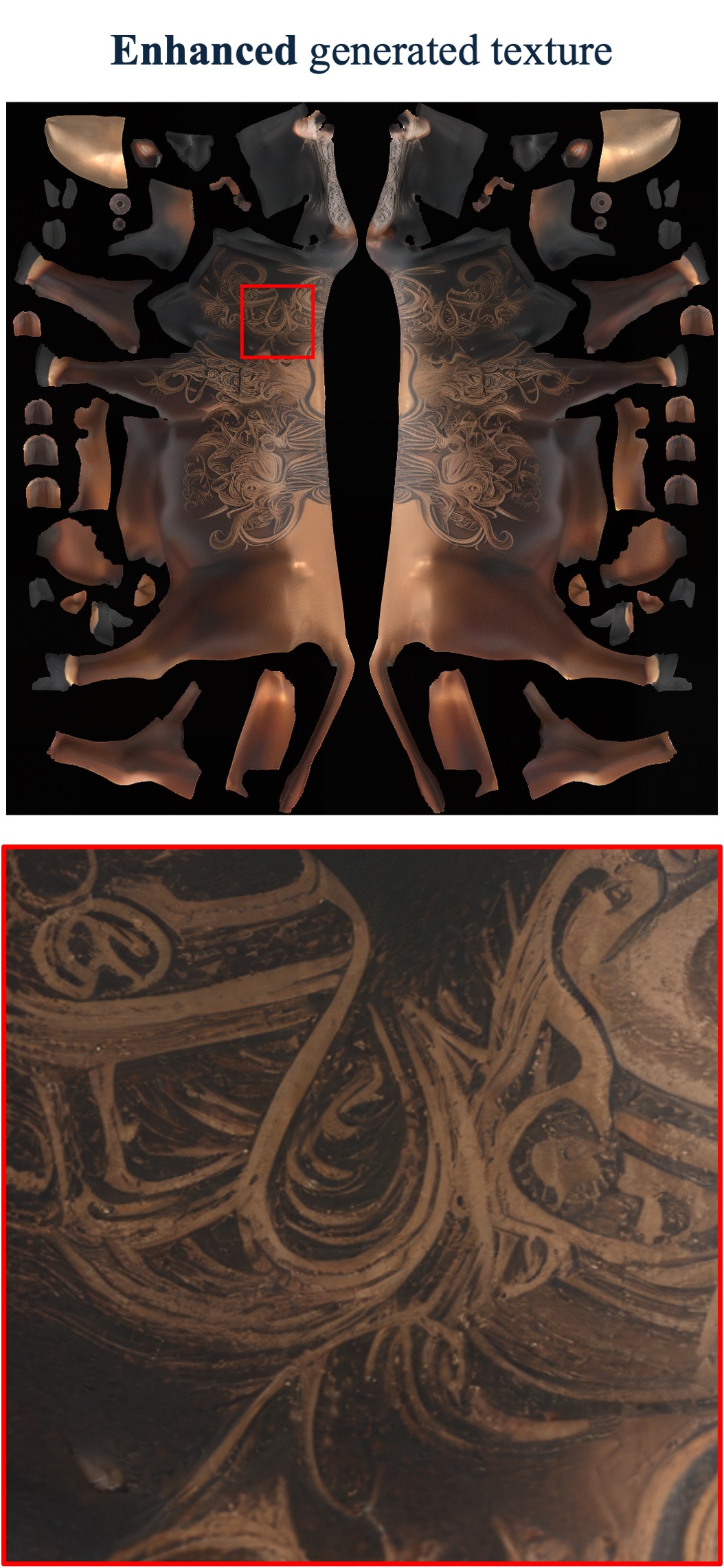}
    \caption*{(b)}
  \end{minipage}
  \caption{\textbf{Texture enhancement UV space.} (a) Generated textures and (b) enhanced generated textures for the text prompt: ``a brown cow covered with an intricate tattoo''. The top row showcases these textures which, despite their initial high quality, have been further enhanced to reveal extremely fine details. The bottom row provides a closer look at these intricate details.} 
  \label{fig::texture_enhancement}
\end{figure*}

\begin{figure*}[h]
  \centering
  \begin{minipage}{.45\linewidth}
    \centering
    \includegraphics[width=\linewidth]{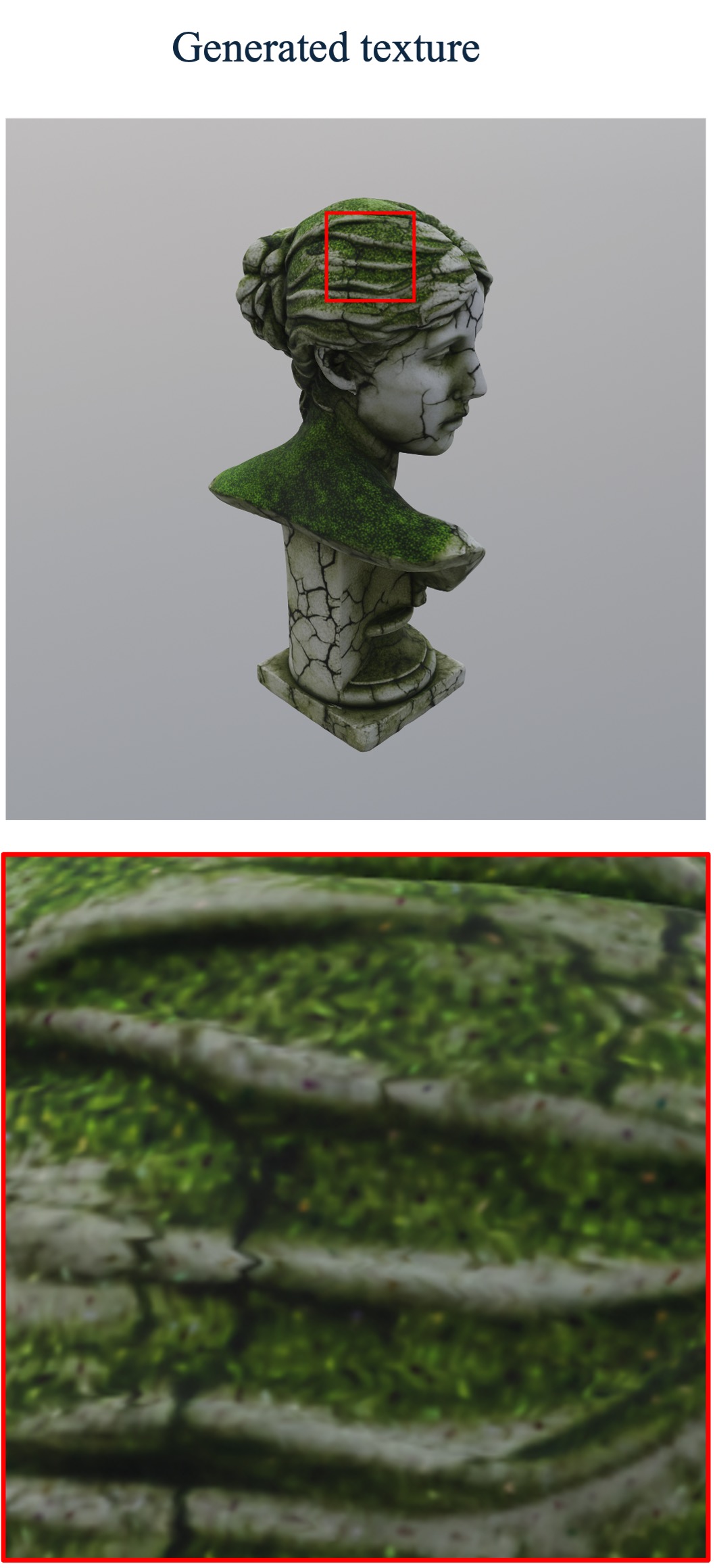}
    \caption*{(a)}
  \end{minipage}\hfill
  \begin{minipage}{.45\linewidth}
    \centering
    \includegraphics[width=\linewidth]{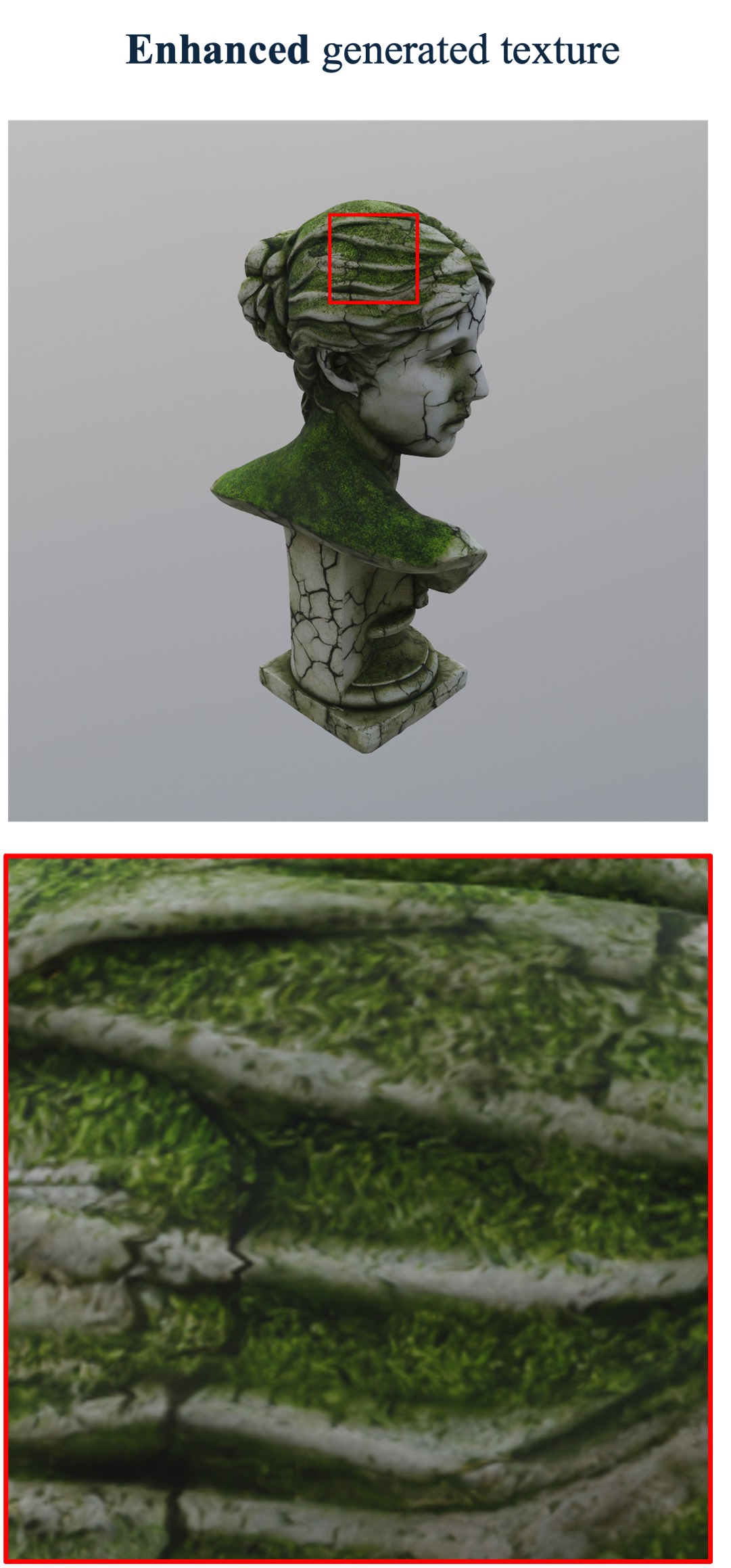}
    \caption*{(b)}
  \end{minipage}
\caption{\textbf{Texture enhancement in 3D.} (a) Generated textures and (b) enhanced generated textures for the text prompt: ``a moss-covered ancient statue made of cracked and semi-shattered stone''. The top row showcases these textures which, despite their initial high quality, have been further enhanced to reveal extremely fine details. The bottom row provides a closer look at these intricate details.}
  \label{fig::texture_enhancement}
\end{figure*}
\section{Experiments details}

\subsection{Evaluation dataset}
All meshes on our evaluation dataset were taken from Sketchfab, under \href{https://creativecommons.org/licenses/by/4.0/}{CC Attribution} license and respecting any NoAI requests by the artists. We present a list of all meshes, with credit to the artists, as well as the prompts we used, in \cref{tab:credit_1,tab:credit_2,tab:credit_3,tab:credit_4,tab:credit_5,tab:credit_6}. Prompts not marked in bold were used during the user study, while those marked in bold were used for FID and KID calculation.

\subsection{Applications}
The vast majority of texture generation evaluation is performed on a single asset detached from any environment (i.e. with a white background). While this is important for capturing fine details and artifacts, it lacks the broader context of a method's ability to produce multiple assets that can blend in an environment, whether realistic or stylized, in a manner that is desirable for real-world applications.
In addition to the single asset evaluations, we demonstrate the usability and applicability of our method in diverse real-world scenarios, utilizing it for building both realistic and stylized environments in virtual reality in \cref{fig::envs} and the supplementary video.

\subsection{User study}

\begin{figure*}
  \includegraphics[width=\linewidth]{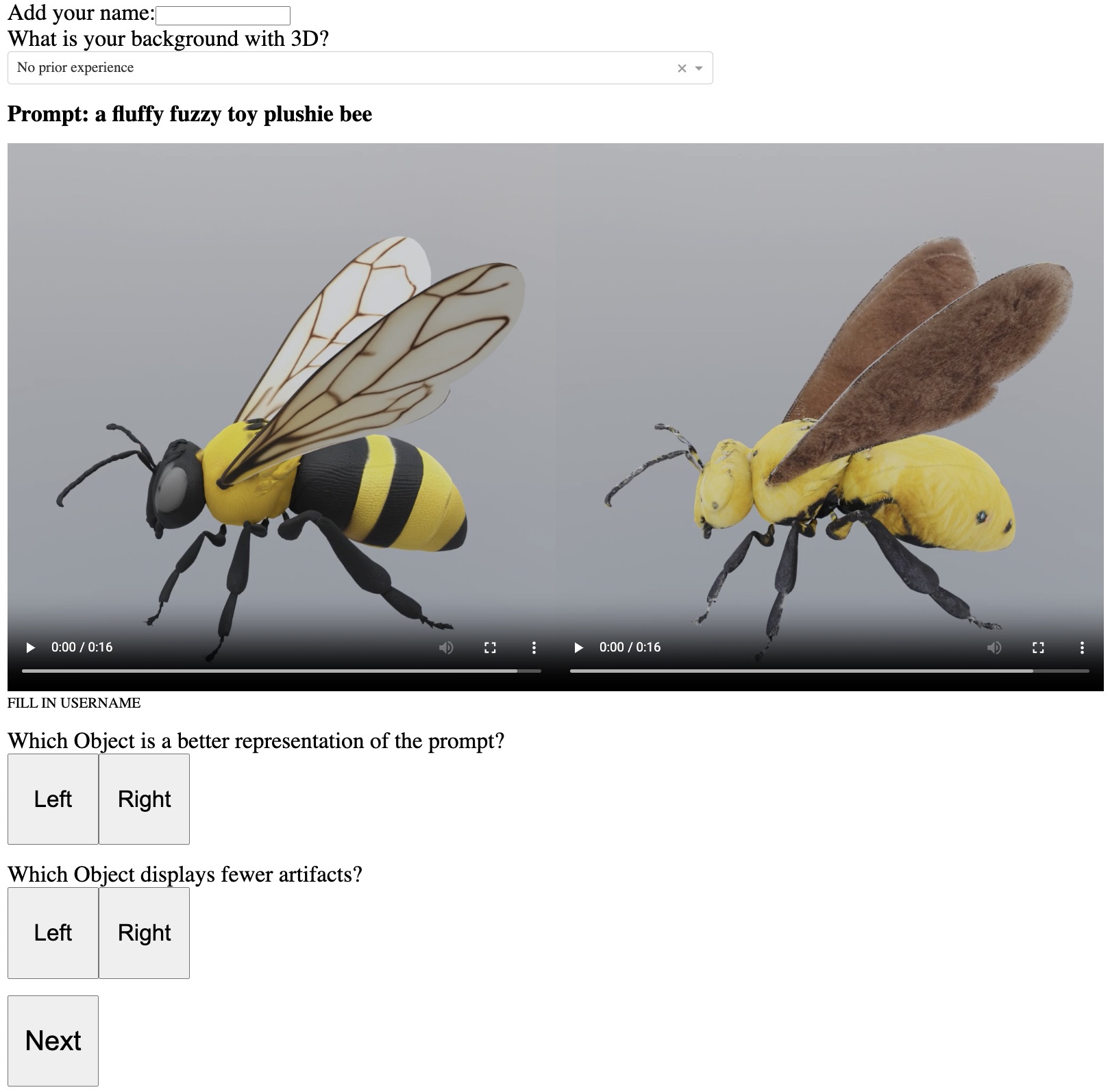}
  \caption{Screenshot from the user study screen.}
  \label{fig:user_survey}
\end{figure*}

We conducted a user study, presenting pair-wise comparisons between our method and five different baselines - TEXTure, Text2Tex, SyncMVD, Paint3D and Meshy on textured meshes. To eliminate biases, left-right ordering, as well as mesh, prompt and baseline orderings have all been randomized.
A screenshot of the survey is shown in \cref{fig:user_survey}.
\Cref{tab:breakdown} includes a breakdown of participants to different backgrounds, according to their familiarity and proficiency with 3D objects. Of the 33 participants we had in our study, 10 were 3D artists, 18 had some proficiency with 3D objects and 5 had no prior background.

\begin{table}
\centering
  \caption{Visualization dataset}
  \vspace{-0.2cm}
  \label{tab:credit_7}
  \begin{tabular}{llll}
  \toprule
    Object & Ref. & Object & Ref.  \\
    \toprule
    Tree 1 & \href{https://sketchfab.com/3d-models/low-poly-stylized-tree-bc1df9f4b7de421886b84af8ee8bcea1}{Link} & Hot Air Balloon & \href{https://sketchfab.com/3d-models/hot-air-balloon-d543cbe5ef774c33b087f56f4d0cb15f}{Link} 
    \\
    Flower & \href{https://sketchfab.com/3d-models/dry-flower-d565e8b73f35430a882789ed9c434299}{Link} & Forest Skybox & \href{https://sketchfab.com/3d-models/forest-clearing-1-top-skybox-ed2c72a25b834f0aac84c4e91b9864cc}{Link} \\
    Tree 2 & \href{https://sketchfab.com/3d-models/lowpoly-tree-b562b2e9f029440c804b4b6d36ebe174}{Link} & Alien Skybox & \href{https://sketchfab.com/3d-models/forest-clearing-1-top-skybox-ed2c72a25b834f0aac84c4e91b9864cc}{Link} \\
    Treasure Chest & \href{https://sketchfab.com/3d-models/stylized-treasure-chest-1c992fcba0a54ff49d817ad27a2b4bd4}{Link} & Fairy Skybox & \href {https://sketchfab.com/3d-models/free-skybox-fairy-forest-day-07f93a261b674a51816ea67ca6882902}{Link} \\
    Treasure Chest 2 & \href{https://sketchfab.com/3d-models/treasure-chest-cece9be19f644b2da6eca2aa69a15fe9}{Link} & Swampland Skybox & \href {https://sketchfab.com/3d-models/free-skybox-pack-fantasy-swamplands-0385b2bf3fe1478fa367824c4dcfef60}{Link}\\
  \bottomrule
\end{tabular}
\end{table}

\begin{table*}
\centering
  \caption{Breakdown of user answers according to proficiency.}
  \label{tab:breakdown}
  \begin{tabular}{lcccccc}
  \toprule
       & \multicolumn{2}{c}{\centering 3D Artists}  
       & \multicolumn{2}{c}{\centering Some Proficiency} 
       & \multicolumn{2}{c}{\centering No Background} \\
       & \multicolumn{1}{c}{\centering Preference}  
       & \multicolumn{1}{c}{\centering Artifacts} 
       & \multicolumn{1}{c}{\centering Preference}  
       & \multicolumn{1}{c}{\centering Artifacts} 
       & \multicolumn{1}{c}{\centering Preference}  
       & \multicolumn{1}{c}{\centering Artifacts} \\
    \midrule
    TEXTure & 78.3\% & 80.7\% & 77\% & 72.1\% & 62.5\% & 50\% \\
    Text2Tex & 83.6\% & 89.6\% & 77.6\% & 75.5\% & 92.3\% & 92.3\% \\
    SyncMVD & 64.3\% & 61.8\% & 67.2\% & 68.3\% & 84.6\% & 84.6\% \\
    Paint3D & 87.3\% & 86.1\% & 68.3\% & 71.2\% & 76.9\% & 76.9\% \\
    Meshy3.0 & 69.1\% & 70.4\% & 56.3\% & 64\% & 80\% & 80\% \\
  \bottomrule
\end{tabular}
\end{table*}

\section{Visualization details}
In addition to meshes used during evaluation, we made use of additional meshes and skyboxes for visualization purposes. These meshes are also under \href{https://creativecommons.org/licenses/by/4.0/}{CC Attribution} license, and can be found in \cref{tab:credit_7}.

\begin{table*}
  \caption{Evaluation dataset. \textbf{Bold} prompts were used for quantitative evaluation (continued in \cref{tab:credit_2}).}
  \label{tab:credit_1}
  \begin{tabularx}{\textwidth}{lll}
    \toprule
       Object & Source & Prompts  \\
    \toprule
    {\multirow{5}{*}{Ant}} & {\multirow{5}{*}{\href{https://sketchfab.com/3d-models/ant-dab7080251674ef98fc83b7604be2ffc}{Link}}} &  a psychedelic colored ant \\ & & a realistic fire ant \\ & & an old ant robot made of rusted metal \\ & & a radioactive green ant \\ && \textbf{a dark brown ant with light brown legs and black eyes} 
    \\
    \midrule
    {\multirow{5}{*}{Bee}} & {\multirow{5}{*}{\href{https://sketchfab.com/3d-models/flying-bee-ba935104d4774583af65828289e8aad5}{Link}}} & rainbow striped bee \\ & & a bee with a green snakeskin pattern on its body \\ & & a fluffy fuzzy toy plushie bee \\ & & a common honey bee \\ && \textbf{a black and yellow fuzzy bee with black eyes}
    \\
    \midrule
    {\multirow{4}{*}{Bottle}} & {\multirow{4}{*}{\href{https://sketchfab.com/3d-models/bottle-checker-669882ec1eac41dfb98bd0d86a4b26e2}{Link}}} & a red wine bottle with a black label with a blue infinity logo \\ & & a bottle filled with layers of colorful sand from the dead sea \\ & & a bottle completely wrapped in old newspapers \\ & & a beer bottle with GenAI written on the label \\
    \midrule
    {\multirow{5}{*}{Boy Room}} & {\multirow{5}{*}{\href{https://sketchfab.com/3d-models/boy-room-19035fc846034003b76e2914ba6dd7a6}{Link}}} & a diorama of a boy and a monster in a room made out of cardboard \\ & & a dollhouse featuring a boy and a big monster \\ & & papercraft diorama of a boy and a big monster, origami folding \\ & & a boy wearing a red cape and a golden sword fighting a green swamp monster \\ && \textbf{an isometric green and gray room with a boy wearing a red cape fighting a purple-yellow monster.}
    \\
    \midrule
    {\multirow{5}{*}{Bracelet}} & {\multirow{5}{*}{\href{https://sketchfab.com/3d-models/bracelet-f1526584726d4e87a735d9732c227a74}{Link}}} & brown leather bracelet with steel studs, horses engraved on the leather \\ & & a golden bracelet with precious stones inlaid around it \\ & & a hand-worn communication device, with a led screen, buttons and lights \\ & & bracelet made of rough weathered wood in a deep forest green color with visible wood grain \\ && \textbf{a dark brown leather bracelet with laces} 
    \\
    \midrule
    {\multirow{5}{*}{Burger}} & {\multirow{5}{*}{\href{https://sketchfab.com/3d-models/burger-realistic-free-18e59d7dbd2243c69f469e0f056f44c4}{Link}}} & an ancient statue of a marble burger in greek or roman style, made of veined red, black, green marble \\ & & a realistic burger with tomatoes, pickles and lettuce in a sesame bun \\ & & a burger-shaped cake, made out of chocolate cake bun, candies and candy floss \\ & & a simple wooden toy shaped like a burger, made out of natural oak wood 
    \\ && \textbf{a succulent burger with poppy seeds on the bun, lettuce, cheese, tomatoes and pickles}
    \\
    \midrule
    {\multirow{5}{*}{Bust}} & {\multirow{5}{*}{\href{https://sketchfab.com/3d-models/sculpture-bust-of-roza-loewenfeld-fc6e731a0131471ba8e45511c7ea9996}{Link}}} & a painted greek sculpture, blonde hair, fair skin, red lipstick, bright blue eyes \\ & & a moss covered ancient statue, made of cracked and semi-shattered stone \\ & & a realistic bust of a heavily tattooed woman, with tribal tattoos covering her face and neck \\ & & a sculpture of a woman painted in the style of Van Gogh \\ && \textbf{a marble sculpture bust of the woman Róża Loewenfeld}
    \\
    \midrule
    {\multirow{5}{*}{Butterfly}} & {\multirow{5}{*}{\href{https://sketchfab.com/3d-models/animated-flying-fluttering-butterfly-loop-80f8d9a6dadc411e89ca366cb0cfb0d9}{Link}}} & a majestic monarch butterfly \\ & & a magical butterfly, arcane sigils on its wings, sparkling glitter on its body \\ & & a crochet butterfly toy from pastel pink and pastel green yarn \\ & & venomous black butterfly with bright red and orange markings\\ && \textbf{a monarch butterfly with orange and yellow wings, white and yellow dots and a black body}
    \\ 
    \midrule
    {\multirow{5}{*}{Cactus}} & {\multirow{5}{*}{\href{https://sketchfab.com/3d-models/stylized-cactus-d515ed6d19184b80830a6feba04863a8}{Link}}} & alien blue cactus plant with red spikes and white flowers \\ & & a realistic model of a cactus \\ & & contemporary plastic statue of a green cactus with white geometric patterns on it \\ & & cactus colored with vertical repeating stripes of purple, black and lavender \\ && \textbf{three green cartoonish cactus with red pink and yellow spikes}
    \\
  \bottomrule
\end{tabularx}
\end{table*}

\begin{table*}
  \caption{Evaluation dataset \textbf{Bold} prompts were used for quantitative evaluation (continued in \cref{tab:credit_3}).}
  \label{tab:credit_2}
  \begin{tabularx}{\textwidth}{lll}
  \toprule
       Object & Reference & Prompts  \\
    \toprule
    {\multirow{5}{*}{Carriage}} & {\multirow{5}{*}{\href{https://sketchfab.com/3d-models/carriage-88051117075346b4bf8c4b1d90679f3c}{Link}}} & a medieval wagon made out of light wood, with colorful striped orange and white awning \\ & & a modern steel carriage with white fabric awning \\ & & a mystical magician’s wagon, colored in purple with a yellow symbol of an eye painted on it \\ & & a wagon made out of unprocessed wood, with awning made out of leaves, vines and branches \\ && \textbf{a wooden carriage with white fabric and white cargo}
    \\
    \midrule
    {\multirow{5}{*}{Cartoon Car}} & {\multirow{5}{*}{\href{https://sketchfab.com/3d-models/car-2019-21067530b93c465fbc7765a717933c4f}{Link}}} & a cartoon car in the style of 3D animation \\ & & a futuristic space car \\ & & an old rusted car for a post apocalyptic game \\ & & a toy car made out of pastel colored plastic with red wheels \\ && \textbf{a busted up cartoonish yellow car with some rust}
    \\
    \midrule
    {\multirow{5}{*}{Cartoon Plane}} & {\multirow{5}{*}{\href{https://sketchfab.com/3d-models/cartoon-plane-edbb12ffdd3c408d8908e02cbc1ba137}{Link}}} & a realistic plane with blue wings and white body \\ & & a futuristic robot car made of white and silver metal \\ & & an old rusted car for a post apocalyptic game \\ & & a toy car made out of blue and white plastic with red wheels \\ && \textbf{a blue and white cartoonish light airplane}
    \\
    \midrule
    {\multirow{5}{*}{Classic Car}} & {\multirow{5}{*}{\href{https://sketchfab.com/3d-models/fancy-car-780c75b7713b4811b0f7b447df310889}{Link}}} & an old classic car made out of wood and brown leather \\ & & a light blue old classic car \\ & & a steampunk car lined with patches of leather \\ & & a wooden carved car embossed with an intricate design \\ && \textbf{a pale green vintage classic car}
    \\
    \midrule
    {\multirow{5}{*}{Cow}} & {\multirow{5}{*}{\href{https://sketchfab.com/3d-models/cow-idle-16493ce34a4f41a6ba7b7d31032317c1}{Link}}} & a cow made of rusted metal \\ & & a realistic black and white holstein cow \\ & & a brown cow covered with an intricate tattoo \\ & & a white cow with golden horns and golden hooves \\ && \textbf{a holstein cow with black and white spots and gray horns}
    \\
    \midrule
    {\multirow{5}{*}{Cutlass}} & {\multirow{5}{*}{\href{https://sketchfab.com/3d-models/octopus-cutlass-f5ba95d8409a44f180ab84e361183d4a}{Link}}} & toy sword made out of red and gold plastic \\ & & a pirate sword with a bronze hilt and metallic blade, japanese kanji written across the blade \\ & & a sword with a lattice of blue neon led-light across its blade \\ & & a wooden sword with barnacles on the hilt \\ && \textbf{a shiny metal sword with a golden blade wrapped with red tape}
    \\
    \midrule
    {\multirow{5}{*}{Eclair}} & {\multirow{5}{*}{\href{https://sketchfab.com/3d-models/fancy-eclair-309600e4f0094a0b97f9c09075704651}{Link}}} & a realistic raspberry eclair with cream and raspberries on top \\ & & a realistic chocolate eclair with rainbow colored cream \\ & & a yellow lemon and vanilla eclair \\ & & a graphite drawing of an eclair with intricate shading patterns \\ && \textbf{an eclair with pink whipped cream and raspberries}
    \\
    \midrule
    {\multirow{5}{*}{Ender Dragon}} & {\multirow{5}{*}{\href{https://sketchfab.com/3d-models/realistic-minecraft-ender-dragon-a43e3ede784f4c52aeeec355784057c1}{Link}}} & a realistic red dragon \\ & & a steampunk dragon wearing a leather armor \\ & & a cobalt dragon with lapis lazuli wings \\ & & a stone statue of a majestic dragon decorated with jewelry \\ && \textbf{a black shiny dragon with grayish wings}
    \\
    \midrule
    {\multirow{5}{*}{Fish}} & {\multirow{5}{*}{\href{https://sketchfab.com/3d-models/koi-fish-236859b809984f52b70c94fd040b9c59}{Link}}} & a white and orange koy fish \\ & & a robotic koi fish made of orange metal \\ & & an alabaster statue of a fish with delicate gold veins inlaid across its body \\ & & a koi fish wearing bronze armor \\ && \textbf{a white koi fish with red and orange spots}
    \\
  \bottomrule
\end{tabularx}
\end{table*}

\begin{table*}
  \caption{Evaluation dataset \textbf{Bold} prompts were used for quantitative evaluation (continued in \cref{tab:credit_4}).}
  \label{tab:credit_3}
  \begin{tabularx}{\textwidth}{lll}
  \toprule
       Object & Reference & Prompts  \\
    \toprule
    {\multirow{5}{*}{Football Helmet}} & {\multirow{5}{*}{\href{https://sketchfab.com/3d-models/american-football-helmet-aa779ac1b7d8459496d92e116b460ab6}{Link}}} & a gamer vr football helmet, cyberpunk style with neon lights \\ & & a football helmet with a mascot painting \\ & & a rainbow football helmet \\ & & a post apocalyptic rusty football helmet, with dirt, dust, and stains on it \\ && \textbf{an old red football helmet with a white drawing of a creepy skull on the back}
    \\
    \midrule
    {\multirow{5}{*}{Game Controller}} & {\multirow{5}{*}{\href{https://sketchfab.com/3d-models/gamepad-low-poly-0431ada1c7234acab1ec7966ffcbcf6f}{Link}}} & a game controller made out of wood with visible wood grain \\ & & a game controller that looks like an exposed green circuit board \\ & & an ancient stone statue of a game controller with colorful buttons \\ & & a pink birthday cake in the shape of a game controller with chocolate buttons \\ && \textbf{a white game controller}
    \\
    \midrule
    {\multirow{5}{*}{Goggles}} & {\multirow{5}{*}{\href{https://sketchfab.com/3d-models/glasses-mixamo-13-custom-30711b2ca5414752ab2fda0203557fee}{Link}}} & goggles with butterflies on the strap \\ & & steampunk goggles made out of brown leather and brass \\ & & goggles with polka dots in the style of yayoi kusama \\ & & goggles painted in the style of van gogh \\ && \textbf{black goggles with silver rings}
    \\
    \midrule

    {\multirow{5}{*}{Grapes}} & {\multirow{5}{*}{\href{https://sketchfab.com/3d-models/grape-downloadable-bf09c469e83045f9b91667fdbb277e22}{Link}}} & a realistic cluster of red grapes \\ & & magical glittering purple grapes, with gold dust on them \\ & & crochet grapes, made of colorful thick wool yarn \\ & & alien metallic grapes with strange engravings on them \\ && \textbf{a bunch of black grapes}
    \\
    \midrule
    {\multirow{5}{*}{Handbag}} & {\multirow{5}{*}{\href{https://sketchfab.com/3d-models/atom-shopping-handbag-4bedaea868b84e33a8b6b7ddecfbb51d}{Link}}} & a cute handbag with a whacky llama illustration on it \\ & & an expensive pale leather handbag, high-end, high fashion \\ & & a pink fluffy and fuzzy handbag with googly eyes \\ & & a handbag made of bricks and metal \\ && \textbf{a leather-like solid light beige handbag with gold-tone metal clasps}
    \\
    \midrule
    {\multirow{5}{*}{Hovercraft}} & {\multirow{5}{*}{\href{https://sketchfab.com/3d-models/free-cyberpunk-hovercar-3205b1075bb44ffc826bce0c2a04d74c}{Link}}} & an art deco hovercraft, with gold geometric patterns \\ & & a neon cyberpunk hovercraft, in the style of japanese neon signs \\ & & a desert hovercraft covered with camouflage nets \\ & & a bright red metallic hovercraft 
    \\ && \textbf{a black futuristic hovercraft}
    \\
    \midrule
    {\multirow{5}{*}{Ice Axe}} & {\multirow{5}{*}{\href{https://sketchfab.com/3d-models/texture-painting-an-ax-006c1bf4bd39479b9134ef32a0618897}{Link}}} & a magical axe with lightning powers set with blue gems \\ & & an axe studded with diamonds and gems \\ & & an old battle axe weathered with scratches and cracks and splattered with blood \\ & & an axe carved out of a single piece of wood with intricate engraving. blade made of same wood as handle \\ && \textbf{a cartoonish axe with a wooden handle and a gray blade with blue lightning}
    \\
    \midrule
    {\multirow{5}{*}{Jellyfish}} & {\multirow{5}{*}{\href{https://sketchfab.com/3d-models/jellyfish-42da7b26f173427f89c9ddb64be1be73}{Link}}} & neon colored purple and green jellyfish \\ & & a robotic jellyfish made out of dark metal, with blue led lights across its tentacles \\ & & a flying jellyfish with tentacles made out of delicate white and light blue feathers \\ & & wooden carving of a jellyfish, unprocessed olive wood \\ && \textbf{a pink and green jellyfish with intricate patterns}
    \\
    \midrule
    {\multirow{5}{*}{Knight}} & {\multirow{5}{*}{\href{https://sketchfab.com/3d-models/medieval-knight-sculpture-game-ready-6cdd055b4afa41eb9360dbbfe75c7f10}{Link}}} & a knight wearing steel armor with a golden lion engraving on his chest \\ & & a knight wearing armor with swirling rainbows on it \\ & & a papercraft knight, origami with colorful papers covered by dots and geometric patterns \\ & & a knight wearing samurai armor resembling a blooming sakura \\ && \textbf{a knight dressed in full-body suit of armor}
    \\
    \midrule
    {\multirow{5}{*}{Laptop}} & {\multirow{5}{*}{\href{https://sketchfab.com/3d-models/cyberpunk-laptop-concept-design-fddc4e68cc6c498b88b19af1a05bd420}{Link}}} & a realistic laptop with a background picture of green fields and blue sky on its screen \\ & & a laptop running a pixelated 80s style video game \\ & & a laptop made entirely out of rock \\ & & a gray laptop with a colorful keyboard and glowing keys \\ && \textbf{a laptop showing an image of a city at night with a car, the laptop features mostly}
     \\
  \bottomrule
\end{tabularx}
\end{table*}

\begin{table*}
  \caption{Evaluation dataset \textbf{Bold} prompts were used for quantitative evaluation (continued in \cref{tab:credit_5}).}
  \label{tab:credit_4}
  \begin{tabularx}{\textwidth}{lll}
  \toprule
       Object & Reference & Prompts  \\
    \toprule
    {\multirow{5}{*}{Mushrooms}} & {\multirow{5}{*}{\href{https://sketchfab.com/3d-models/mushrooms-c2dbb88a5e9c4e82a180e13b472015ed}{Link}}} & toadstool, red cap with white polka dots \\ & & magical deep purple mushrooms, sparkling glitter on the cap \\ & & cute cartoon mushrooms with a cute face, big eyes on the stalk \\ & & realistic brown mushrooms \\ && \textbf{dark red and orange mushrooms with white stems}
    \\
    \midrule
    {\multirow{5}{*}{Ottomans}} & {\multirow{5}{*}{\href{https://sketchfab.com/3d-models/1800s-cushioned-foot-stools-35b9301bc0134330a80860c2542f89cc}{Link}}} & 3 wooden ottomans with portuguese azulejo patterned cushions \\ & & colorful patchwork ottomans with elaborate wooden carvings on the bases \\ & & ottomans made of dark wooden planks and deep scarlet velvet cushions \\ & & Bamboo ottomans with simple bamboo cushion, ink paintings on the bases \\ && \textbf{three wooden ottoman stools with floral patterns}
    \\
    \midrule
    
    {\multirow{5}{*}{Potted Plant}} & {\multirow{5}{*}{\href{https://sketchfab.com/3d-models/potted-plant-83986836419445edb3e507b9eb0c786b}{Link}}} & a realistic potted plant \\ & & a realistic potted plant with colorful leaves \\ & & a plant made out of colorful origami paper \\ & & a potted plant made out of newspaper \\ && \textbf{a simple potted plant with flat green leaves and a reddish pot}
    \\
    \midrule
    {\multirow{5}{*}{Pterodactyl}} & {\multirow{5}{*}{\href{https://sketchfab.com/3d-models/animated-flying-pteradactal-dinosaur-loop-09f6b1ff7bfd4a2d8d9c0b9827f2a708}{Link}}} & a white marble pterodactyl \\ & & a steampunk pterodactyl wearing a leather armor \\ & & a realistic pterodactyl with bat wings \\ & & a pterodactyl made of rocks and lava \\ && \textbf{a brown-yellow pterodactyl}
    \\
    \midrule
    {\multirow{5}{*}{Roman Helmet}} & {\multirow{5}{*}{\href{https://sketchfab.com/3d-models/roman-helmet-fe4dc3a4c6a141b795c97e2e94b336a3}{Link}}} & a bronze metal helmet with vibrant red plume adorning it \\ & & a crochet helmet in rainbow colors, a rainbow colored plume on top \\ & & a yellow-green helmet made of snakeskin, with purple ruffle on top \\ & & an old rusty cracked helmet \\ && \textbf{a shiny roman corinthian helmet with a red plume adorned with intricate engravings} \\ & & \textbf{and patterns}
    \\
    \midrule
    {\multirow{5}{*}{Rose}} & {\multirow{5}{*}{\href{https://sketchfab.com/3d-models/rose-3d-4aad2c8cd1564bc49fddb00df80d5776}{Link}}} & a red rose \\ & & a rose with a pattern of pink-white stripes on its petals \\ & & a rainbow colored rose with each petal leaf in a different color \\ & & a magical blue rose with gold sparkling dusting \\ && \textbf{a red rose with green leaves}
    \\
    \midrule
    {\multirow{5}{*}{Row Boat}} & {\multirow{5}{*}{\href{https://sketchfab.com/3d-models/row-boat-d163bf57f22f430b96118b174b12a2d8}{Link}}} & a wooden red row boat, a black thick stripe at the bottom, a repeating pattern of shells across the body \\ & & metallic row boat, made out of sheets of thin lightweight metal \\ & & ancient egyptian boat with elaborate egyptian paintings and hieroglyphics painted on it \\ & & a wooden boat made of untreated wood planks with visible wood grain \\ && \textbf{an old wooden row boat}
    \\
    \midrule
    {\multirow{5}{*}{Seahorse}} & {\multirow{5}{*}{\href{https://sketchfab.com/3d-models/seahorse-sf-a803bd39df534becb587453c27a0a9a9}{Link}}} & fantastic magical seahorse covered in sparkling gems \\ & & a realistic yellow seahorse \\ & & a seahorse wearing elaborate metal armor with engravings of sealife \\ & & a futuristic robotic seahorse, with led lights and metal platings \\ && \textbf{a yellow orange seahorse with orange-brown stripes and black eyes}
    \\
  \bottomrule
\end{tabularx}
\end{table*}

\begin{table*}
  \caption{Evaluation dataset \textbf{Bold} prompts were used for quantitative evaluation - continued in \cref{tab:credit_6}).}
  \label{tab:credit_5}
  \begin{tabularx}{\textwidth}{lll}
  \toprule
       Object & Reference & Prompts  \\
    \toprule
    {\multirow{6}{*}{Shield}} & {\multirow{6}{*}{\href{https://sketchfab.com/3d-models/gil-galads-shield-76a5f135cf2746c2a4425f5bd6af8021}{Link}}} & gold and blue high fantasy decorated shield \\ & & black obsidian shield with a large ruby in its center \\ & & shield made out of opaque yellow crystallized amber \\ & & wooden shield with a painting of ouroboros on it \\ && \textbf{an elegant and elongated copper and blue shiny shield adorned with intricate designs and} \\ &&  \textbf{a blue gem stone in the middle}
    \\
    \midrule
    {\multirow{6}{*}{Spider}} & {\multirow{6}{*}{\href{https://sketchfab.com/3d-models/spiderthing-take-3-10bb4cf49d304d64afd2b829666f6caf}{Link}}} & a black widow spider with vibrant red markings on its back \\ & & a monstrous flesh spider, sewn together from various parts \\ & & a spider wearing an engraved bronze plated armor \\ & & a pastel colored crochet spider \\ && \textbf{a scary black spider-like monster with a grayish-red bottom and yellow spikes coming} \\ && \textbf{out of the end of each leg}
    \\ 
    \midrule
    {\multirow{5}{*}{Teacup}} & {\multirow{5}{*}{\href{https://sketchfab.com/3d-models/old-tea-cup-eec2cfa2536d42239b3b3c1ca2021c51}{Link}}} & moroccan-style ceramic tea cup with arabesque patterns \\ & & english ceramic tea cup with floral design, the painted flowers are pink\\ & & brute concrete gray tea cup \\ & & fine china tea cup, pink blue and green colors with design of a crane near a river \\ && \textbf{a white tea cup and coaster with blue intricate drawings}
    \\
    \midrule
    {\multirow{5}{*}{Teapot}} & {\multirow{5}{*}{\href{https://sketchfab.com/3d-models/teapot-d9dba4e568cc4f6786febf6dc1fef91d}{Link}}} & black ceramic teapot in the style of ancient greece, with red clay drawings of greek myths \\ & & futuristic hi-tech smart teapot with led lights, sensors and a temperature measurement system \\ & & a ceramic teapot with a intricate carvings on it \\ & & a delicate floral teacup, vines and flowers spanning around the container \\ && \textbf{a grayish teapot with drawings of pink flower vines}
    \\
    \midrule
    {\multirow{5}{*}{Telescope}} & {\multirow{5}{*}{\href{https://sketchfab.com/3d-models/steampunk-telescope-eeae8668106246398fc7956d2fb225e5}{Link}}} & a rusty brass telescope from the golden age of piracy \\ & & a futuristic neon-lit spyglass, black chassis with neon pink and blue lights \\ & & a steampunk spyglass made of leather \\ & & an ancient moss covered rock telescope 
    \\ && \textbf{an antique golden telescope with a wooden patch and black lenses}
    \\
    \midrule
    {\multirow{6}{*}{Toad Hall}} & {\multirow{6}{*}{\href{https://sketchfab.com/3d-models/the-wind-in-the-willows-toad-hall-04c1d5a63a8448daa0ac8f8bcd380206}{Link}}} & an alsatian timber house and a green frog wearing a black and white tux \\ & & a frog wearing a white outfit in front of a santorini style white house with blue rooftop \\ & & a brown toad wearing red near a scary halloween mansion \\ & & a snow covered diorama of a house and a toad wearing a victorian outfit \\ && \textbf{a green toad with khaki clothes standing on a green grassy island with a wooden town} \\ && \textbf{and green trees}
    \\
    \midrule
    {\multirow{6}{*}{Togo Cup}} & {\multirow{6}{*}{\href{https://sketchfab.com/3d-models/coffee-to-go-cup-cdb8ee2283664400939454f36b68fb53}{Link}}} & a togo coffee cup with a blue GenAI logo on it \\ & & a old and dirty coffee cup with coffee stains \\ & & a metal coffee cup that looks like a high-tech gadget \\ & & a take-away coffee cup embossed with gold and precious gems \\ && \textbf{a brown take-away cup with white drawings of houses trees and windmills with a black top} \\ && \textbf{and a white bottom}
    \\
    \midrule
    {\multirow{5}{*}{Toy Gun}} & {\multirow{5}{*}{\href{https://sketchfab.com/3d-models/toy-gun-2b4f4fdccde8429a84c17ef2e3363884}{Link}}} & a plastic toy gun made out of blue and orange plastic \\ & & a high tech gun made out of metal and mechanical parts \\ & & a toy gun with the word “GenAI” written on it \\ & & a toy gun made out of wood with visible wood grain \\ && \textbf{a blue orange and black nerf gun with white markings}
    \\
    \midrule
    {\multirow{5}{*}{Toy Plane}} & {\multirow{5}{*}{\href{https://sketchfab.com/3d-models/toy-plane-9b4e13700da14d31b9d923c499e18f64}{Link}}} & a toy plane made out of purple and yellow plastic with chrome propeller \\ & & a toy plane made out of wood \\ & & a realistic stunt biplane with red and blue stripes \\ & & an old rusted war biplane \\ && \textbf{a purple cartoonish airplane with light yellow flame drawings}
    \\
  \bottomrule
\end{tabularx}
\end{table*}

\begin{table*}
  \caption{Evaluation dataset \textbf{Bold} prompts were used for quantitative evaluation - continued.}
  \label{tab:credit_6}
  \begin{tabularx}{\textwidth}{lll}
  \toprule
    Object & Reference & Prompts  \\
    \toprule
    {\multirow{5}{*}{Tree Stump}} & {\multirow{5}{*}{\href{https://sketchfab.com/3d-models/drevo-7a0f2d413b5846f591b40580f783c53c}{Link}}} & a light gray petrified wood on a dirt mound \\ & & snow covered tree stump during winter \\ & & a moss covered tree stump on fresh green grass \\ & & burnt wood on soot, scorch marks on the tree stump \\ && \textbf{a wood stump covered with some snow laying on a snowy patch of ground}
    \\
    \midrule
    {\multirow{5}{*}{Triceratops}} & {\multirow{5}{*}{\href{https://sketchfab.com/3d-models/triceratops-f87fdfc7193047218db5f7a8ae173e0d}{Link}}} & a hyperrealistic triceratops with pronounced grooves on its skin \\ & & a plush toy orange triceratops with white horns and white claws \\ & & a triceratops painted in the style of van gogh \\ & & a triceratops covered with tattoos of intricate designs \\ && \textbf{a green triceratops with brown stripes on its back and yellow horns and nails}
    \\
    \midrule
    {\multirow{5}{*}{Unicorn}} & {\multirow{5}{*}{\href{https://sketchfab.com/3d-models/unicorn-wip-b5963898b2af4b25ad2598b07f14f0fd}{Link}}} & a pastel colored magical unicorn \\ & & a forest unicorn, covered with swirling blue shiny patterns amidst leaves covering its body \\ & & a unicorn with a snakeskin body in the colors of black and yellow \\ & & a nightmarish hell horse in the color of soot, fiery flaming mane and ruby eyes \\ && \textbf{a white unicorn with black eyes and snout}
    \\
    \midrule
    {\multirow{5}{*}{Vase}} & {\multirow{5}{*}{\href{https://sketchfab.com/3d-models/jarregrandetexture-ff819642d09446fa92d9a0f477657453}{Link}}} & an old vase made out of clay covered with ancient hieroglyphs \\ & & a priceless ming vase made out of white porcelain with a blue pattern \\ & & a black and red chinese vase with an intricate drawing \\ & & a cracked terracotta vase with various cracks \\ && \textbf{a brown-orange base with some rust}
    \\
    \midrule
    {\multirow{5}{*}{Watermelon}} & {\multirow{5}{*}{\href{https://sketchfab.com/3d-models/jiggly-watermelon-jello-c15e41a62b46487fa6dcc67af7f7acee}{Link}}} & a jelly cartoon watermelon \\ & & an autumn pumpkin pie \\ & & a slice of yellow lemon \\ & & a metallic round puzzle key with mysterious engravings on it \\ && \textbf{a cartoonish black watermelon with green skin and black interior}
    \\ 
    \midrule
    {\multirow{5}{*}{Wine Barrel}} & {\multirow{5}{*}{\href{https://sketchfab.com/3d-models/vine-barrel-64369e53063d4fc3a04cdd042ae1d33b}{Link}}} & a realistic wooden wine barrel with deep scarlet wine stains on them \\ & & a circus barrel, colored in bright simple colors with an image of a clown painted on it \\ & & a rusted metal barrel with graffiti on it \\ & & a dark red wooden barrel with steel hoops \\ && \textbf{a wooden light colored wine barrel with metal stripes and brown base legs}
    \\
    \midrule
    {\multirow{5}{*}{Wooden Crate}} & {\multirow{5}{*}{\href{https://sketchfab.com/3d-models/acg-final-wooden-crate-2ccac0eaa9b14b059be313dc079433eb}{Link}}} & an old wooden crate made out of rotten wood \\ & & a blue plastic crate covered with graffiti \\ & & an old rusted metal crate \\ & & a new wooden storage crate with chinese characters on it \\ && \textbf{an old wooden crate with black markings}
    \\
    \midrule
    Wooden Toy & \href{https://sketchfab.com/3d-models/block-creation-29f8ef63aad24b6c99a6c3bca5f7ab55}{Link} & \textbf{a wooden toy tower}
    \\
    \midrule
    {\multirow{4}{*}{Stanford Bunny}} & {\multirow{4}{*}{\href{https://graphics.stanford.edu/data/3Dscanrep/}{Link}}} & realistic white rabbit with long fur, pink eyes, and black paws \\ & & a bunny made out of small pebbles of many shades of gray \\ & & a futuristic bunny with dark fur and stripes of bright neon colors \\ && a sand sculpture of a bunny with engraving of an intricate pattern 
    \\
    \midrule
    {\multirow{5}{*}{Stanford Armadillo}} & {\multirow{5}{*}{\href{https://graphics.stanford.edu/data/3Dscanrep/}{Link}}} & an armadillo man wearing a beautiful medieval armor crafted with gold and precious gems \\ & & an armadillo creature made out of a mosaic of small squares with red, blue, and white colors \\ & & a realistic armadillo creature with a shell like a green turtle on its back \\ & & an armadillo man action figure made out of pieces of brown and purple plastic with white claws \\ & & and black eyes 
    \\
  \bottomrule
\end{tabularx}
\end{table*}

%% file: figures/diversity/fig_diversity_bunny.tex
\begin{figure*}[h]
\centering





    \begin{tabular}{ccccccc}

		\includegraphics[width=0.12\linewidth]{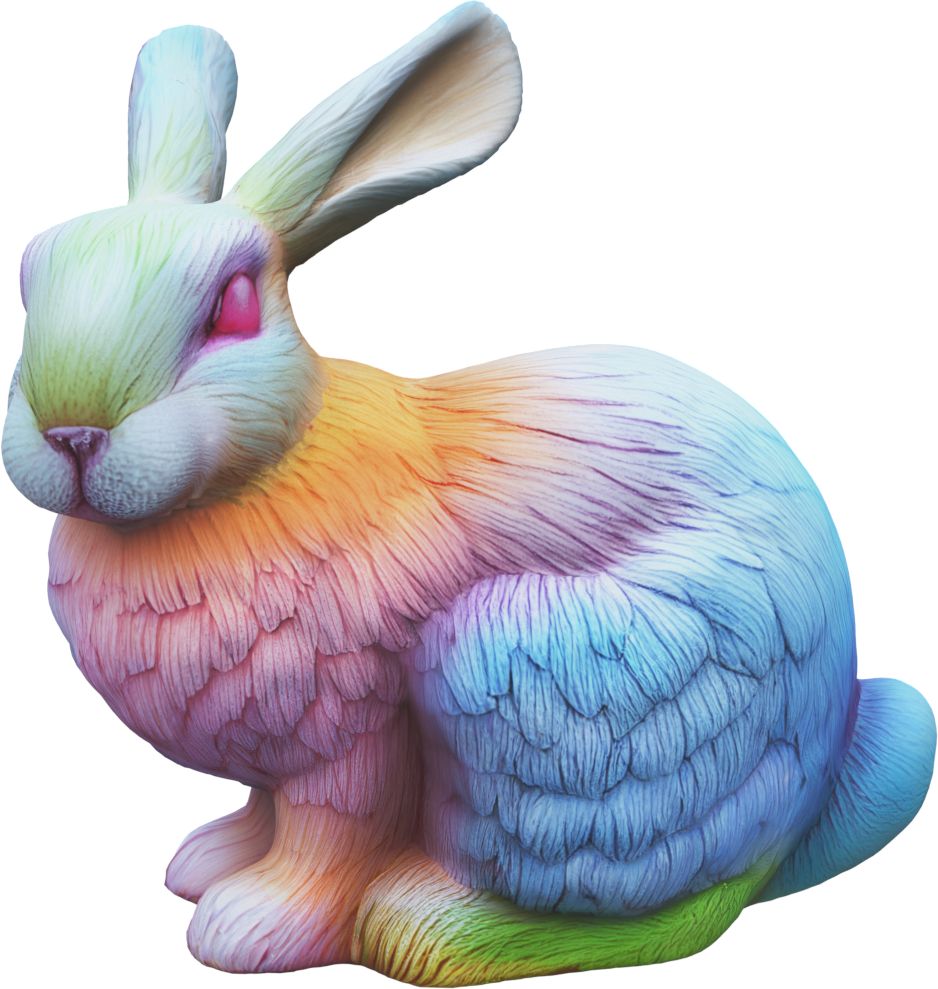} &
		\includegraphics[width=0.12\linewidth]{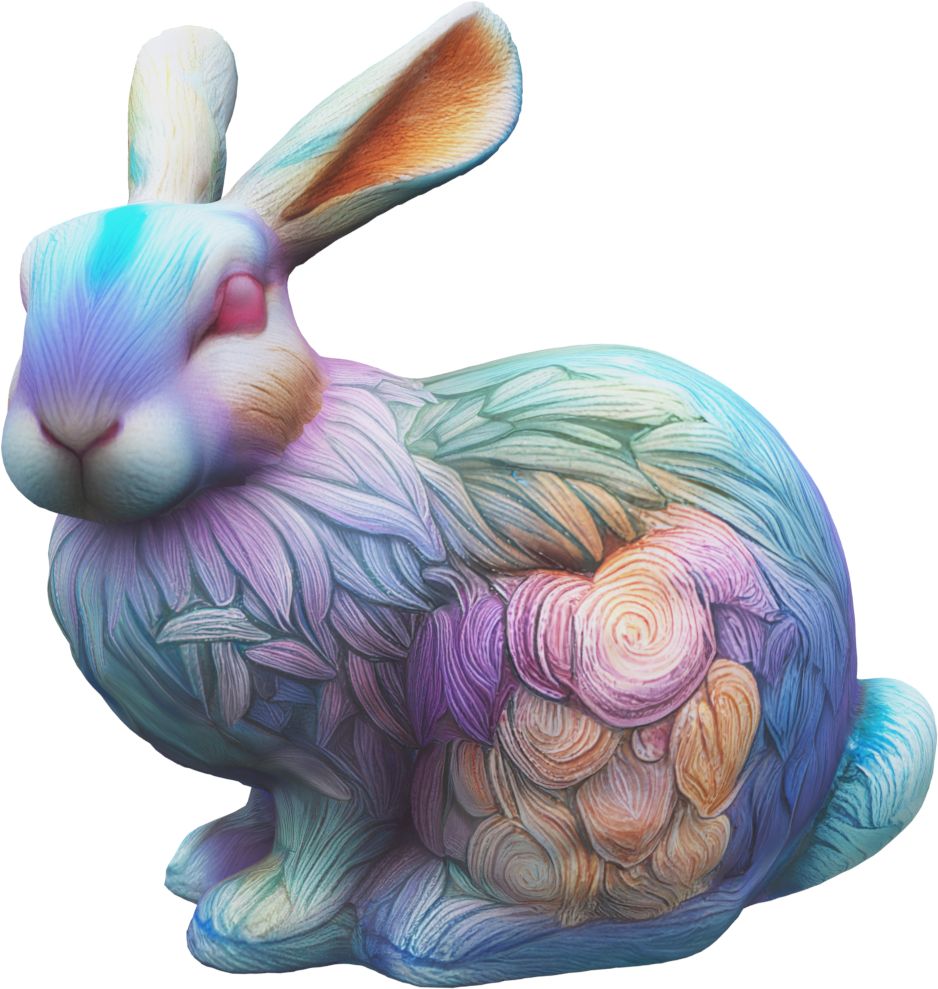} &
		\includegraphics[width=0.12\linewidth]{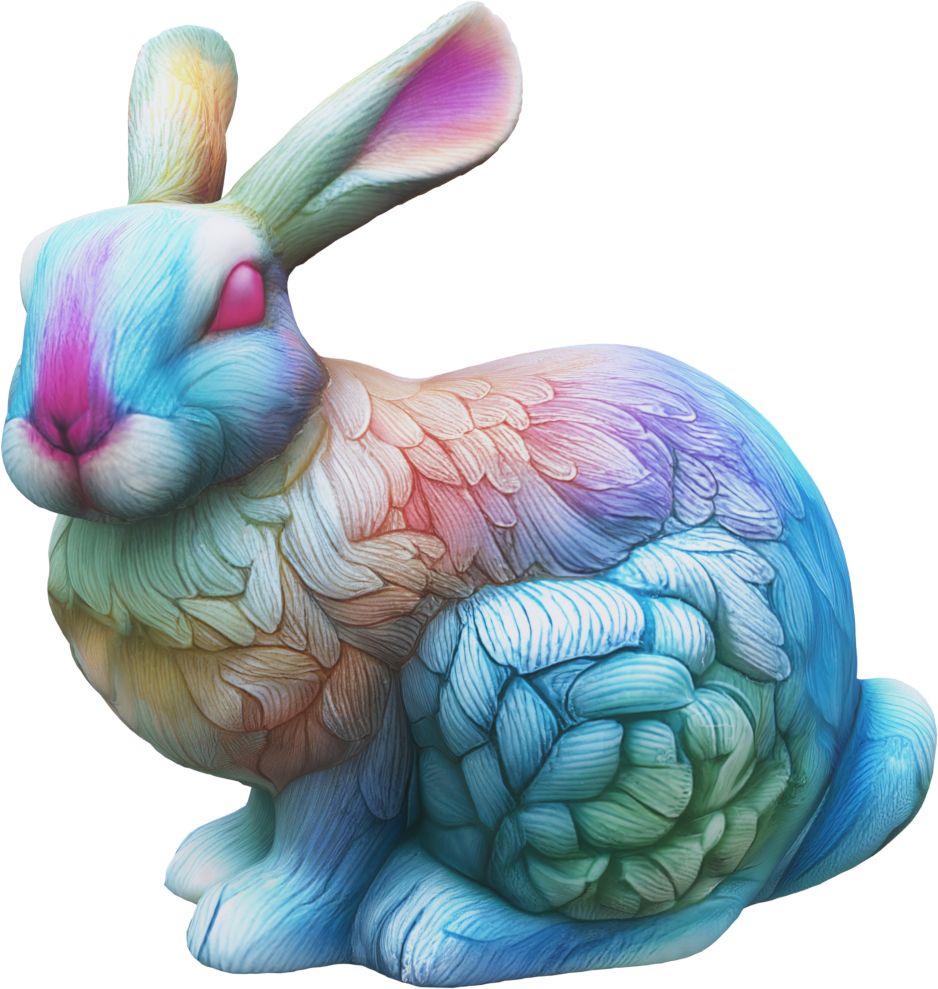} &
		\includegraphics[width=0.2\linewidth]{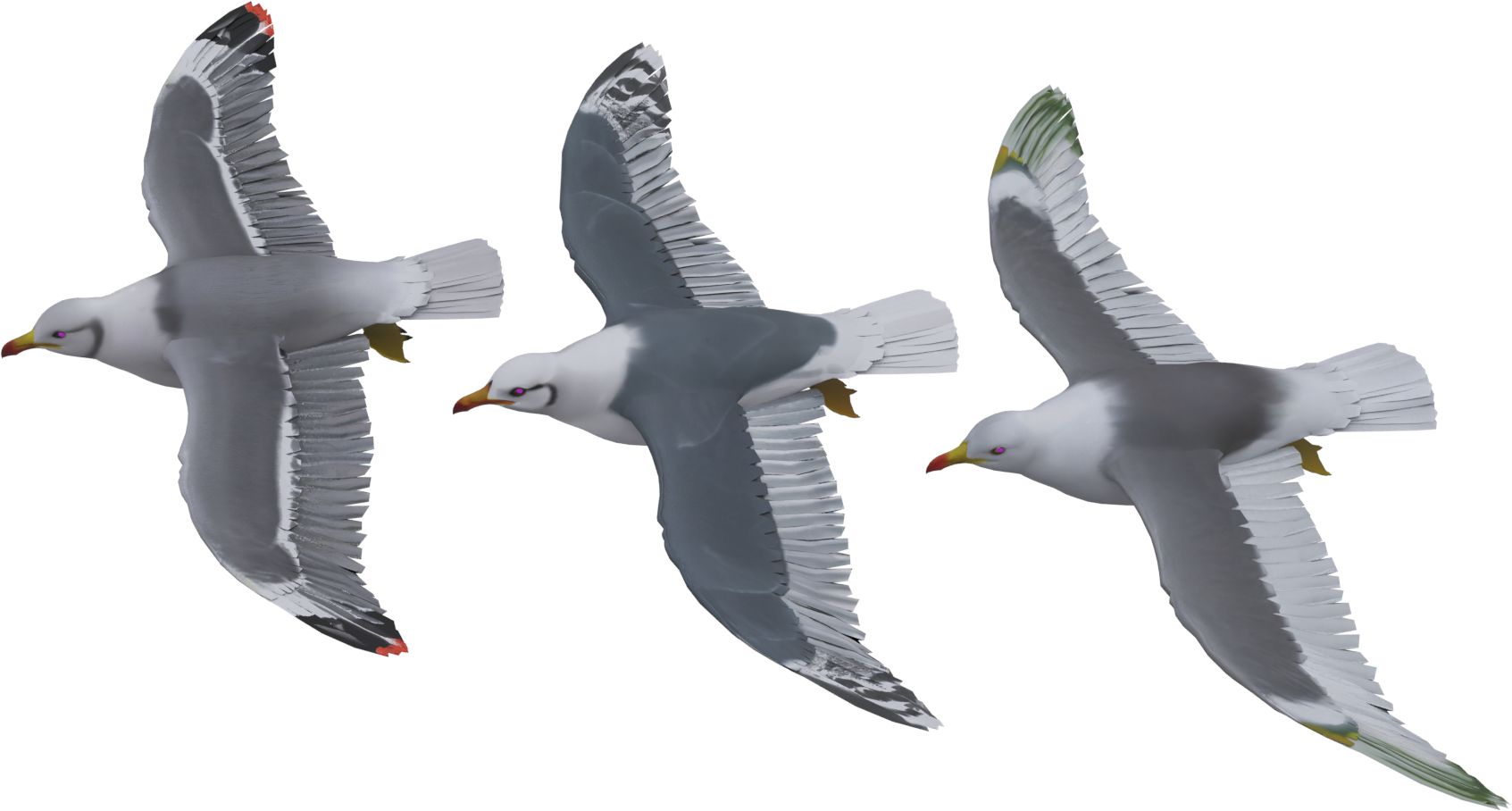} &
		\includegraphics[width=0.1\linewidth]{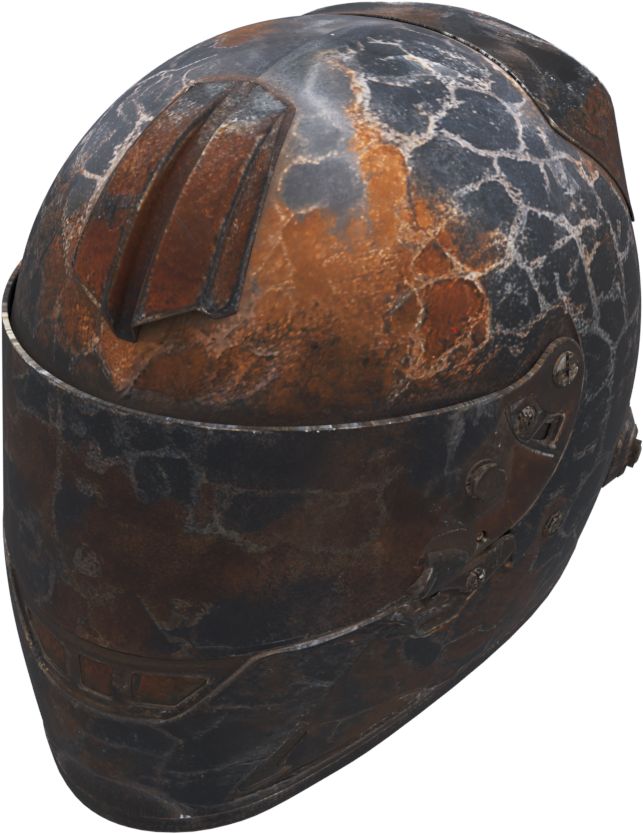} &
		\includegraphics[width=0.1\linewidth]{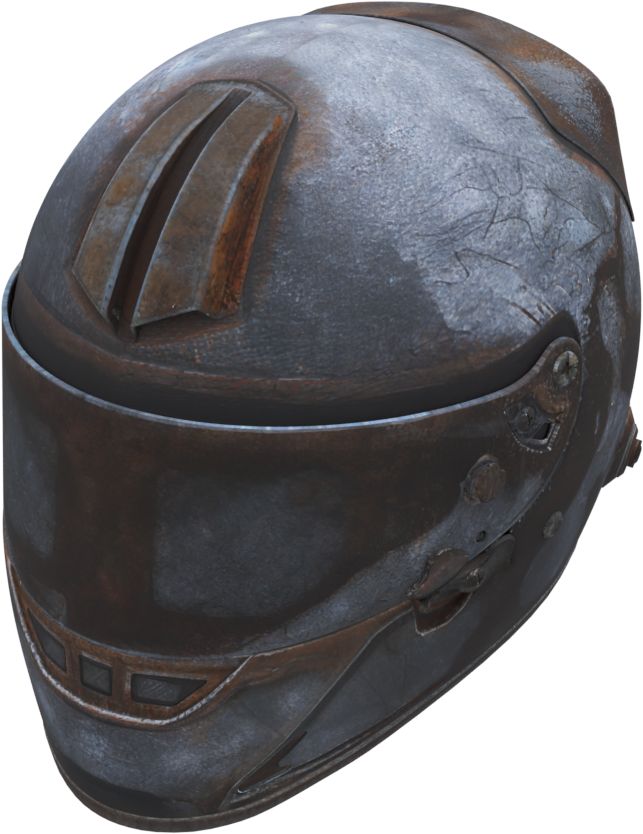} &
		\includegraphics[width=0.1\linewidth]{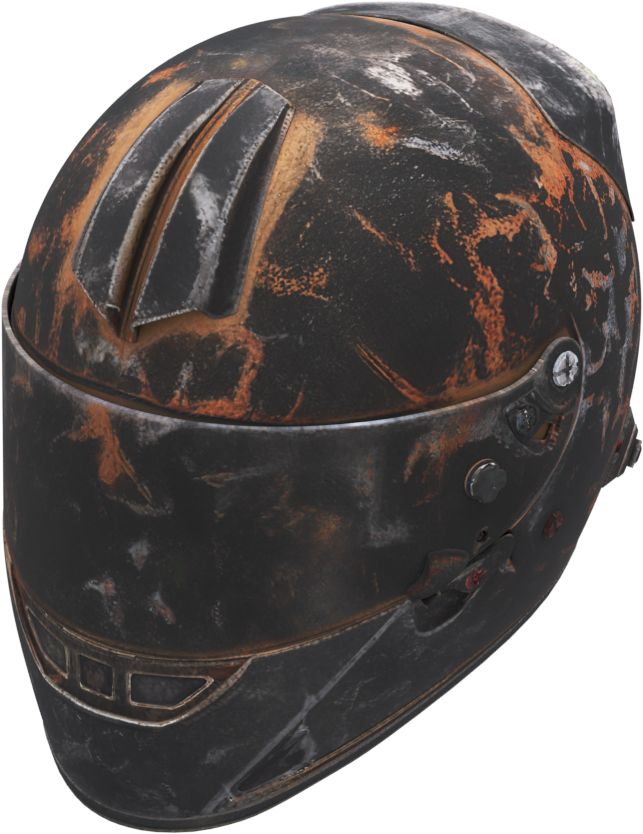} \\
		
		\includegraphics[width=0.12\linewidth]{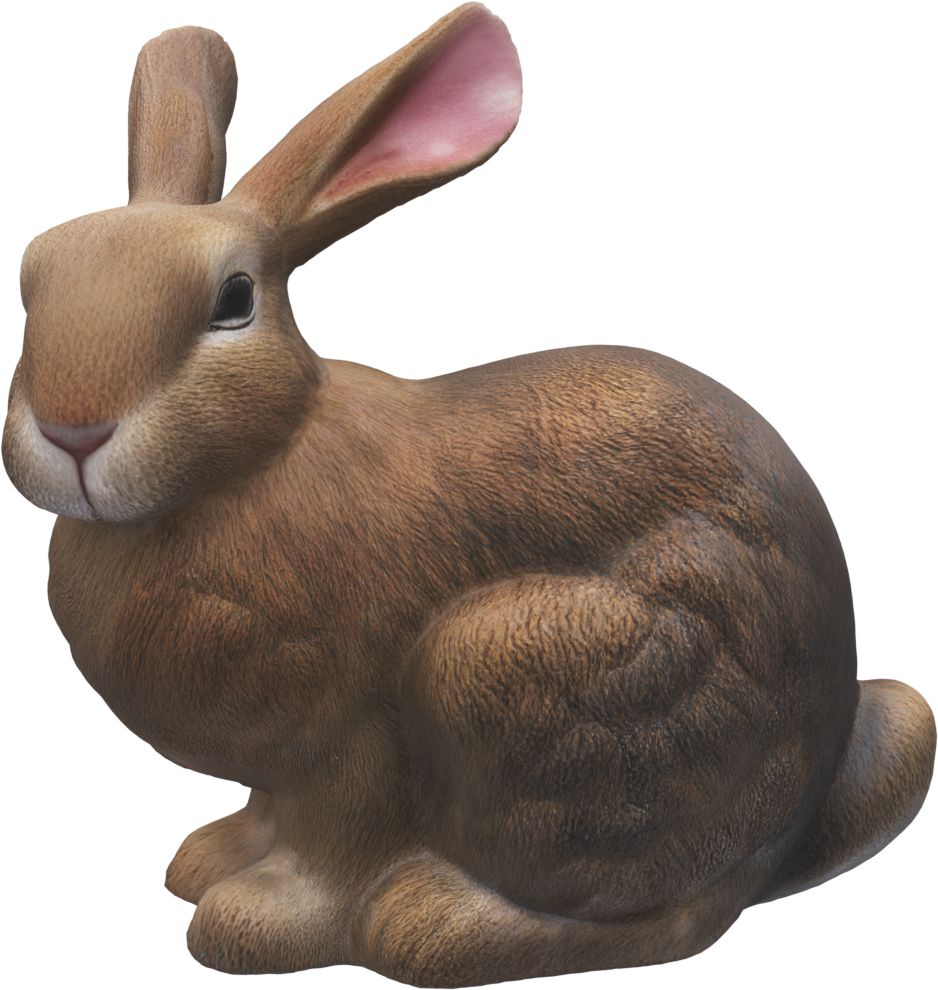} &
		\includegraphics[width=0.12\linewidth]{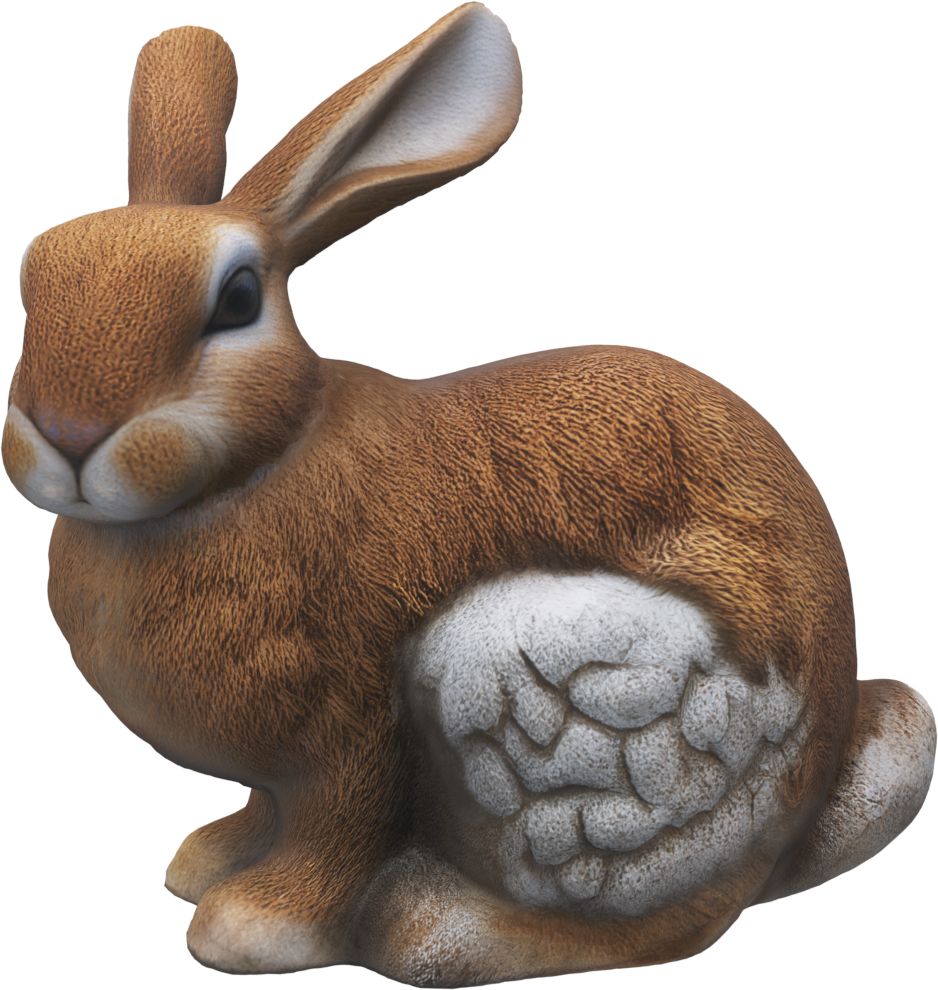} &
		\includegraphics[width=0.12\linewidth]{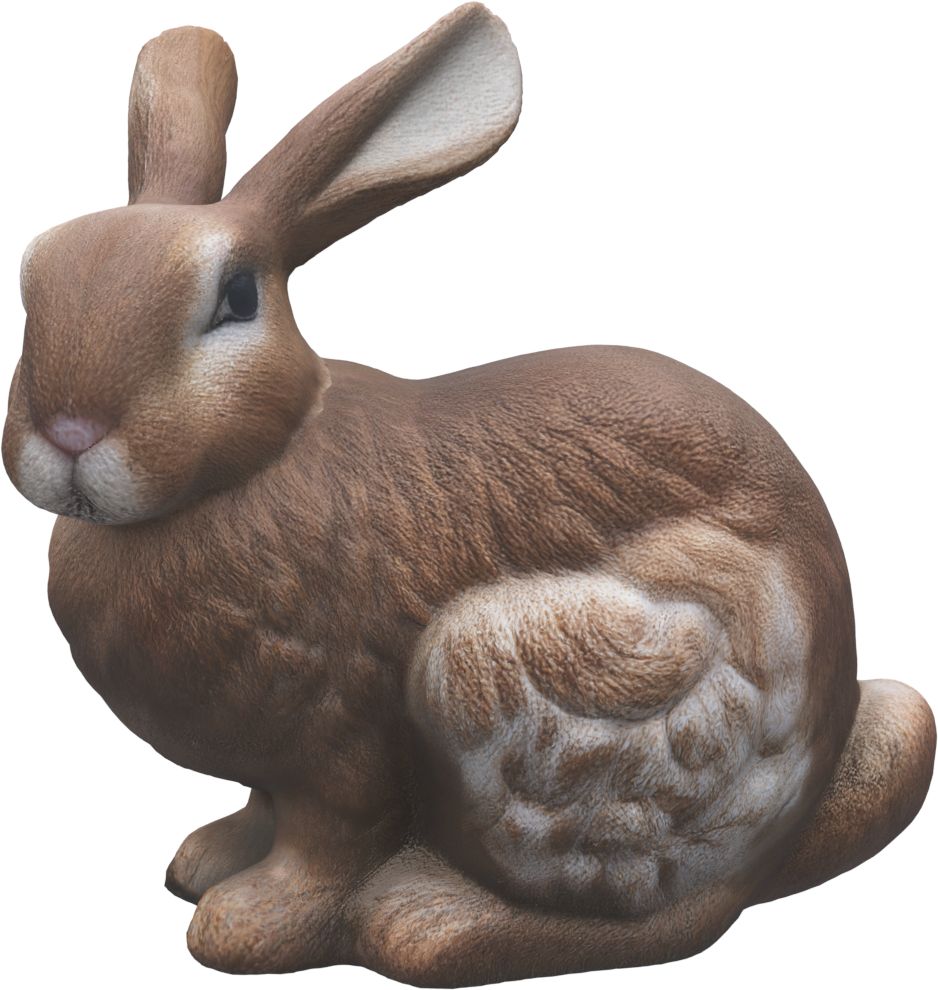} &
		\includegraphics[width=0.2\linewidth]{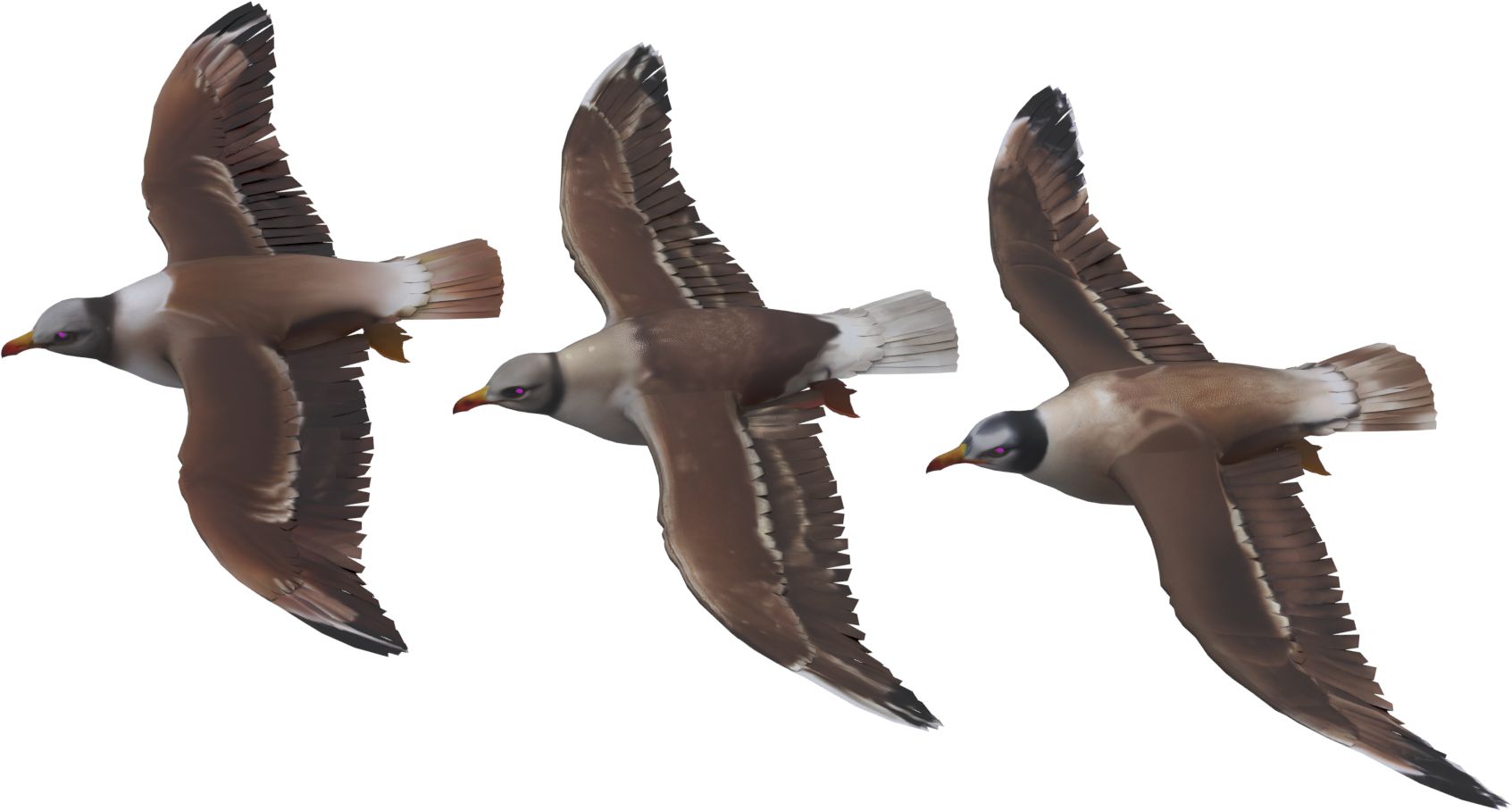} &
		\includegraphics[width=0.1\linewidth]{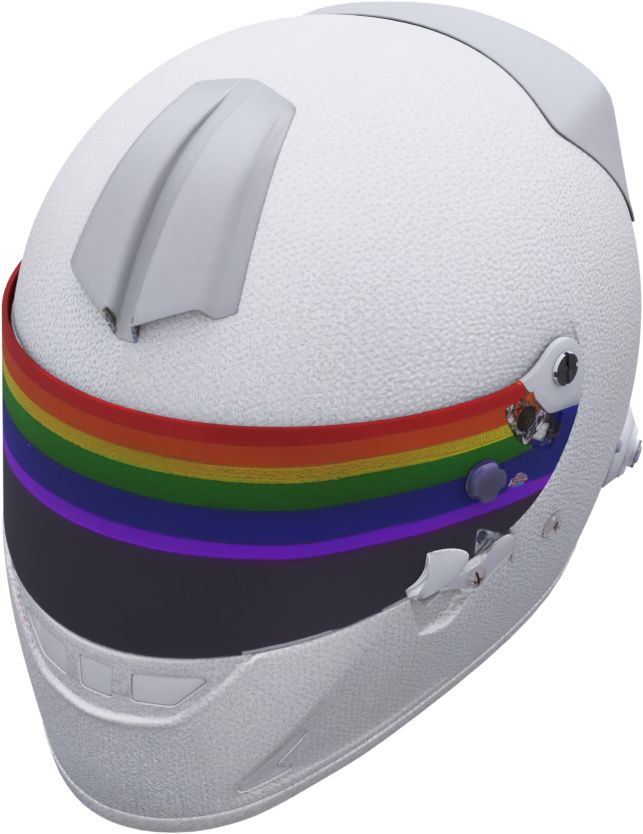} &
		\includegraphics[width=0.1\linewidth]{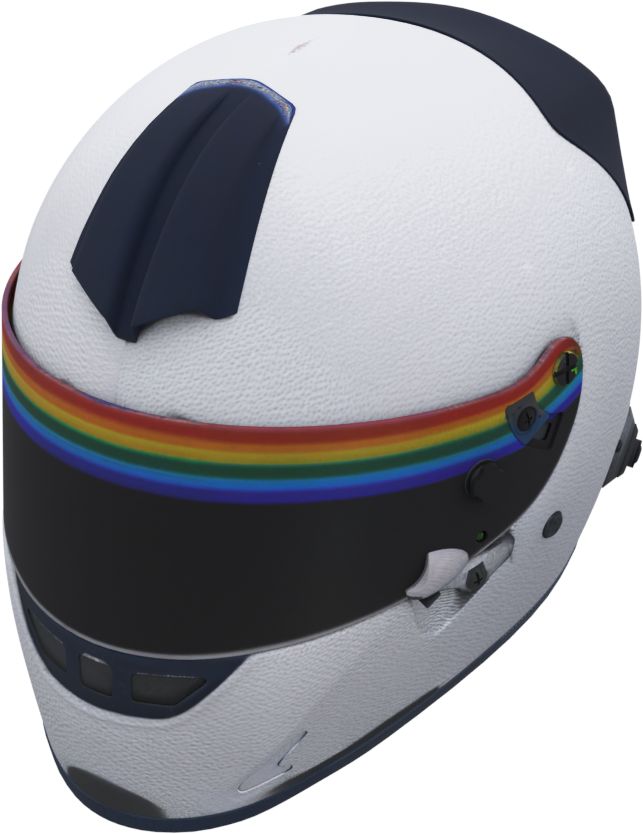} &
		\includegraphics[width=0.1\linewidth]{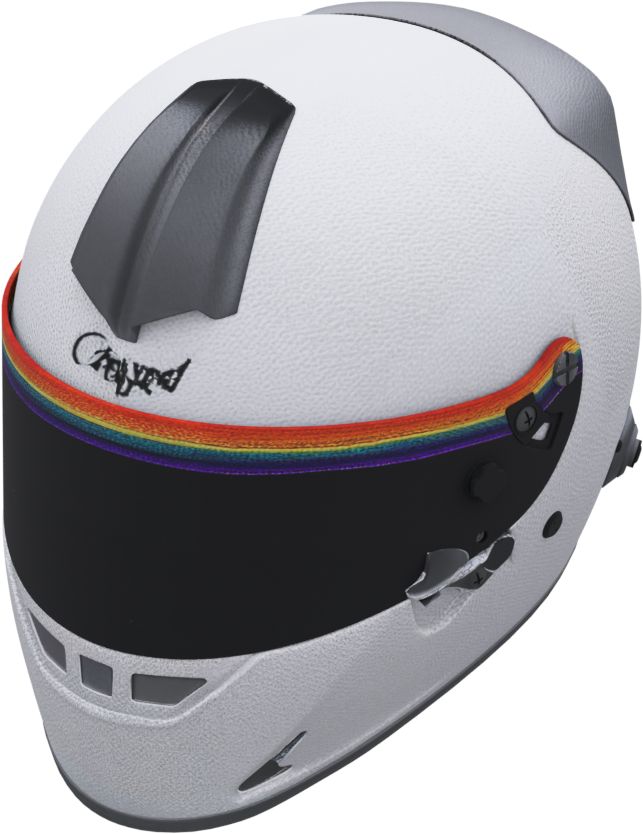} \\

		\includegraphics[width=0.12\linewidth]{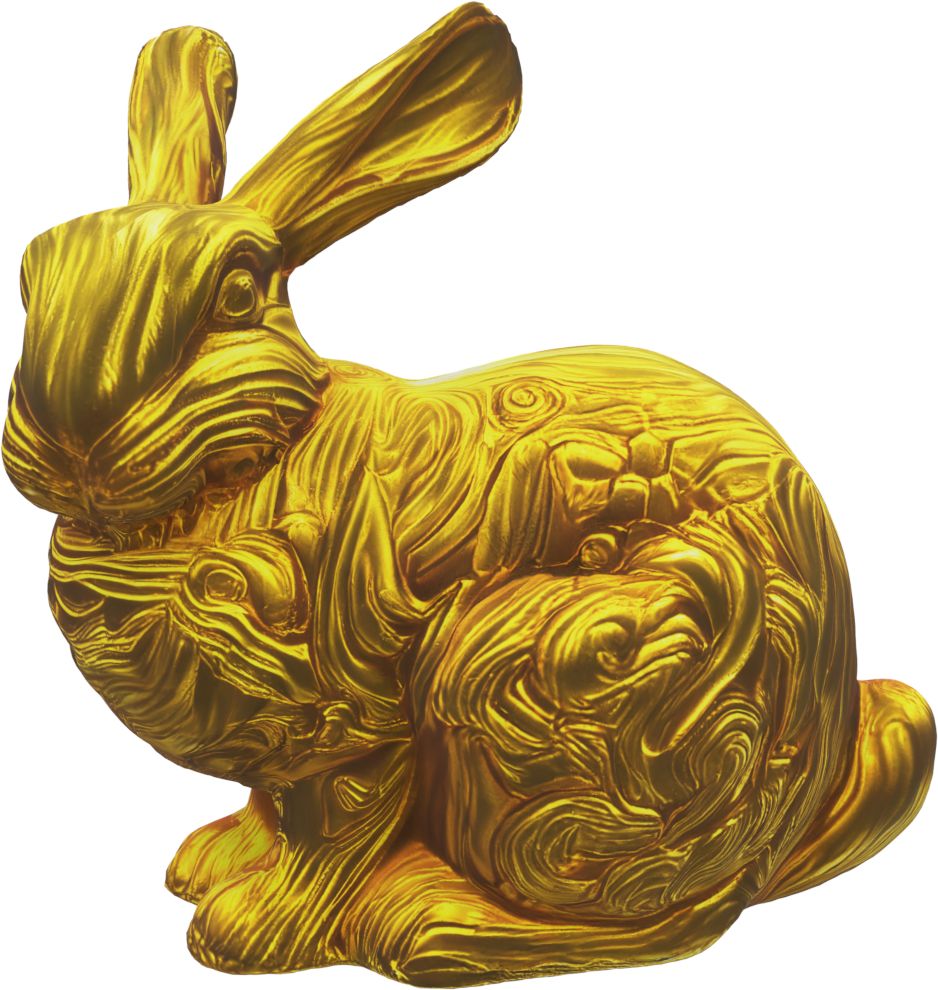} &
		\includegraphics[width=0.12\linewidth]{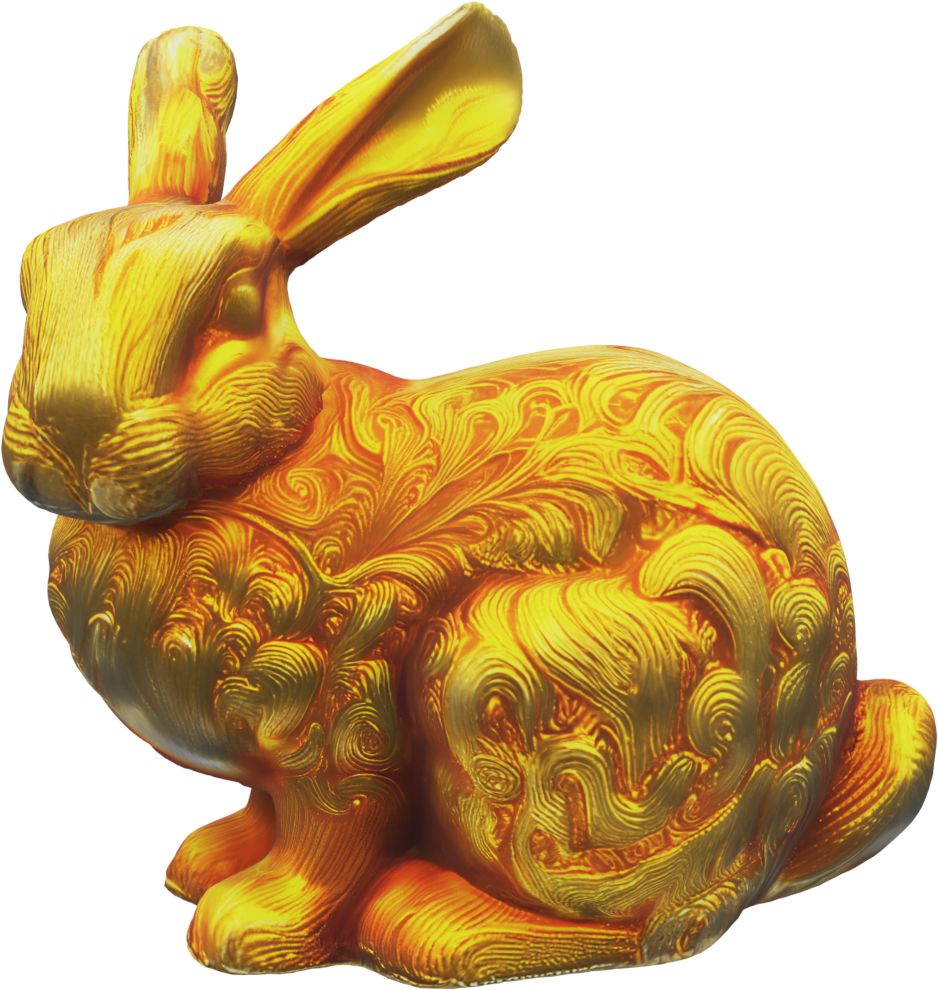} &
		\includegraphics[width=0.12\linewidth]{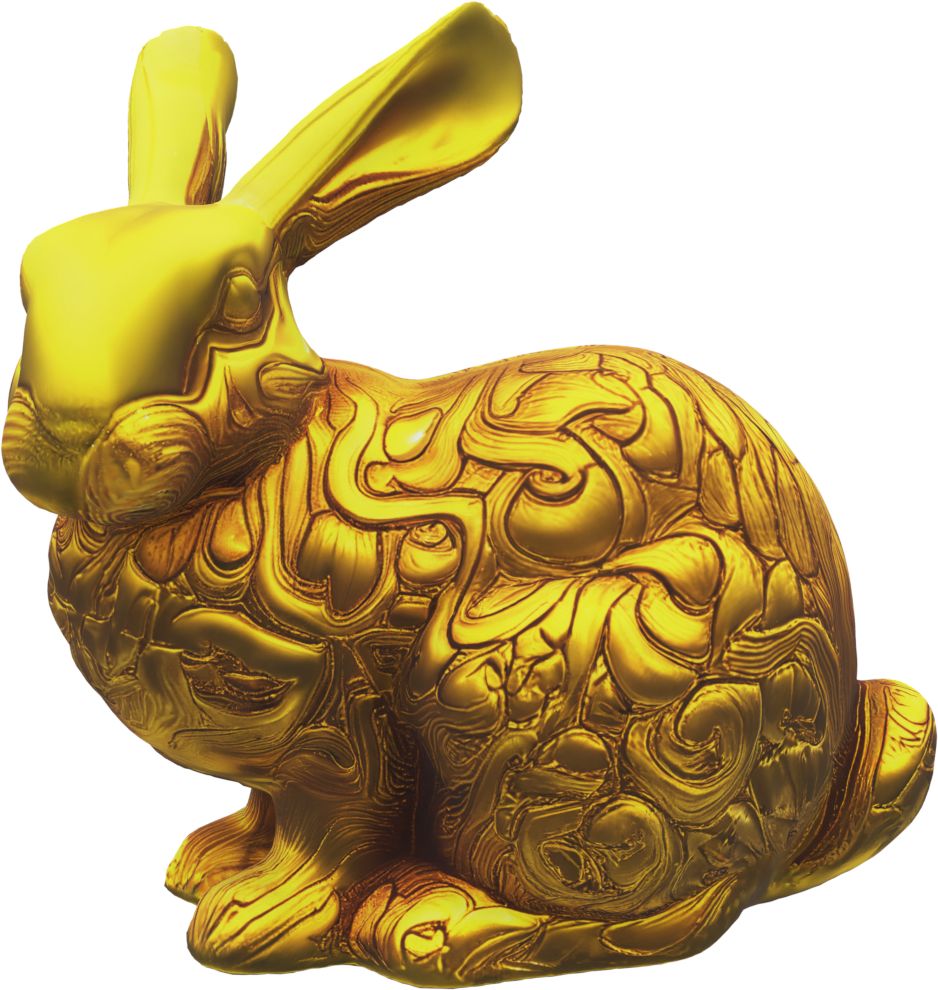} &
		\includegraphics[width=0.2\linewidth]{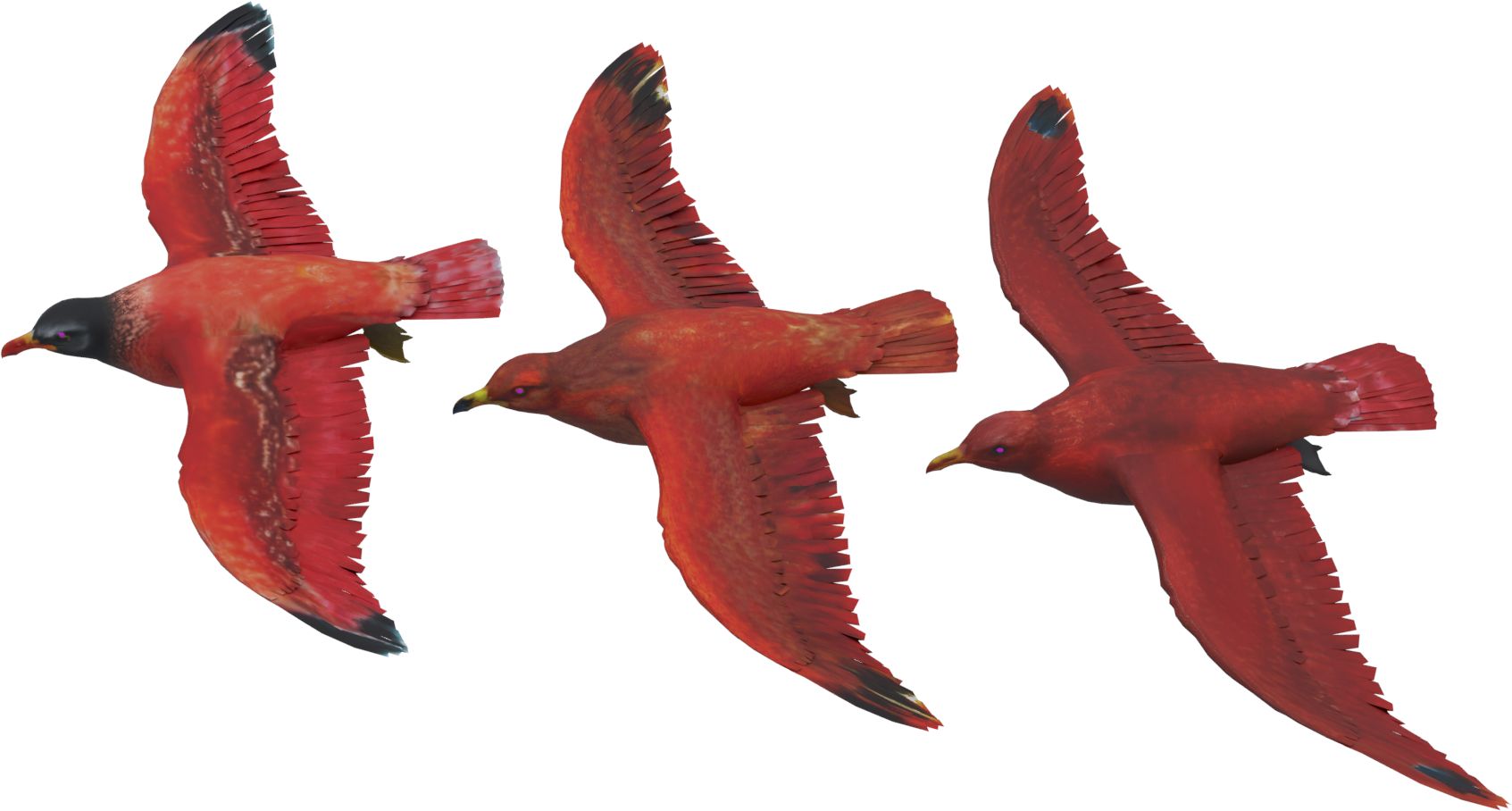} &
		\includegraphics[width=0.1\linewidth]{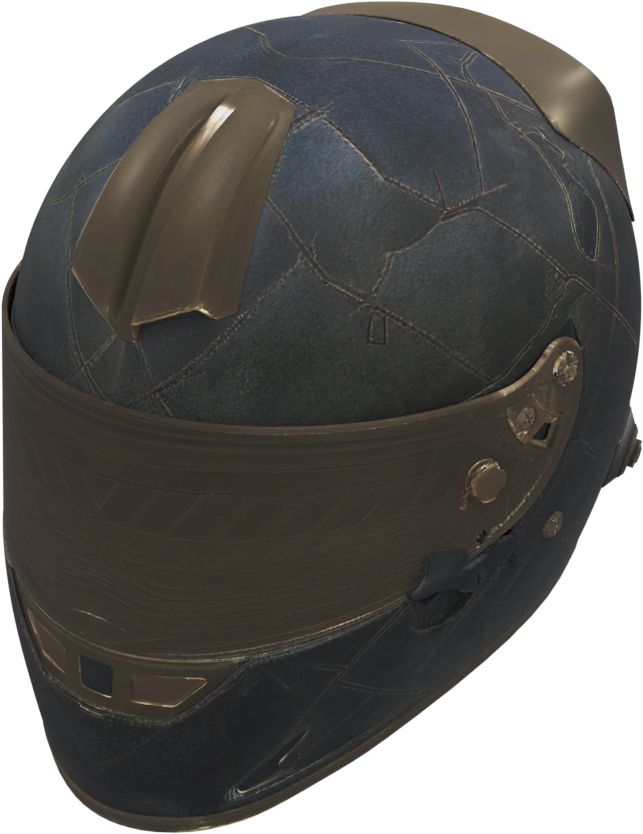} &
		\includegraphics[width=0.1\linewidth]{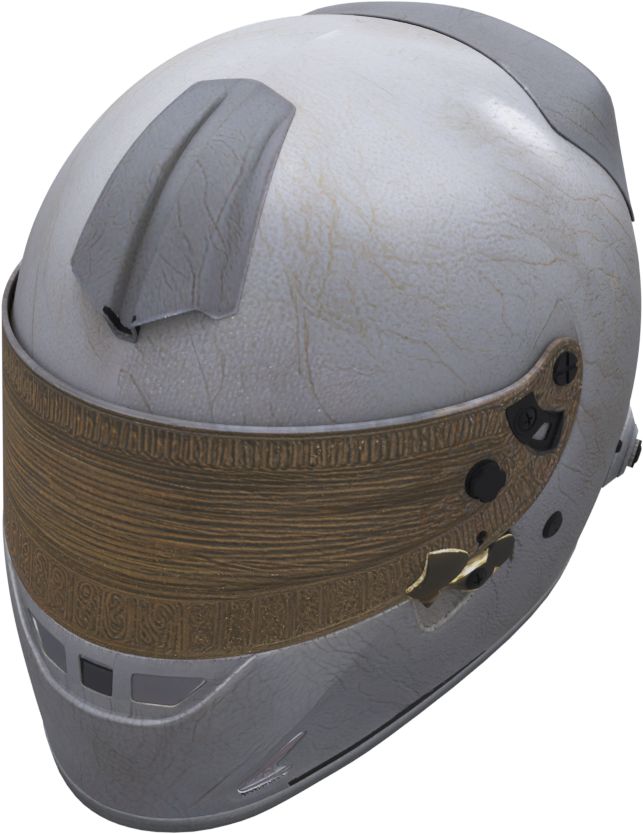} &
		\includegraphics[width=0.1\linewidth]{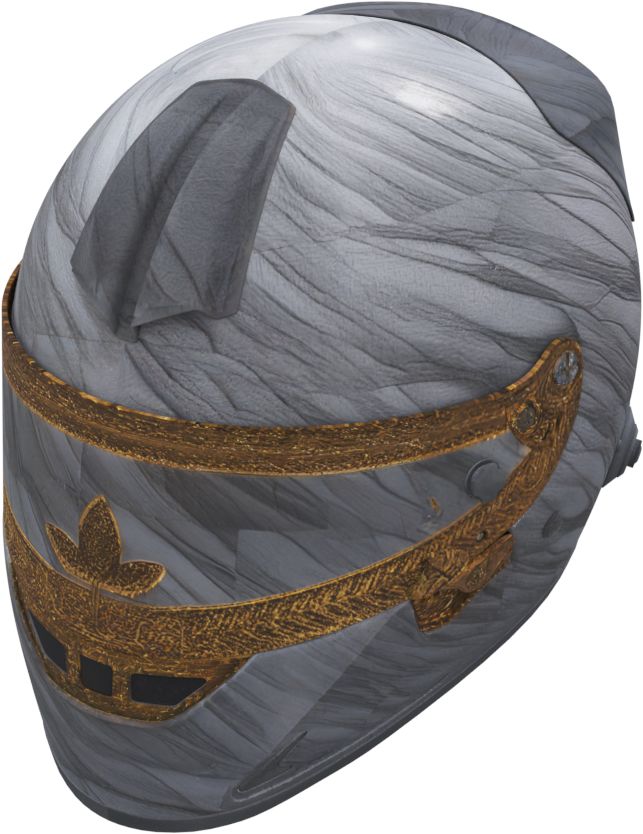} \\

        \ & (a) & \ & (b) & \ & (c) \\

	\end{tabular}

\caption{\textbf{Diverse samples.} For each column, each row was generated using the same prompt with a different seed.} 
  \label{fig:diversity_bunnies}
\end{figure*}